\documentclass{article}
\usepackage[affil-it]{authblk}

     \usepackage[final]{neurips_2019}

\usepackage[utf8]{inputenc} 
\usepackage[T1]{fontenc}    
\usepackage{hyperref}       
\usepackage{url}            
\usepackage{booktabs}       
\usepackage{amsfonts}       
\usepackage{nicefrac}       
\usepackage{microtype}      
\usepackage{amsmath}
\usepackage{underscore}
\usepackage{xcolor}
\newcommand{\todo}[1]{}
\renewcommand{\todo}[1]{{\color{red} {#1}}}
\usepackage{graphicx}
\usepackage{subfigure}
\usepackage{stmaryrd}
\usepackage{listings}
\usepackage{xr}
\usepackage{paralist}
\usepackage{caption}
\usepackage{multirow}
\usepackage{adjustbox} 

\usepackage{amsmath,amsfonts,bm}

\def\eqref#1{equation~\ref{#1}}

\def\1{\bm{1}}

\def\vtheta{{\bm{\theta}}}

\DeclareMathAlphabet{\mathsfit}{\encodingdefault}{\sfdefault}{m}{sl}
\SetMathAlphabet{\mathsfit}{bold}{\encodingdefault}{\sfdefault}{bx}{n}

\title{Anti-efficient encoding in emergent communication}

\author[1,2]{Rahma Chaabouni}
\author[1]{Eugene Kharitonov}
\author[1,2]{Emmanuel Dupoux}
\author[1,3]{Marco Baroni}
\affil[1]{Facebook AI Research}
\affil[2]{Cognitive Machine Learning (ENS - EHESS - PSL Research University - CNRS - INRIA)}
\affil[3]{ICREA}
\affil[ ]{\tt {\{rchaabouni,kharitonov,dpx,mbaroni\}@fb.com}}

\begin{document}

\maketitle

\begin{abstract}
  Despite renewed interest in emergent language simulations with
  neural networks, little is known about the basic properties of the
  induced code, and how they compare to human language. One
  fundamental characteristic of the latter, known as Zipf's Law of
  Abbreviation (ZLA), is that more frequent words are efficiently
  associated to shorter strings. We study whether the same pattern
  emerges when two neural networks, a ``speaker'' and a ``listener'',
  are trained to play a signaling game. Surprisingly, we find that
  networks develop an \emph{anti-efficient} encoding scheme, 
  in which the most frequent inputs are associated to the longest messages, 
  and messages in general are skewed towards the maximum length threshold. 
  This anti-efficient code appears easier to discriminate for the listener,
  and, unlike in human communication, the speaker does not impose a
  contrasting least-effort pressure towards brevity. Indeed, when the
  cost function includes a penalty for longer messages, the resulting
  message distribution starts respecting ZLA. Our analysis stresses
  the importance of studying the basic features of emergent
  communication in a highly controlled setup, to ensure the latter
  will not depart too far from human language. Moreover, we present a
  concrete illustration of how different functional pressures can lead
  to successful communication codes that lack basic properties of
  human language, thus highlighting the role such pressures play in
  the latter.
\end{abstract}

\section{Introduction} 

There is renewed interest in  simulating language emergence among
neural networks that interact to solve a task, motivated by the desire
to develop automated agents that can communicate with humans
\citep[e.g.,][]{Havrylov:Titov:2017,Lazaridou:etal:2017,Lazaridou:etal:2018,Lee:etal:2018b}. As part of this trend, several recent studies analyze the properties of the
emergent codes \citep[e.g.,][]{Kottur:etal:2017,
  Bouchacourt:Baroni:2018,Evtimova:etal:2018,Lowe:etal:2019,Graesser:etal:2019}. However,
these analyses generally consider relatively complex setups, when
very basic characteristics of the emergent codes have yet to be
understood. We focus here on one such characteristic, namely the length
distribution of the messages that two neural networks playing a simple
signaling game come to associate to their inputs, in function of input
frequency.

In his pioneering studies of lexical statistics,
George Kingsley Zipf noticed a robust trend in human language that
came to be known as Zipf's Law of Abbreviation (ZLA): There is an
inverse (non-linear) correlation between word frequency and length
\citep{Zipf:1949,teahan2000,sigurd2004word,Strauss:etal:2007}. Assuming that
shorter words are easier to produce, this is an efficient encoding
strategy, particularly effective given Zipf's other important
discovery that word distributions are highly skewed, following a power-law distribution. Indeed, in this way language approaches an optimal code in
information-theoretic terms \citep{Cover:Thomas:2006}. Zipf, and many
after him, have thus used ZLA as evidence that language is shaped by
functional pressures toward effort minimization
\citep[e.g.,][]{piantadosi2011,Mahowald:etal:2018,Gibson:etal:2019}. However, others %
\citep[e.g.,][]{mandelbrot1954,miller,ferrer2011,moscoso2013} noted
 that some random-typing distributions also respect ZLA, casting doubts
 on functional explanations of the observed pattern.

We study a \emph{Speaker} network that gets one
out of $1K$ distinct one-hot vectors as input,
randomly drawn from a power-law distribution (so that
frequencies are extremely skewed, like in natural language). Speaker
transmits a variable-length \emph{message} to a \emph{Listener}
network. Listener outputs a one-hot vector, and the networks are
rewarded if the latter is identical to the input. There is no
direct supervision on the message, so that the networks are free to
create their own ``language''. The networks develop a successful
communication system that does \emph{not} exhibit ZLA, and is indeed
\emph{anti-efficient}, in the sense that all messages are long, and
the most frequent inputs are associated to the longest
messages. Interestingly, a similar effect is observed in artificial human
communication experiments, in conditions in which 
longer messages do not demand extra effort to speakers, so that they
are preferred as they ease the listener discrimination task
\citep{kanwal2017}. Our Speaker network, unlike humans, has no
physiological pressure towards brevity \citep{Chaabouni:etal:2019b}, and our Listener network
displays an \emph{a priori} preference for longer messages. Indeed,
when we penalize Speaker for producing longer strings, the emergent
code starts obeying ZLA. We examine the implications of our findings
in the Discussion.

\section{Setup} 

\subsection{The game}
We designed a variant of the Lewis signaling game
\citep{david1969convention} in which the input distribution follows a
power-law distribution. We think of these inputs as a vocabulary of
distinct abstract \emph{word types}, to which the agents will assign
specific word forms while learning to play the game. We leave it to
further research to explore setups in which word type and
form distributions co-evolve
\citep{FerrerICancho:DiazGuilera:2007}. Importantly, our basic
inefficient encoding result also holds when the inputs are uniformly
distributed (Appendix \ref{sec:unifrom}). Formally, the game proceeds as follows:

\begin{enumerate}
\item The Speaker network receives one of $1K$ distinct one-hot vectors
  as input $i$. Inputs are not drawn uniformly, but, like
  in natural language, from a power-law
  distribution. That is, the $r^{th}$ most frequent input $i_r$ has probability
  $\frac{1}{r \times \sum_{k=1}^{1000}{\frac{1}{k}}}$ to be sampled,
  with $r \in \llbracket 1,..., 1000 \rrbracket $. Consequently, the
  probability of sampling the $1^{st}$ input is $0.13$ while the
  probability of sampling the $1000^{th}$ one is $1000$ times lower.
\item Speaker chooses a sequence of symbols from its alphabet $A = \{s_1, s_2 ..., s_{a-1}, \texttt{eos}\}$ of size $|A| =a$ to construct a message $m$, terminated as soon as Speaker produces the `end-of-sequence' token $\texttt{eos}$. If Speaker has not yet emitted $\texttt{eos}$ at $\texttt{max_len}-1$, it is stopped and $\texttt{eos}$ is appended at the end of its message (so that all messages are suffixed with $\texttt{eos}$ and no message is longer than $\texttt{max_len}$). 
\item The Listener network consumes $m$ and outputs $\hat{i}$.
\item The agents are successful if $i=\hat{i}$, that is, Listener reconstructed Speaker's input.
\end{enumerate}

The game is implemented using the EGG toolkit \citep{kharitonov2019egg}, and the code can be found at \url{https://github.com/facebookresearch/EGG/tree/master/egg/zoo/channel}.

\subsection{Architectures}
As standard in current emergent-language simulations \citep[e.g.,][]{Lazaridou:etal:2018}, both agents are implemented as single-layer LSTMs~\citep{Hochreiter:Schmidhuber:1997}. Speaker's input is a $1K$-dimensional one-hot vector $i$, and the output is a sequence of symbols, defining message $m$. This sequence is generated as follows. A linear layer maps the input vector into the initial hidden state of Speaker's LSTM cell. Next, a special start-of-sequence symbol is fed to the cell. At each step of the sequence, the output layer defines a Categorical distribution over the alphabet. At training time, we sample from this distribution. During evaluation, we select the symbol greedily. Each selected symbol is fed back to the LSTM cell. The dimensionalities of the hidden state vectors are part of the hyper-parameters we explore (Appendix \ref{sec:hyperparams}). Finally, we initialize the weight matrices of our agents with a uniform distribution with support in [$-\frac{1}{\sqrt{\texttt{input_size}}}$, $\frac{1}{\sqrt{\texttt{input_size}}}$], where $\texttt{input_size}$ is the dimensionality of the matrix input (Pytorch default initialization).

Listener consumes the entire message $m$, including \texttt{eos}. After \texttt{eos} is received, Listener's hidden state is passed through a fully-connected layer with softmax activation, determining a Categorical distribution over $1K$ indices. This distribution is used to calculate the cross-entropy loss w.r.t.\ the ground-truth input, $i$.

The joint Speaker-Listener architecture can be seen as a discrete auto-encoder \citep{autoencoder}.

\subsection{Optimization}

The architecture is not directly differentiable, as messages are discrete-valued. In language emergence, two approaches are dominantly used: Gumbel-Softmax relaxation~\citep{Maddison2016,Jang2016} and REINFORCE~\citep{Williams1992}.  We also experimented with the approach of \cite{Schulman2015}, combining REINFORCE and stochastic backpropagation to estimate gradients. Preliminary experiments showed that the latter algorithm (to be reviewed next) results in the fastest and most stable convergence, and we used it in all the following experiments. However, the main results we report were also observed with the other algorithms, when successful.


We denote by $\vtheta_s$ and $\vtheta_l$ the  Speaker and Listener parameters, respectively. $\mathcal{L}$ is the cross-entropy loss, that takes the ground-truth one-hot vector $i$ and Listener's output $L(m)$ distribution as inputs. 
We want to minimize the expectation of the cross-entropy loss $\mathbb{E} ~ \mathcal{L}(i, L(m))$, where the expectation is calculated w.r.t.\ the joint distribution of inputs and message sequences.  The gradient of the following surrogate function is an unbiased estimate of the gradient $\nabla_{\vtheta_s \cup \vtheta_l}\mathbb{E} ~ \mathcal{L}(i, L(m))$:
\begin{equation}
    \label{eq:surrogate}
    \mathbb{E} \left[ \mathcal{L}(i, L(m; \vtheta_l)) + \left( \{ \mathcal{L}(i, L(m; \vtheta_l) \} - b \right)  \log P_s(m | \vtheta_s)  \right]
\end{equation}
where $\{\cdot\}$ is the stop-gradient operation, $P_s(m | \vtheta_s)$ is the probability of producing the sequence $m$ when Speaker is parameterized with vector $\vtheta_s$, and $b$ is a running-mean baseline used to reduce the estimate variance without introducing a bias. To encourage exploration, we also apply an entropy regularization term~\citep{Williams1991} on the output distribution of the speaker agent.

Effectively, under Eq.~\ref{eq:surrogate}, the gradient of the loss w.r.t.\ the Listener parameters is found via conventional backpropagation (the first term in Eq.~\ref{eq:surrogate}), while Speaker's gradient is found with a REINFORCE-like procedure (the second term). Once the gradient estimate is obtained, we feed it into the Adam~\citep{Kingma2014} optimizer. We explore different learning rate and  entropy regularization coefficient values (Appendix \ref{sec:hyperparams}).

We train agents for $2500$ episodes, each consisting of $100$
mini-batches, in turn including $5120$ inputs sampled from the
power-law distribution with replacement. After training, we present to
the system each input once, to compute accuracy by giving equal
weight to all inputs, independently of amount of training exposure.

\subsection{Reference distributions}

As ZLA is typically only informally defined, we introduce 3 reference distributions that display efficient encoding and arguably respect
ZLA.

\subsubsection{Optimal code} 

Based on standard coding theory \citep{Cover:Thomas:2006}, we design an \emph{optimal code} (OC) guaranteeing  the shortest
average message length given a certain alphabet size and the
constraint that all messages must end with \texttt{eos}. The \emph{shortest} messages are deterministically associated
to the \emph{most frequent} inputs, leaving longer ones for less
frequent ones. The length of the message associated to an input
is determined as follows. Let
$A=\{s_1, s_2 ... s_{a-1}, \texttt{eos}\}$ be the alphabet of size $a$
and $i_r$ be the $r^{th}$ input when ranked by frequency. Then
$i_r$ is mapped to a message of length
\begin{equation}
	l_{i_r} = min\{n:  \sum_{k=1}^{n}{(a-1)^{k-1}} \ge r\}
\end{equation}
For instance, if $a=3$, then there is only one message of length $1$ (associated to the most frequent referent), $2$ of length $2$, $4$ of length $3$ etc.\footnote{There is always only one message of length 1 (that is, \texttt{eos}), irrespective of alphabet size.} Section 2 of \citet{ferrer2013} presents a proof of how this encoding is the maximally efficient one.

\subsubsection{Monkey typing}
\label{sec:is}
Natural languages respect ZLA without being as efficient as OC. It has
been observed that \emph{Monkey typing} (MT) processes, whereby a
monkey hits random typewriter keys including a space character, produce word
length distributions remarkably similar to those attested in natural languages
\citep{simon, miller}. We thus adapt a MT process to our setup, as a less
strict benchmark for network efficiency.\footnote{No actual monkey was harmed in the definition of the process.} 


We first sample an input without replacement according to the power-law distribution, then generate the message to be associated with it. We repeat the process until all inputs are assigned a unique message. The message is constructed by letting a monkey hit the $a$ keys of a typewriter uniformly at random ($p=1/a$), subject to these constraints:
\begin{inparaenum}[(i)]
\item The message ends when the monkey hits $\texttt{eos}$.
\item A message cannot be longer than a specified length $\texttt{max_len}$. 
    If the monkey has not yet emitted $\texttt{eos}$ at $\texttt{max_len}-1$, it is stopped and $\texttt{eos}$ is appended at the end of the message.
\item If a generated message is identical to one already used, it is rejected and another is generated.
\end{inparaenum}

For a given length $l$, there are only $(a-1)^{l-1}$ different messages. Moreover, for a random generator with the $\texttt{max_len}$ constraint, the probability of generating a message of length $l$ is: 
\begin{equation}\label{eq:pl}
P_l=p \times (1- p)^{l-1}, \mbox{if } l<\texttt{max_len} \mbox{ and } P_{\texttt{max_len}}=(1- p)^{\texttt{max_len}-1}
\end{equation}
From these calculations, we derive two qualitative
observations about MT. First, as we fix \texttt{max_len} and increase $a$
(decrease $p=1/a$), more generated messages will reach
\texttt{max_len}. Second, when $a$ is small and
\texttt{max_len} is large (as in early MT studies where \texttt{max_len}
was infinite), a ZLA-like distribution emerges, due to the finite
number of \emph{different} messages of length $l$. Indeed, for any $l$
less than \texttt{max_len}, $P_l$ strictly decreases as $l$
grows. Then, for given inputs, the monkey is likely to start by
generating messages of the most probable length (that is, $1$). As we
exhaust all unique messages of this length, the process starts
generating messages of the next probable length (i.e., $2$) and so
on. 
Figure \ref{fig:intermittentSilence} in Appendix \ref{sec:Intersilence} confirms experimentally that our MT distribution respects ZLA for $a\leq10$ and various $\texttt{max_len}$.

\subsubsection{Natural language}
\label{sec:naturallanguages}

We finally consider word length distributions in natural
language corpora. We used pre-compiled English, Arabic, Russian and
Spanish frequency lists from \url{http://corpus.leeds.ac.uk/serge/},
extracted from corpora of internet text containing between ~$200M$
(Russian) and ~$16M$ words (Arabic). For direct comparability with
input set cardinality in our simulations, we only looked at the
distribution of the top $1000$ most frequent words, after merging
lower- and upper-cased forms, and removing words containing
non-alphabetical characters. The resulting word frequency distributions 
obeyed power laws  with  exponents between $-0.81$ and $-0.92$ 
(we used $-1$ to generate our inputs). Alphabet sizes are as follows: $30$
(English), $31$ (Spanish), $47$ (Russian), $59$ (Arabic). These are
larger than normative  sizes, as unfiltered Internet text will
occasionally include foreign characters (e.g., accented letters in
English text). Contrary to previous reference distributions, we cannot control
$\texttt{max_len}$ and alphabet size.  We hence compare human and
network distributions only in the adequate settings. In the main text,
we present results for the languages with the smallest (English) and
largest (Arabic) alphabets. The distributions of the other
languages are comparable, and presented in Appendix \ref{sec:NatLang}.


\section{Experiments} 

\subsection{Characterizing the emergent encoding}
\label{sec:emergentenc}
We experiment with alphabet sizes $a \in [3, 5, 10 , 40, 1000]$. We
chose mainly small alphabet sizes to minimize a potential bias
in favor of long messages: For high $a$, randomly generating long
messages becomes more likely, as the probability of outputting
$\texttt{eos}$ at random becomes lower. At the other extreme, we also
consider $a=1000$, where the Speaker could in principle successfully communicate
using at most $2$-symbol messages (as Speaker needs to produce $\texttt{eos}$). 
Finally, $a=40$ was chosen to be close to the alphabet size of the
natural languages we study (mean alphabet size: $41.75$).

After fixing $a$, we choose $\texttt{max_len}$ so that agents have
enough capacity to describe the whole input space ($|I|=1000$). For
a given $a$ and $\texttt{max_len}$, Speaker cannot encode more
inputs than the message space size
$M_{a}^{\texttt{max_len}} =
\sum_{j=1}^{\texttt{max_len}}{(a-1)^{j-1}}$. We experiment with
$\texttt{max_len} \in [2,6,11,30]$. We couldn't use higher values
because of memory limitations. Furthermore, we studied the effect of
$D=\frac{M_{a}^{\texttt{max_len}}}{|I|}$. While making sure that this
ratio is at least $1$, we experiment with low values, where Speaker
would have to use nearly the whole message space to successfully
denote all inputs. We also considered settings with
significantly larger $D$, where constructing $1K$
distinct messages might be an easier task.

We train models for each $(\texttt{max_len}, a)$ setting and agent
hyperparameter choice (4 seeds per choice). We consider runs successful if, after training, they
achieve an accuracy above $99\%$ on the full input set (i.e., less than
$10$ miss-classified inputs). As predicted, the higher
$D$ is, the more accurate the agents become. Indeed, agents
need much larger $D$ than strictly necessary in order to
converge. We select for further analysis only those
$(\texttt{max_len}, a)$ choices that resulted in more than $3$
successful runs (mean number of successful runs across the reported
configurations is $25$ out of $48$). Moreover,
we focus here on configurations with $\texttt{max_len}=30$, as the
most comparable to natural language.\footnote{Natural languages have
  no rigid upper bound on length, and 30 is the highest
  \texttt{max_len} we were able to train models for. Qualitative
  inspection of the respective corpora suggest that 30 is anyway a
  reasonable ``soft'' upper bound on word length in the languages we studied
  (longer strings are mostly typographic detritus).} We present
results for all selected configurations (confirming the same trends)
in Appendix \ref{sec:antieff}.

Figure \ref{fig:3encodings} shows message length distribution
(averaged across all successful runs) in function of input
frequency rank, compared to our reference distributions. The MT results are averaged across $25$ different runs. We show the Arabic and English distributions in the plot containing the most comparable simulation settings $(30, 40)$.

Across configurations, we observe that Speaker messages greatly depart
from ZLA. There is a clear general preference for longer messages,
that is strongest \emph{for the most frequent inputs}, where Speaker
outputs messages of length $\texttt{max_len}$. That is, in the
emergent encoding, more frequent words are longer, making the system
obey a sort of ``\emph{anti}-ZLA'' (see Appendix \ref{sec:radomtest}
for confirmation that this anti-efficient pattern is statistically
significant). Consequently, the emergent language distributions are
well above all reference distributions, except for MT with $a=1000$,
where the large alphabet size leads to uniformly long words, for
reasons discussed in Section \ref{sec:is}. Finally, the lack of
efficiency in emergent language encodings is also observed when inputs
are uniformly distributed (see Appendix \ref{sec:unifrom}).

Although some animal signing systems disobey ZLA, due to specific environmental constraints \citep[e.g.,][]{heesen2019}, a large survey of human and animal communication did not find any case of significantly \emph{anti-}efficient systems \citep{ferrer2013}, making our finding particularly intriguing.

\begin{figure*}[ht]
\centering
\subfigure[\hspace{-.3\baselineskip}$\texttt{max_len}$=$30$, $a$=$5$]{
    \hspace{-1.5\baselineskip}
    \includegraphics[width=0.25\textwidth, keepaspectratio]{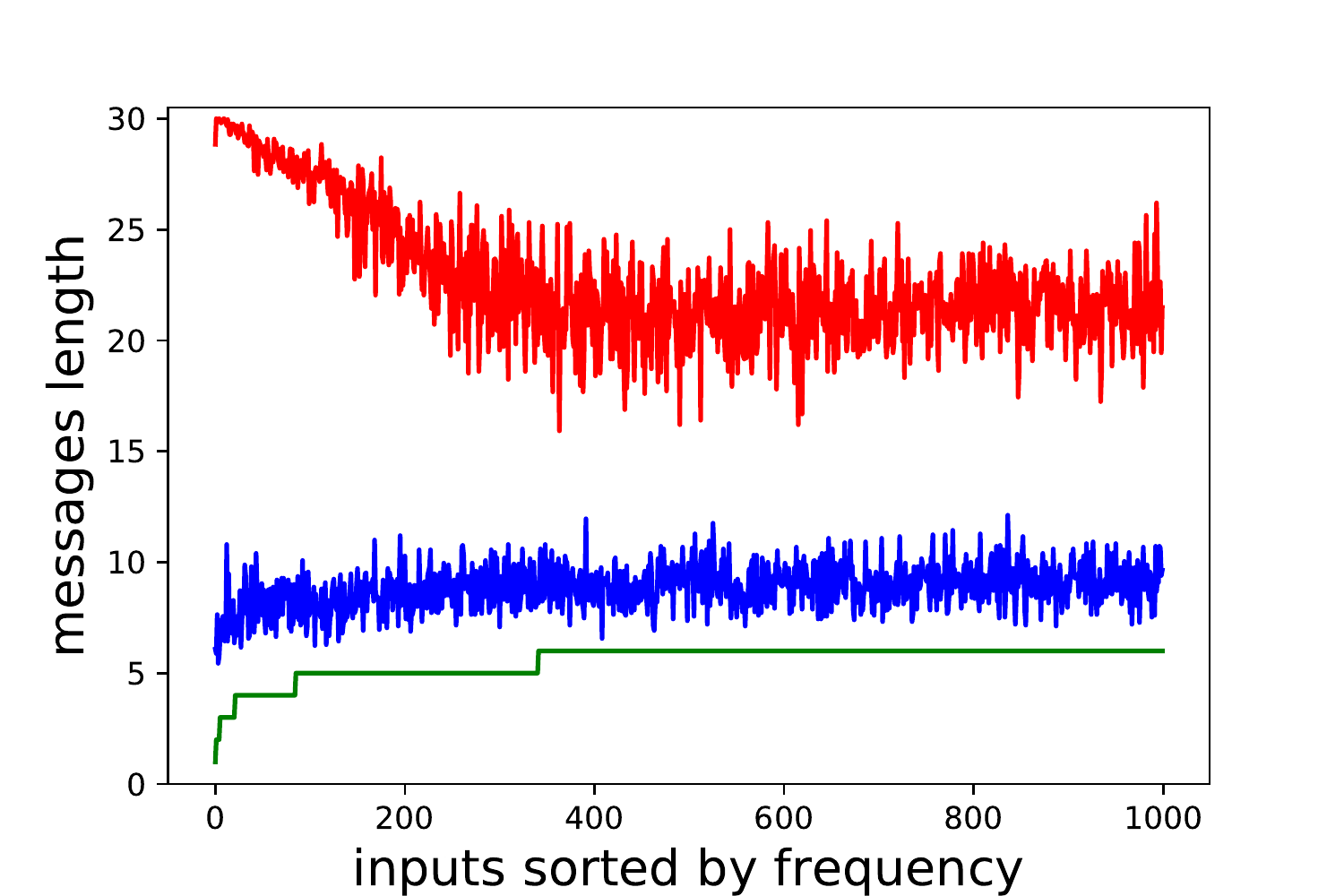}    
}
\subfigure[\hspace{-.3\baselineskip}$\texttt{max_len}$=$30$, $a$=$10$]{
    \hspace{-1.5\baselineskip}
    \includegraphics[width=0.25\textwidth, keepaspectratio]{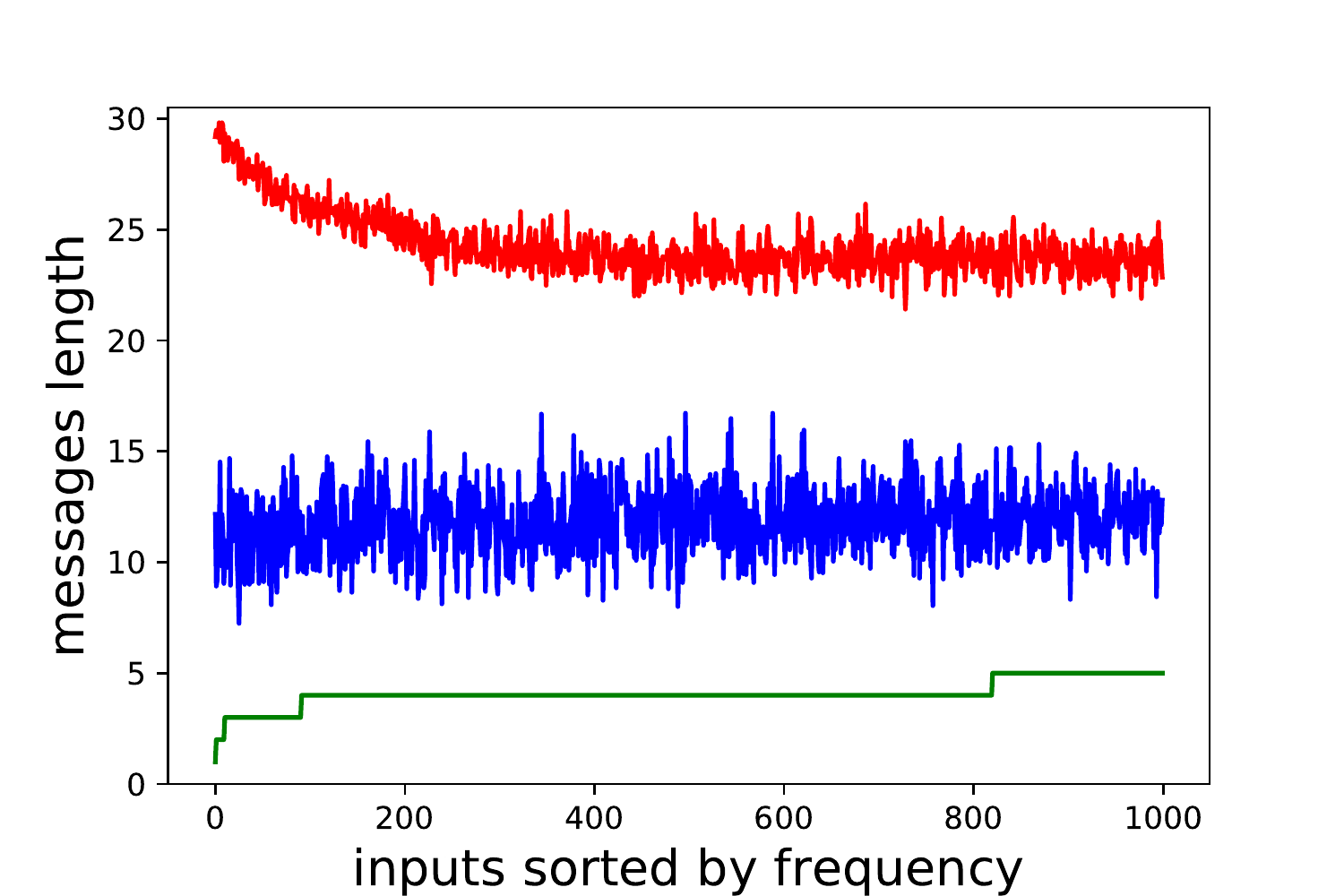}    
}
\subfigure[\hspace{-.4\baselineskip}$\texttt{max_len}$=$30$, $a$=$40$]{
    \hspace{-1.5\baselineskip}
    \includegraphics[width=0.25\textwidth, keepaspectratio]{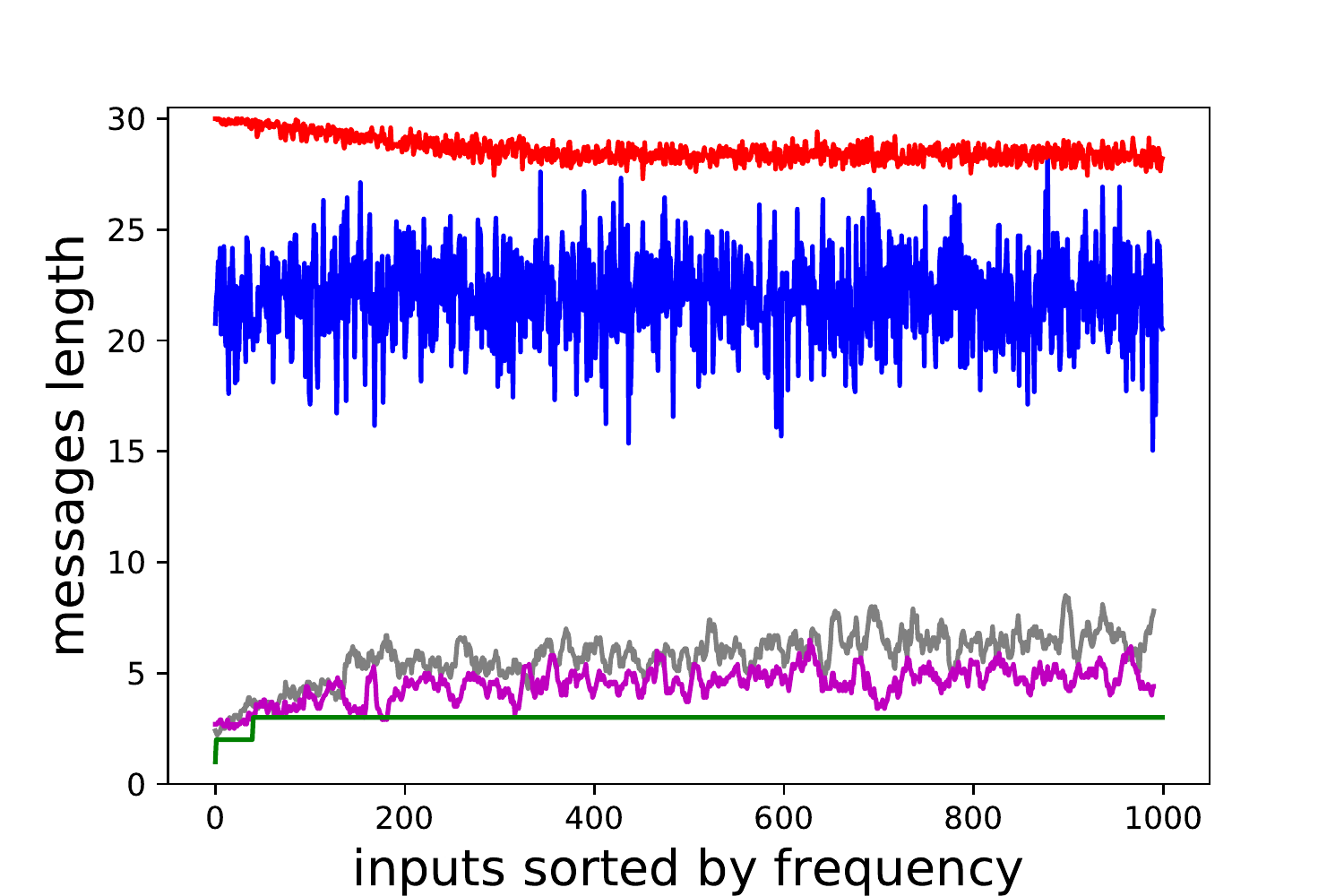}    
}
\subfigure[\hspace{-0.3\baselineskip}$\texttt{max_len}$=$30$, $a$=$1000$]{
    \hspace{-1.5\baselineskip}
    \includegraphics[width=0.25\textwidth, keepaspectratio]{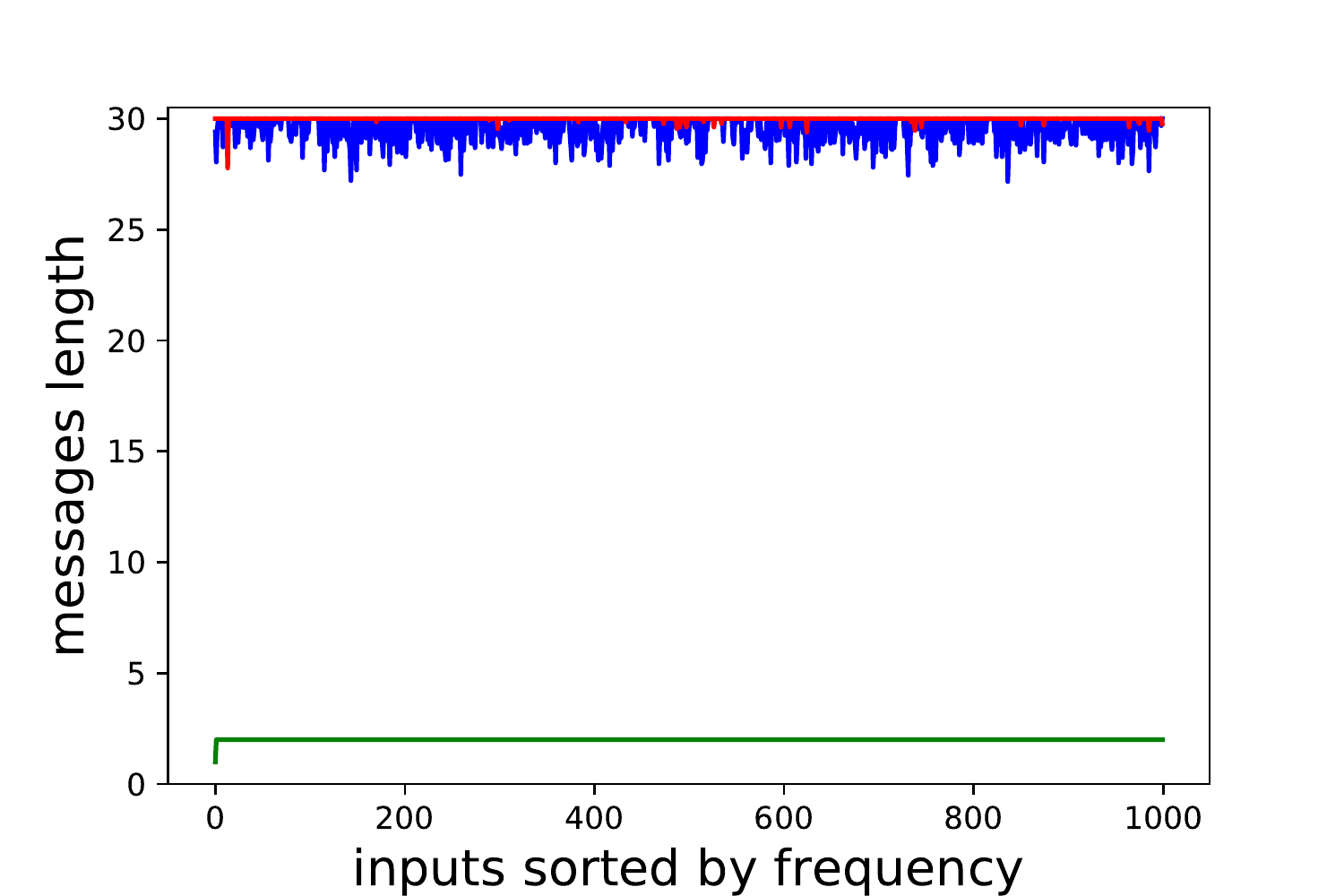} 
}
{
    \includegraphics[width=\textwidth]{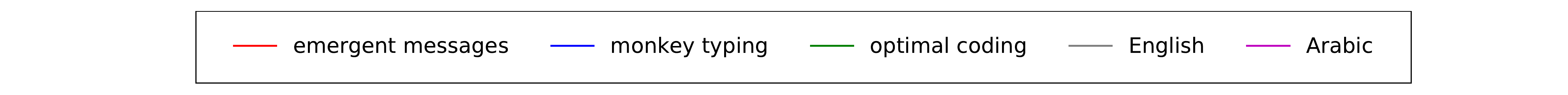}    
}

\caption{Mean message length across successful runs as a function of input frequency rank, with reference distributions. For readability, we smooth natural language distributions by reporting the sliding average of $10$ consecutive lengths.}\label{fig:3encodings}
\end{figure*}

\subsection{Causes of anti-efficient encoding}


We explore the roots of anti-efficiency by looking at the behavior of untrained Speakers and Listeners. Earlier work conjectured that ZLA emerges from the competing pressures to communicate in a perceptually distinct \emph{and} articulatorily efficient manner \citep{Zipf:1949,kanwal2017}. For our networks, there is a clear pressure from Listener in favour of ease of message discriminability , but Speaker has no obvious reason to save on ``articulatory'' effort. We thus predict that the observed pattern is driven by a Listener-side bias. 


\subsubsection{Untrained Speaker behavior}
\label{sec:speakerbias}

%
For each $i$ drawn from the power-law distribution without replacement, we get a message $m$ from $90$ distinct \emph{untrained} Speakers ($30$ speakers for each hidden size in $[100, 250, 500]$). We experiment with $2$ different association processes. In the first, we associate the first generated $m$ to $i$, irrespective of whether it was already associated to another input. In the second, we keep generating a $m$ for $i$ until we get a message that was not already associated to a distinct input. The second version is closer to the  MT process (see Section \ref{sec:is}). Moreover, message uniqueness is a reasonable constraint, since, in order to succeed, Speakers need first of all to keep messages denoting different inputs apart.

\begin{figure*}[ht]
\centering
\subfigure[$\texttt{max_len}=30$, $a=3$]{
    \includegraphics[width=0.33\textwidth, keepaspectratio]{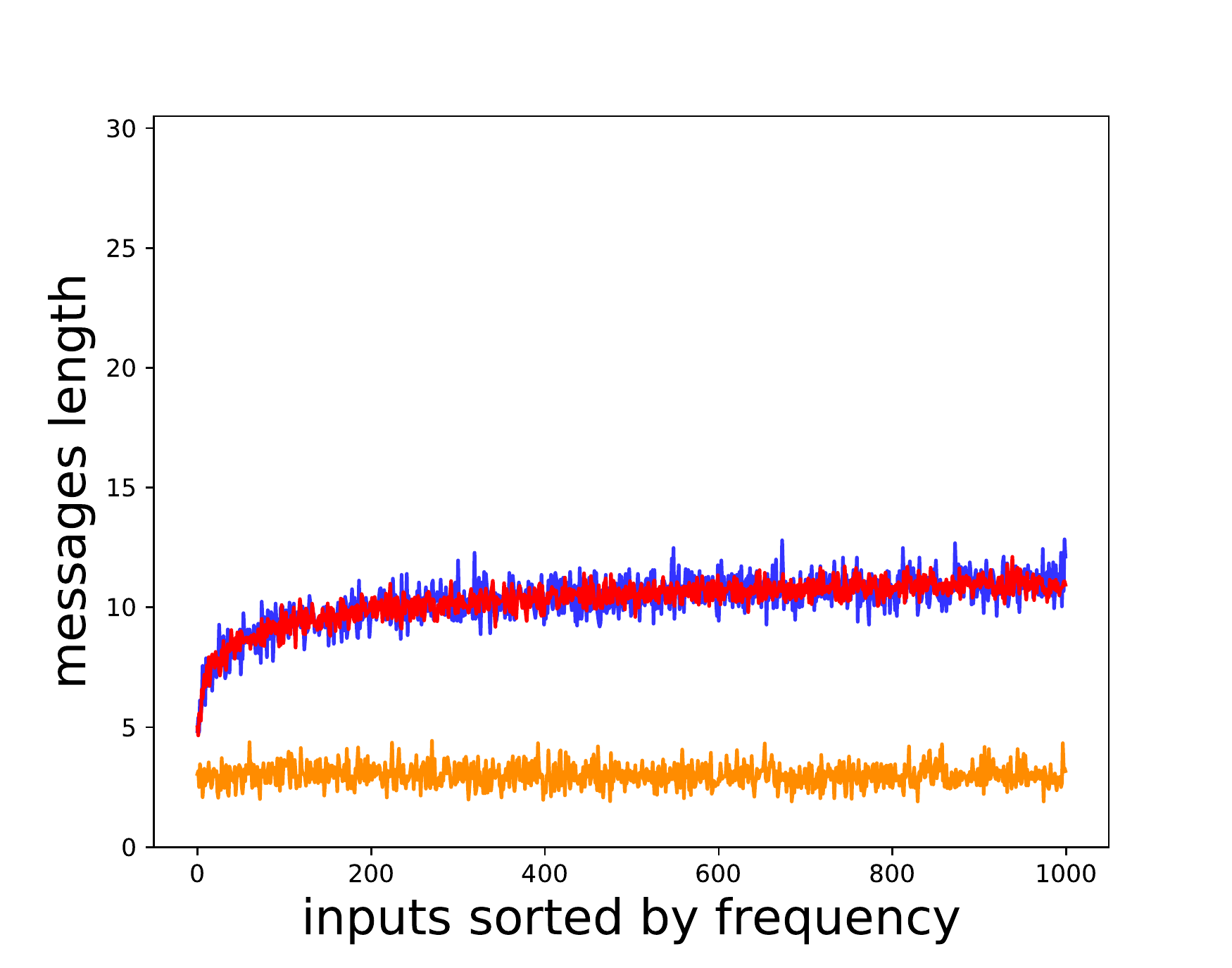} 
    \label{fig:A}
}
\subfigure[$\texttt{max_len}=30$, $a=5$]{
   \hspace{-1.5\baselineskip}
    \includegraphics[width=0.33\textwidth, keepaspectratio]{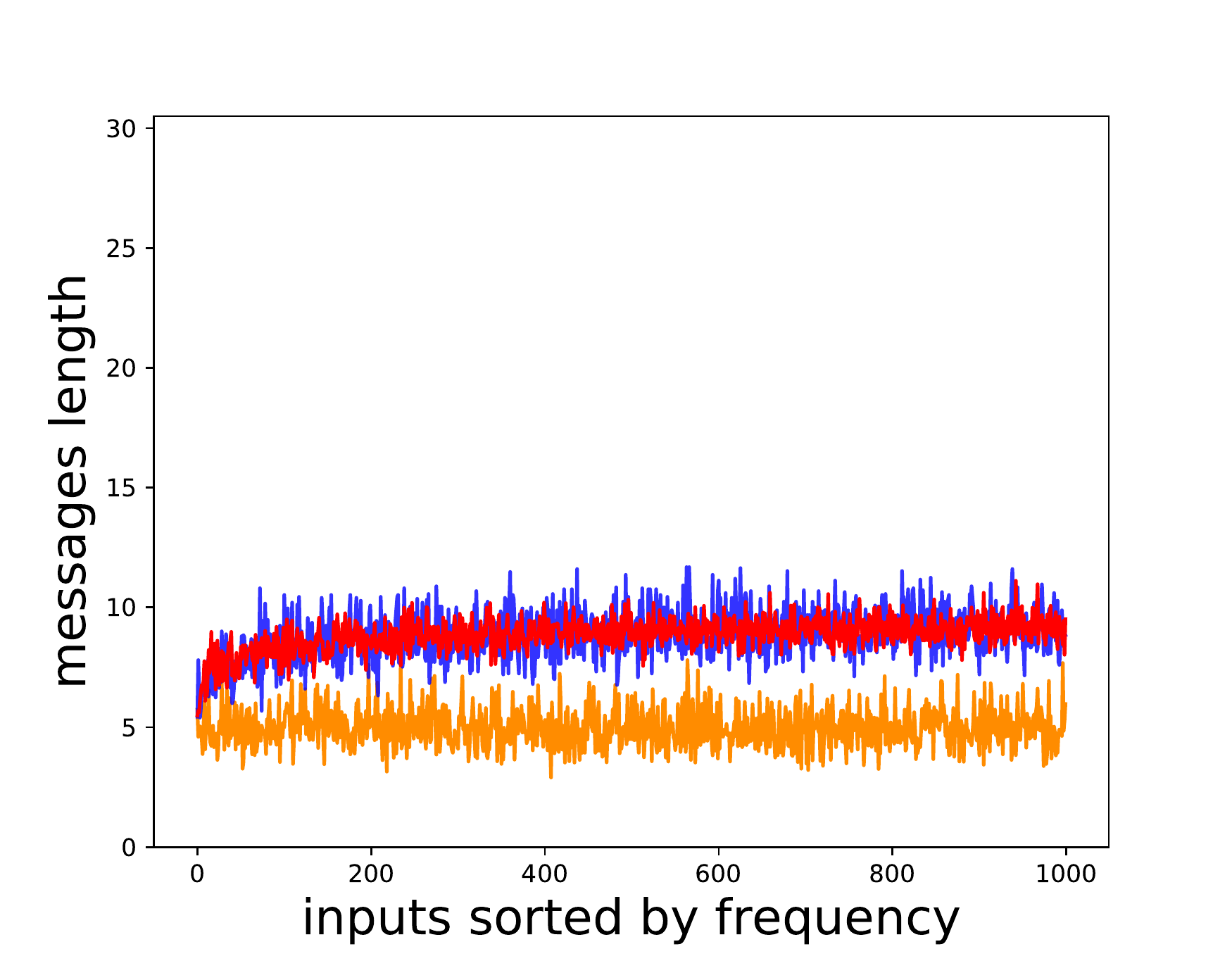} 
    \label{fig:A}
}
\subfigure[$\texttt{max_len}=30$, $a=40$]{
    \hspace{-1.5\baselineskip}	
    \includegraphics[width=0.33\textwidth, keepaspectratio]{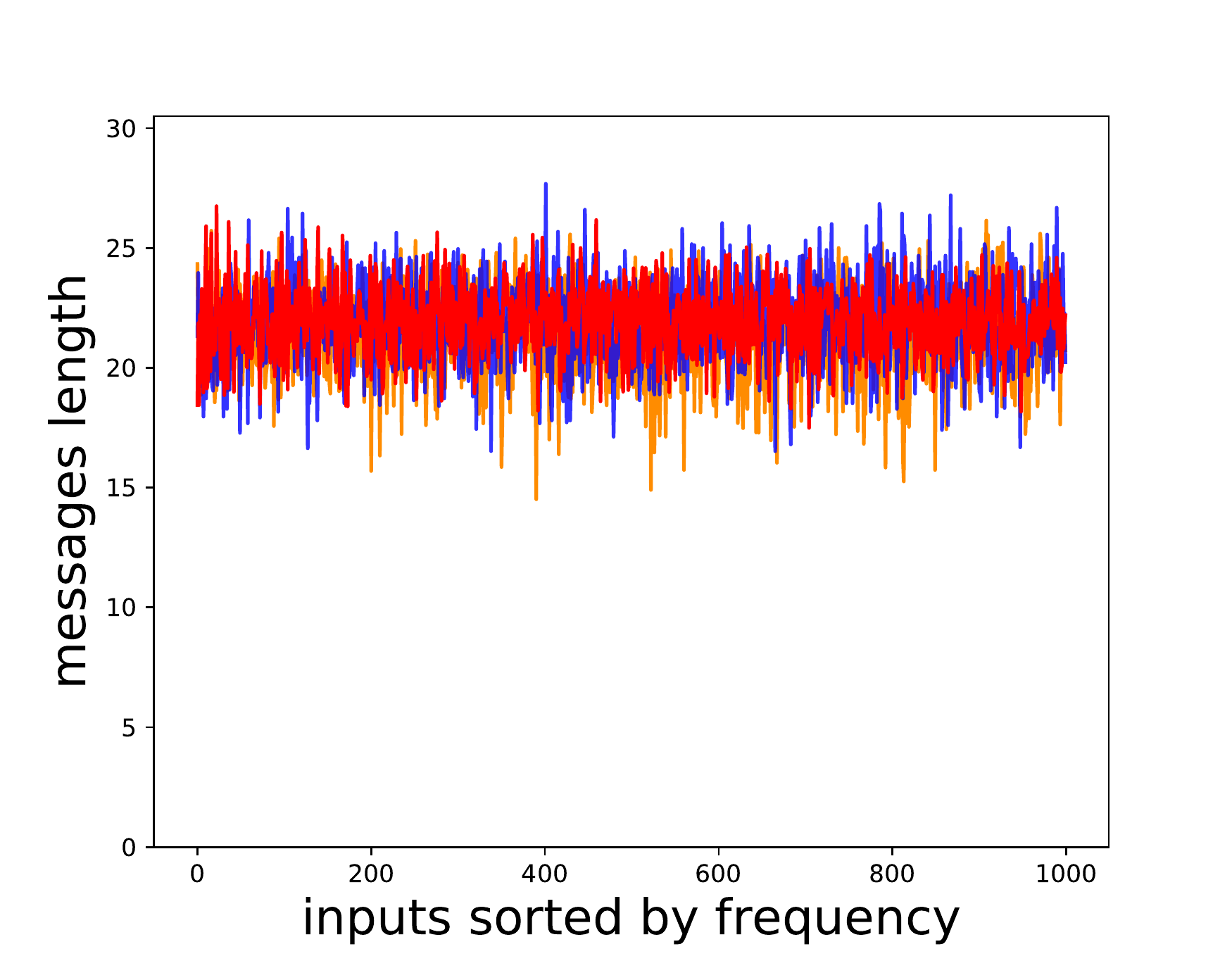}
}
{
    \includegraphics[width=\textwidth]{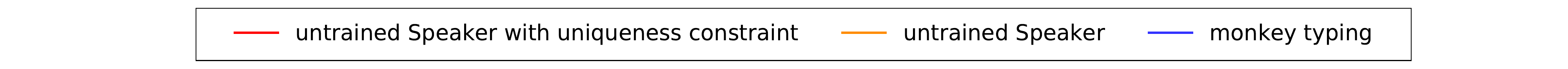}    
}
\caption{Average length of messages by input frequency rank for untrained Speakers, compared to MT. See Appendix \ref{sec:speakerb} for more settings.}\label{fig:SendBias}
\end{figure*}

Figure \ref{fig:SendBias} shows that untrained Speakers have no prior
toward outputting long sequences of symbols. Precisely, from Figure~\ref{fig:SendBias} we see that the untrained Speakers' average
message length coincides with the one produced by the random process defined in Eq.~\ref{eq:pl} where $p=\frac{1}{a}$.\footnote{Note that we did not use the uniqueness-of-messages constraint to define $P_l$.} In other words, untrained
Speakers are equivalent to a random generator with uniform probability over symbols.\footnote{We verified that indeed untrained Speakers have uniform probability over the different symbols.}
Consequently, when imposing message uniqueness, non-trained Speakers
become identical to MT. Hence, Speakers faced with the task of
producing distinct messages for the inputs, if vocabulary size is
not too large, would naturally produce a ZLA-obeying distribution,
that is radically altered in joint Speaker-Listener training.

\subsubsection{Untrained Listener behavior}
\begin{figure*}[ht]
\centering
\includegraphics[width=0.5\textwidth, keepaspectratio]{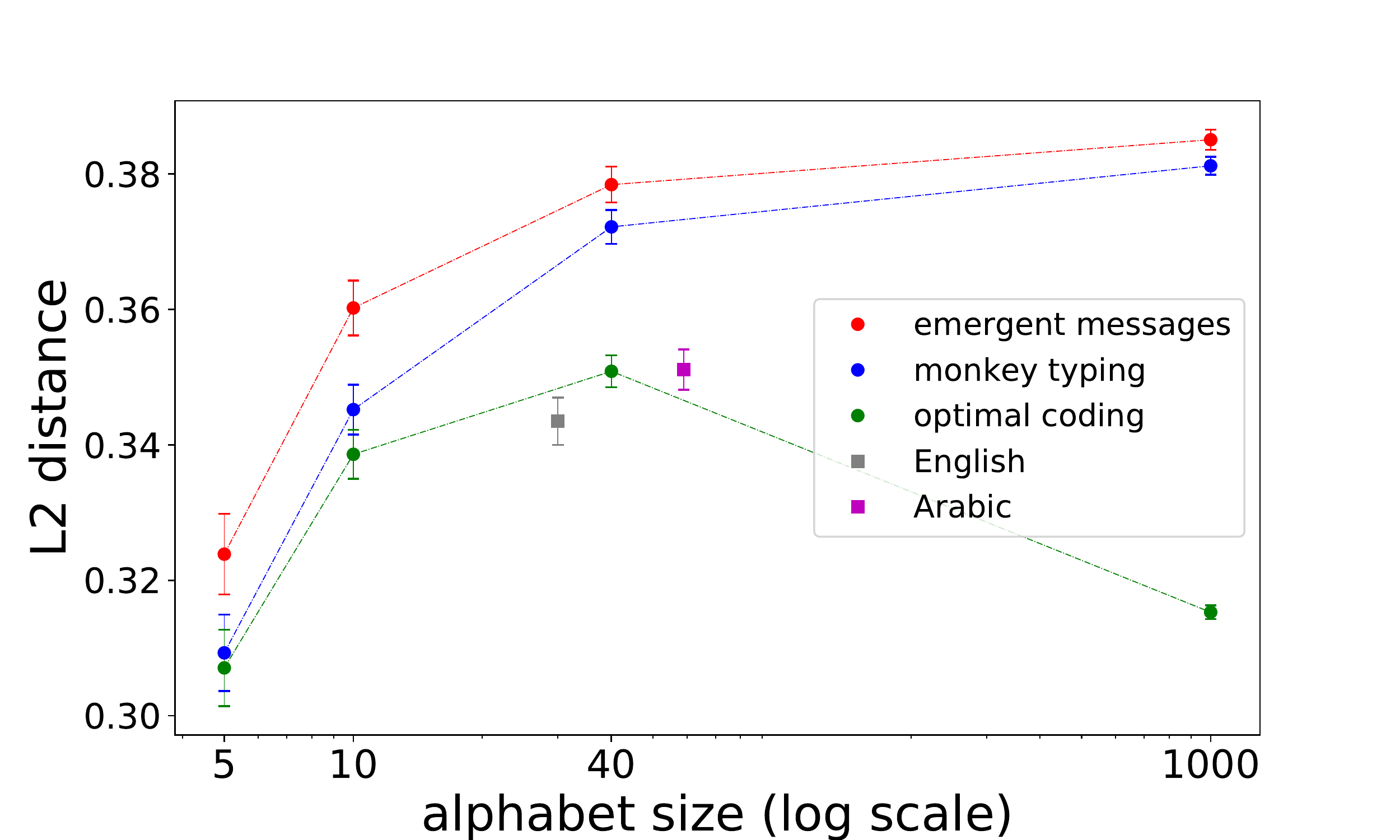} 
\caption{Average pairwise distance between messages' representation in Listener's hidden space, across all considered non-trained Listeners. Vertical lines mark standard deviations across Listeners.}
\label{fig:ListenerBias}
\end{figure*}

Having shown that untrained Speakers do not favor long messages, we ask next if the
emergent anti-efficient language is easier to discriminate by
untrained Listeners than other encodings. To this end, we compute the
average pairwise L2 distance of the hidden representations produced by untrained Listeners in response to messages associated to all inputs.\footnote{Results are similar if looking at the softmax layer instead.}  Messages that are further apart in the
representational space of the untrained Listener should be easier to
discriminate. Thus, if Speaker associates such messages to the inputs, it will be easier for Listener to distinguish them. Specifically, we use $50$ distinct untrained Listeners with 100-dimensional
hidden size.\footnote{We fix this value because, unlike for Speaker, it has considerable impact on performance, with
  $100$ being the preferred setting.} We test $4$ different encodings:
(1) emergent messages (produced by \emph{trained} Speakers) (2) MT
messages ($25$ runs) (3) OC messages and (4) human languages. Note that MT is equivalent
to untrained Speaker, as their messages share the same length
\emph{and} alphabet distribution (see Section
\ref{sec:speakerbias}). We study Listeners' biases
with $\texttt{max_len} = 30$ while varying $a$ as messages are more distinct from reference distributions in that case (see Figure \ref{fig:AllFigs} in Appendix \ref{sec:antieff}). 
Results are reported
in Figure \ref{fig:ListenerBias}. Representations produced in response
to the emergent messages have the highest average distance. MT only
approximates the emergent language for $a=1000$, where, as seen in
Figure \ref{fig:3encodings} above, MT is anti-efficient. The trained Speaker messages are hence \emph{a priori} easier for non-trained Listeners. The length of these  messages could thus be explained by an intrinsic Listener's bias, as conjectured above. Also, interestingly, 
natural languages are not easy to process by Listeners. This suggests that the emergence of ``natural'' languages in LSTM agents is unlikely, without imposing \emph{ad-hoc} pressures.

\subsubsection{Adding a length minimization pressure}

\begin{figure*}[ht]
\centering
{
   \includegraphics[width=0.5\textwidth, keepaspectratio]{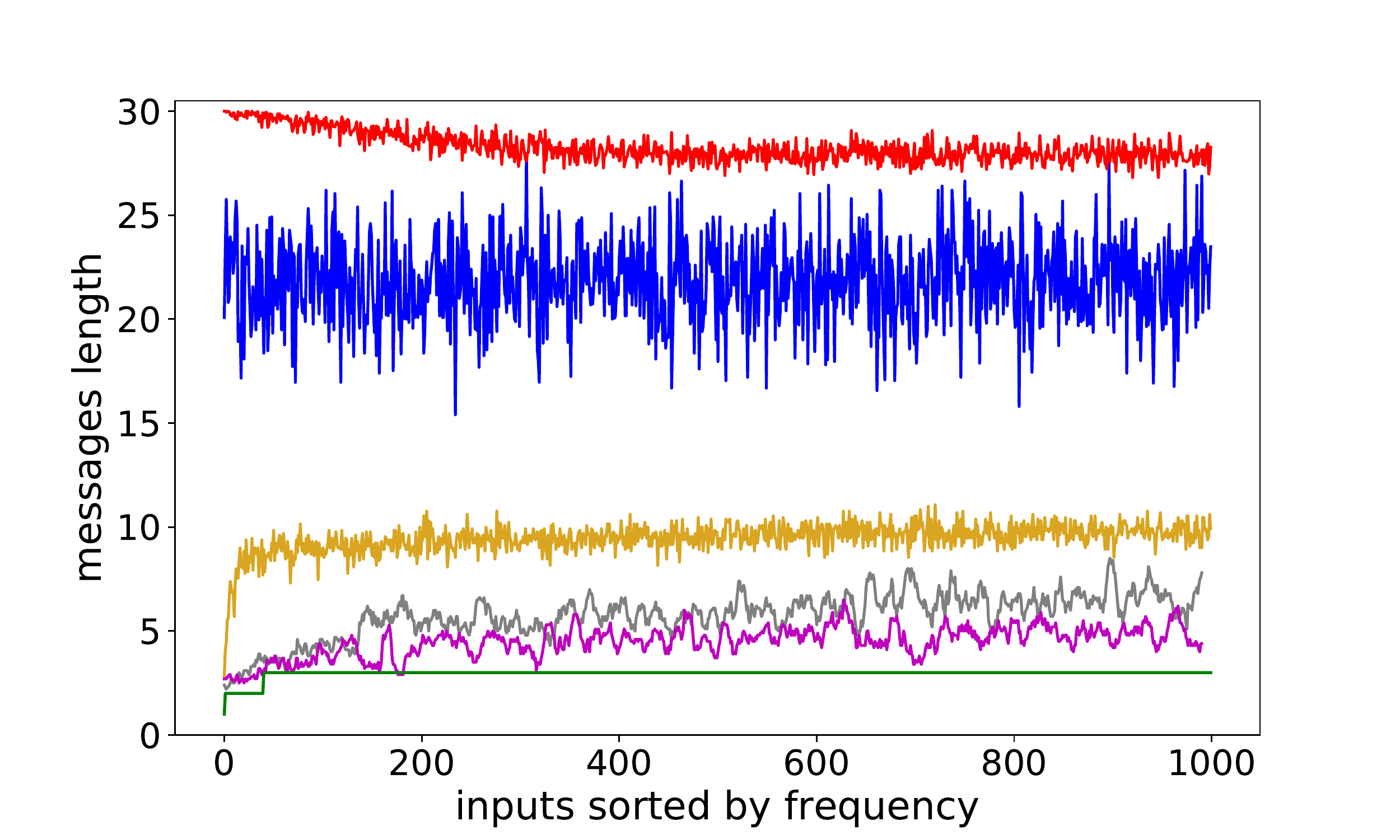} 
}
{
    \includegraphics[width=\textwidth]{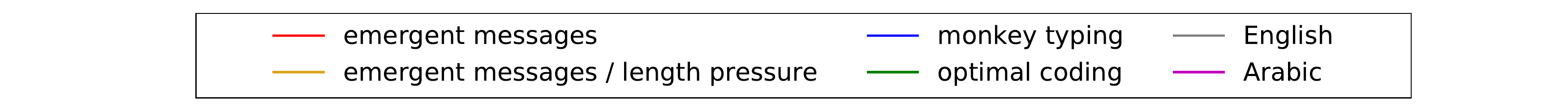}    
}
\caption{Mean length of messages across successful runs as a function of input frequency rank for $\texttt{max_len}=30$, $a=40$, $\alpha=0.5$. Natural language distributions are smoothed as in Fig.~\ref{fig:3encodings}.}\label{fig:Regularization}
\end{figure*}

We next impose an artificial pressure on Speaker to produce short messages, to counterbalance Listener's preference for longer ones. Specifically, we add a regularizer disfavoring longer messages to the original loss:
\begin{equation}
	\label{eq:penalty}
    \mathcal{L}'(i, L(m), m)  = \mathcal{L}(i, L(m)) + \alpha \times |m|
\end{equation}
where $\mathcal{L}(i, L(m))$ is the cross-entropy loss used before,
$|.|$ denotes length, and $\alpha$ is a hyperparameter. The
non-differentiable term $\alpha \times |m|$ is handled seamlessly as
it only depends on Speaker's parameters $\vtheta_s$ (which specify the
distribution of the messages $m$), and the gradient of the loss
w.r.t.\ $\vtheta_s$ is estimated via a REINFORCE-like term
(Eq.~\ref{eq:surrogate}). Figure~\ref{fig:Regularization} shows
emergent message length distribution under this objective, comparing it to
other reference distributions in the most human-language-like setting:
($\texttt{max_len}$=$30$, $a$=$40$). The same pattern is
observed elsewhere (see Appendix \ref{sec:regeffect}, that also
evaluates the impact of the $\alpha$ hyperparameter). The emergent
messages clearly follow ZLA. Speaker now assigns messages of ascending
length to the $40$ most frequent inputs. For the remaining
ones, it chooses messages with relatively similar, but notably
shorter, lengths (always much shorter than MT messages). Still, the
encoding is not as efficient as the one observed in natural language
(and OC). Also, when adding length regularization, we noted a slower convergence, with a smaller number of successful runs, that further diminishes when $\alpha$ increases.

\subsection{Symbol distributions in the emergent code}
\label{sec:symbolDist}
\begin{figure*}[ht]
\centering
\subfigure[Symbol unigram distributions]{
    \includegraphics[width=0.50\textwidth, keepaspectratio]{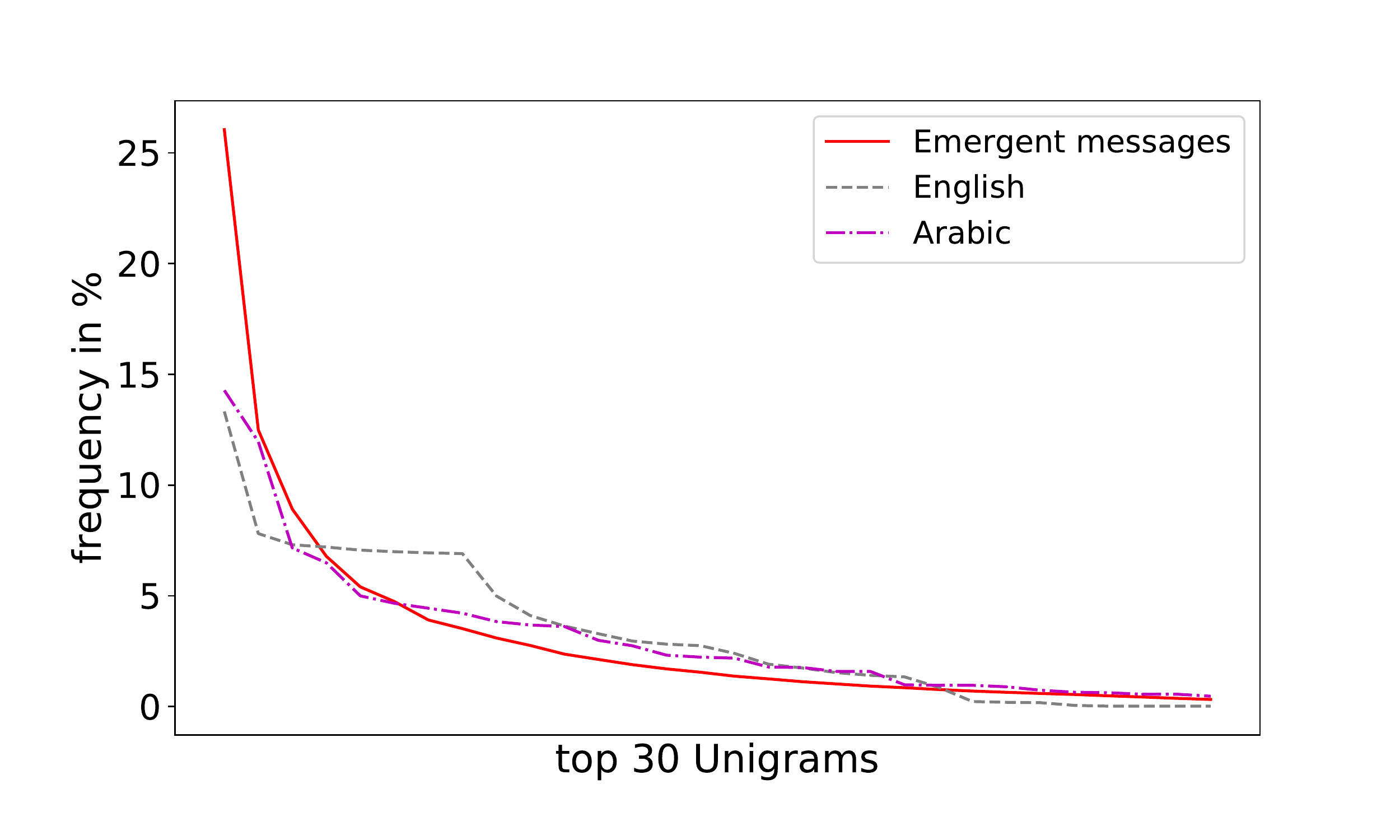} 
    \label{fig:unigramDist}
}
\subfigure[Symbol bigram distributions]{
   \hspace{-1.5\baselineskip}
    \includegraphics[width=0.50\textwidth, keepaspectratio]{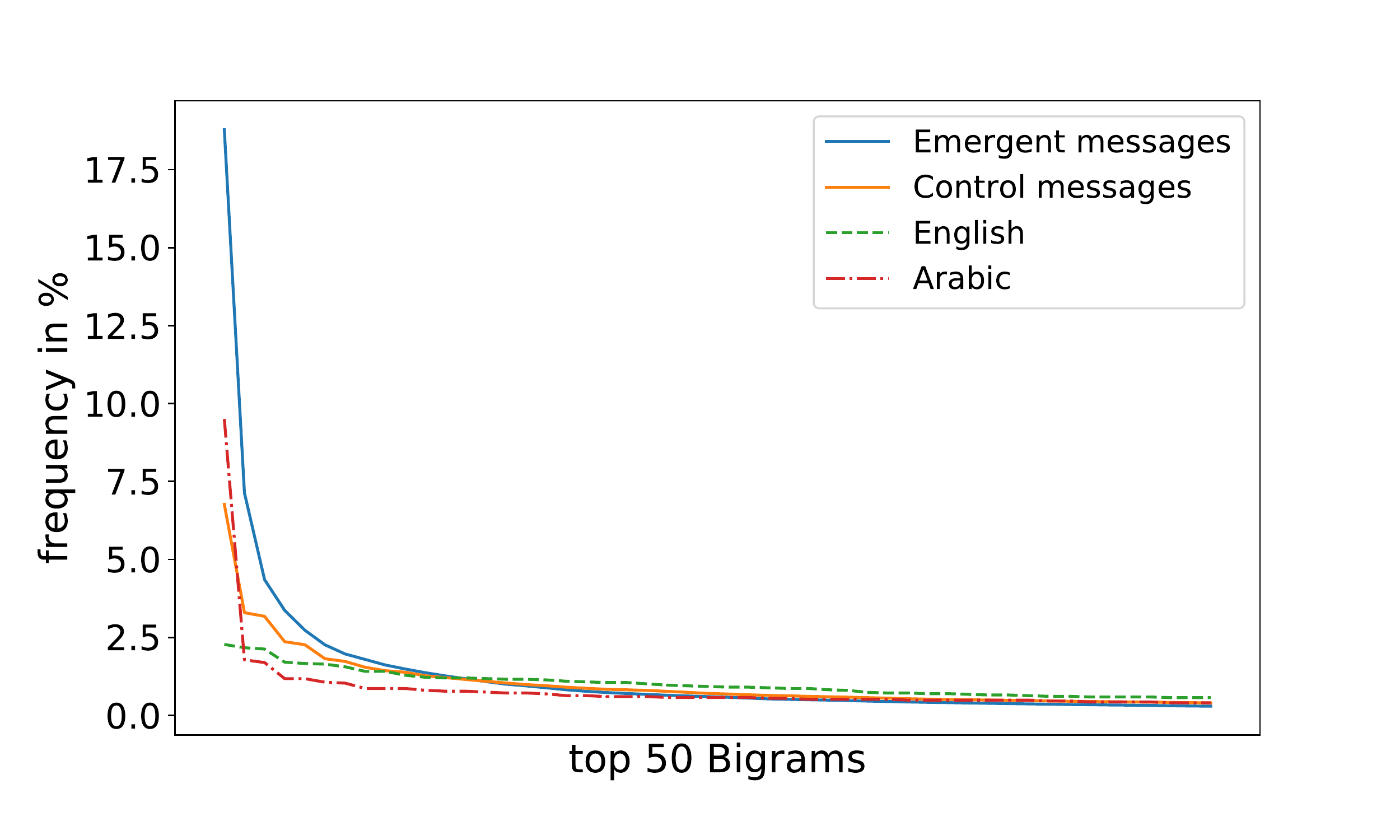} 
    \label{fig:bigramDist}
}
\caption{Distribution of top symbol unigrams and bigrams (ordered by frequency) in different codes. Emergent and control messages are averaged across successful runs and different simulations respectively in the ($\texttt{max_len}$=$30$,$a$=$40$) setting.}\label{fig:vocabUse}
\end{figure*}

We conclude with a high-level look at what the long emergent messages
are made of. Specifically, we inspect symbol unigram and bigram
frequency distributions in the messages produced by trained Sender in
response to the $1K$ inputs (the $\texttt{eos}$ symbol is excluded from counts). For direct comparability with
natural language, we report results in the
($\texttt{max_len}$=$30$,$a$=$40$) setting, but the patterns
are general. We observe in Figure \ref{fig:unigramDist} that, even if
at initialization Speaker starts with a uniform distribution over its
alphabet (not shown here), by end of training it has converged to
a very skewed one. Natural languages follow a similar trend, but their
distributions are not nearly as skewed (see Figure \ref{fig:unigramEnt} in Appendix \ref{sec:entropy} for entropy analysis). We then investigate message
structure by looking at symbol bigram distribution. To this end,
we build $25$ randomly generated \emph{control codes}, constrained to have the same mean length and unigram symbol
distribution as the emergent
code. 
Intriguingly, we observe in Figure \ref{fig:bigramDist} a significantly more
skewed emergent bigram distribution, compared to the controls. This suggests that, despite the
lack of phonetic pressures, Speaker is respecting
``phonotactic'' constraints that are even sharper than those reflected
in the natural language bigram distributions (see Figure \ref{fig:bigramEnt} in Appendix \ref{sec:entropy} for entropy analysis). In other words, the
emergent messages are clearly not built out of random unigram
combinations. Looking at the pattern more closely, we find the skewed
bigram distribution to be due to a strong tendency to repeat the same
character over and over, well beyond what is expected given the unigram
symbol skew (see typical message examples in Appendix \ref{sec:examples}).
More quantitatively, across all runs with $\texttt{max_len}$=$30$, if
we denote the $10$ most probable symbols with $s_1,...,s_{10}$, then
we observe $P(s_r,s_r)>P(s_r)^{2}$ with
$r \in \llbracket 1,.., 10 \rrbracket$, in more than $97.5\%$ runs. We
leave a better understanding of the causes and implications of these
distributions to future work.

\section{Discussion}

We found that two neural networks faced with a simple communication
task, in which they have to learn to generate messages to refer to a
set of distinct inputs that are sampled according to a power-law
distribution, produce an \emph{anti-efficient} code where more
frequent inputs are significantly associated to longer messages, and
all messages are close to the allowed maximum length threshold. The
results are stable across network and task hyperparameters (although
we leave it to further work to replicate the finding with different
network architectures, such as transformers or CNNs). Follow-up
experiments suggest that the emergent pattern stems from an \emph{a
  priori} preference of the listener network for longer, more
discriminable messages, which is not counterbalanced by a need to
minimize articulatory effort on the side of the speaker. Indeed, when
an artificial penalty against longer messages is imposed on the
latter, we see a ZLA distribution emerging in the networks' communication code.

From the point of view of AI, our results stress the importance of
controlled analyses of language emergence. Specifically, if we want to
develop artificial agents that naturally communicate with humans, we want to ensure
that we are aware of, and counteract, their unnatural biases, such as the
one we uncovered here in favor of anti-efficient encoding. We presented
 a proof-of-concept example of how to get rid of this specific
bias by directly penalizing long messages in the cost function, but
future work should look into less \emph{ad hoc} ways to condition the
networks' language. Getting the encoding right seems particularly
important, as efficient encoding has been observed to interact in subtle ways
with other important properties of human language, such as regularity
and compositionality \citep{kirby2001spontaneous}. We also emphasize
the importance of using power-law input distributions when studying
language emergence, as the latter are a universal property of human
language \citep{Zipf:1949,Baayen:2001} largely ignored in previous
simulations, that assume uniform input distributions.

ZLA is observed in all studied human languages. As mentioned above,
some animal communication systems violate it \citep{heesen2019}, but such systems are 1)
limited in their expressivity; and 2) do not display a significantly
\emph{anti-}efficient pattern. We complemented this earlier comparative
research with an investigation of emergent language among artificial
agents that need to signal a large number of different inputs.  We
found that the agents develop a successful communication system that
does \emph{not} exhibit ZLA, and is actually significantly anti-efficient. We
connected this to an asymmetry in speaker vs.~listener biases. This in turn suggests that ZLA in
communication in general does not emerge from trivial statistical
properties, but from a delicate balance of speaker and listener
pressures. Future work should investigate emergent distributions in a
wider range of artificial agents and environments, trying to understand which factors
are determining them.

\section{Acknowledgments}
We would like to thank Fermín Moscoso del Prado Martín, Ramon Ferrer i Cancho, Serge Sharoff, the audience at REPL4NLP 2019 and the anonymous reviewers for helpful comments and suggestions.

\bibliography{marco,other}
\bibliographystyle{unsrtnat}

\setcounter{section}{0}
\setcounter{figure}{0}
\renewcommand\thesection{A.\arabic{section}}
\renewcommand\thetable{A.\arabic{table}}
\renewcommand{\thefigure}{A\arabic{figure}}

\section{Supplementary}

\subsection{Hyperparameters}
\label{sec:hyperparams}

Both speaker and listener agents are single-layer LSTMs \citep{Hochreiter:Schmidhuber:1997}. We experiment with the combinations (Speaker's hidden size, Listener's hidden size) in $[(100, 100), (250, 100), (250, 250), (500,250)]$. We only experiment  with  combinations where Speaker's hidden-size is bigger or equal to Listener's,  because of the asymmetry in their tasks. Indeed, as discussed in Section \ref{sec:emergentenc} of the main paper, the Speaker's search space $M_a^{\texttt{max_len}}$ is generally larger than the one of the Listener $R$. 

We use the Adam optimizer, with learning rate $0.001$. We apply entropy regularization to Speaker's optimization. The values of the regularization's coefficient are chosen in $[1, 1.5, 2]$. We run the simulation with each hyperparameter setting $4$ times with different random seeds.

\subsection{Monkey typing}
\label{sec:Intersilence}
\begin{figure*}[ht]
\centering
\subfigure[$\texttt{max_len}$=$6$, $a$=$5$]{
    \includegraphics[width=0.33\textwidth, keepaspectratio]{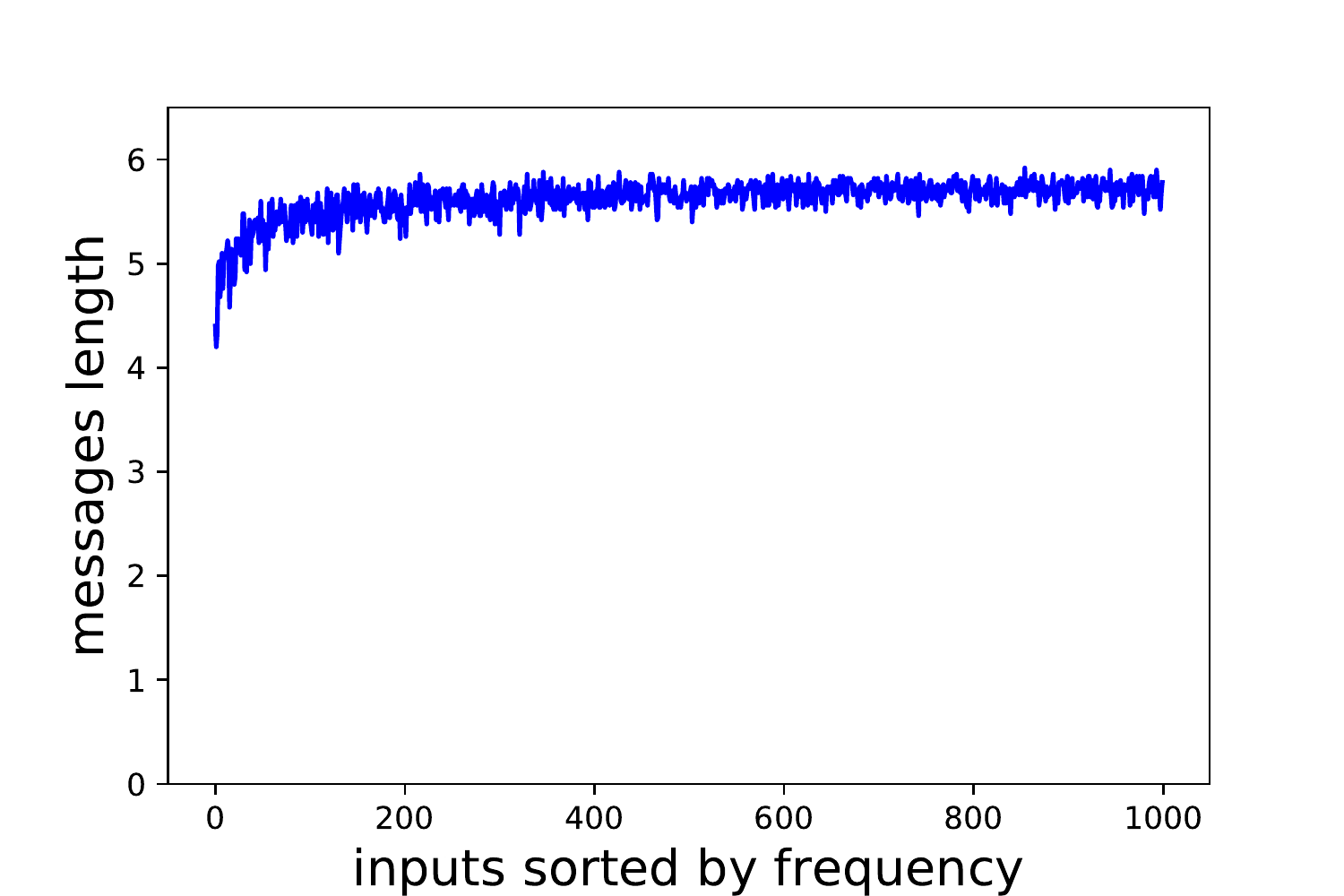}    
}
\subfigure[\hspace{-.3\baselineskip}$\texttt{max_len}$=$6$, $a$=$10$]{
    \hspace{-1.5\baselineskip}
    \includegraphics[width=0.33\textwidth, keepaspectratio]{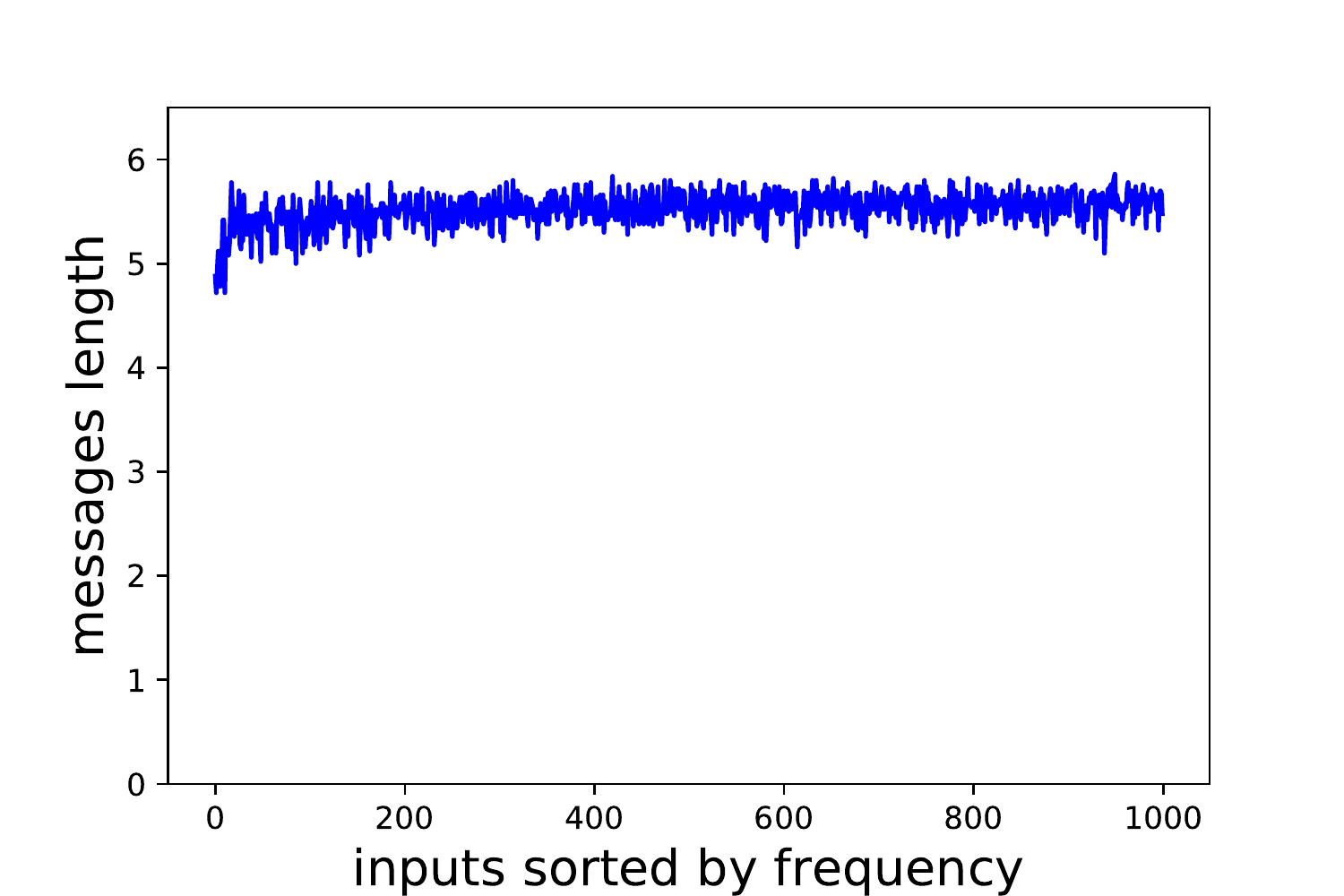}    
}
\subfigure[\hspace{-.3\baselineskip}$\texttt{max_len}$=$6$, $a$=$40$]{
    \hspace{-1.5\baselineskip}
    \includegraphics[width=0.33\textwidth, keepaspectratio]{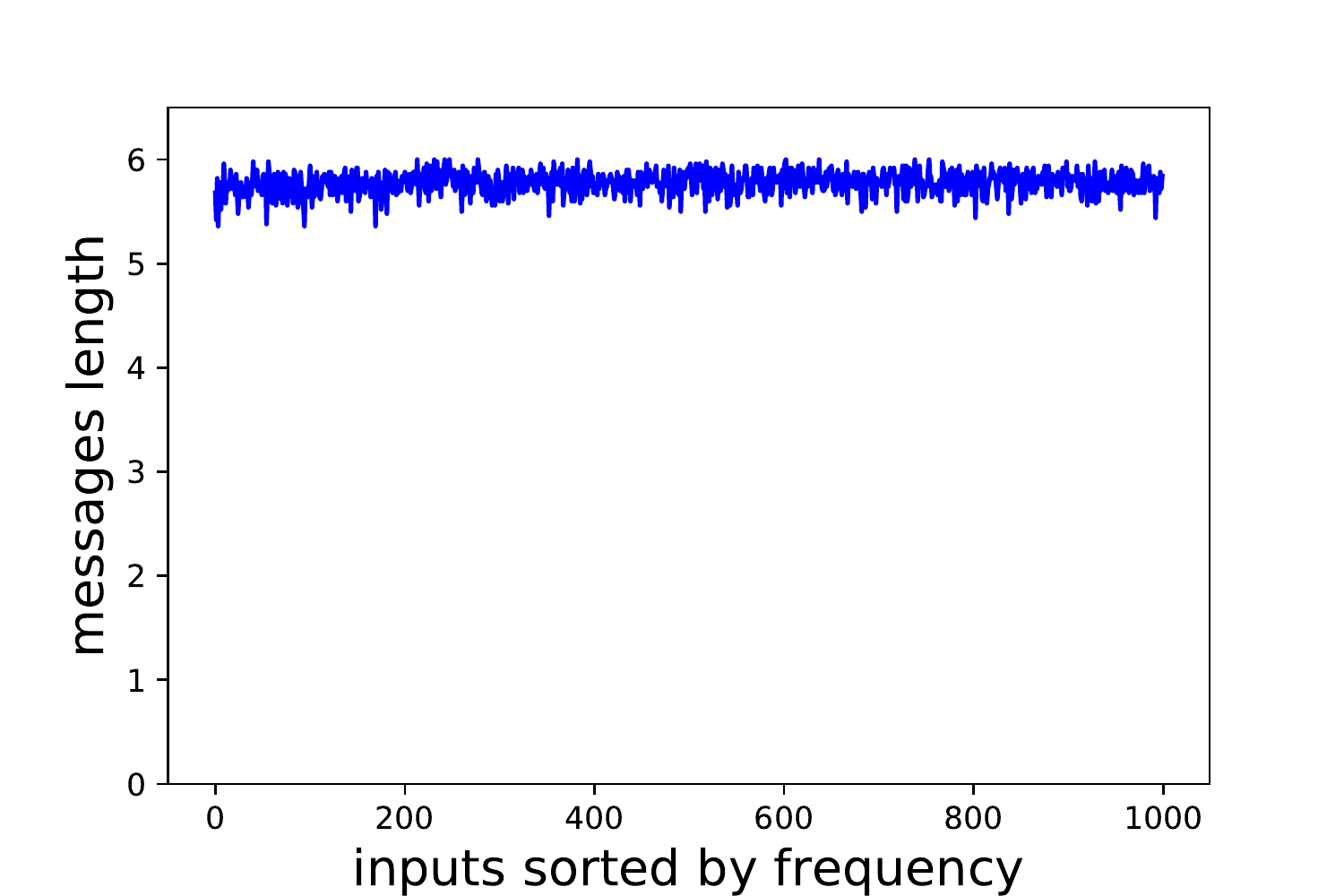}    
}
\subfigure[$\texttt{max_len}$=$11$, $a$=$3$]{

    \includegraphics[width=0.25\textwidth, keepaspectratio]{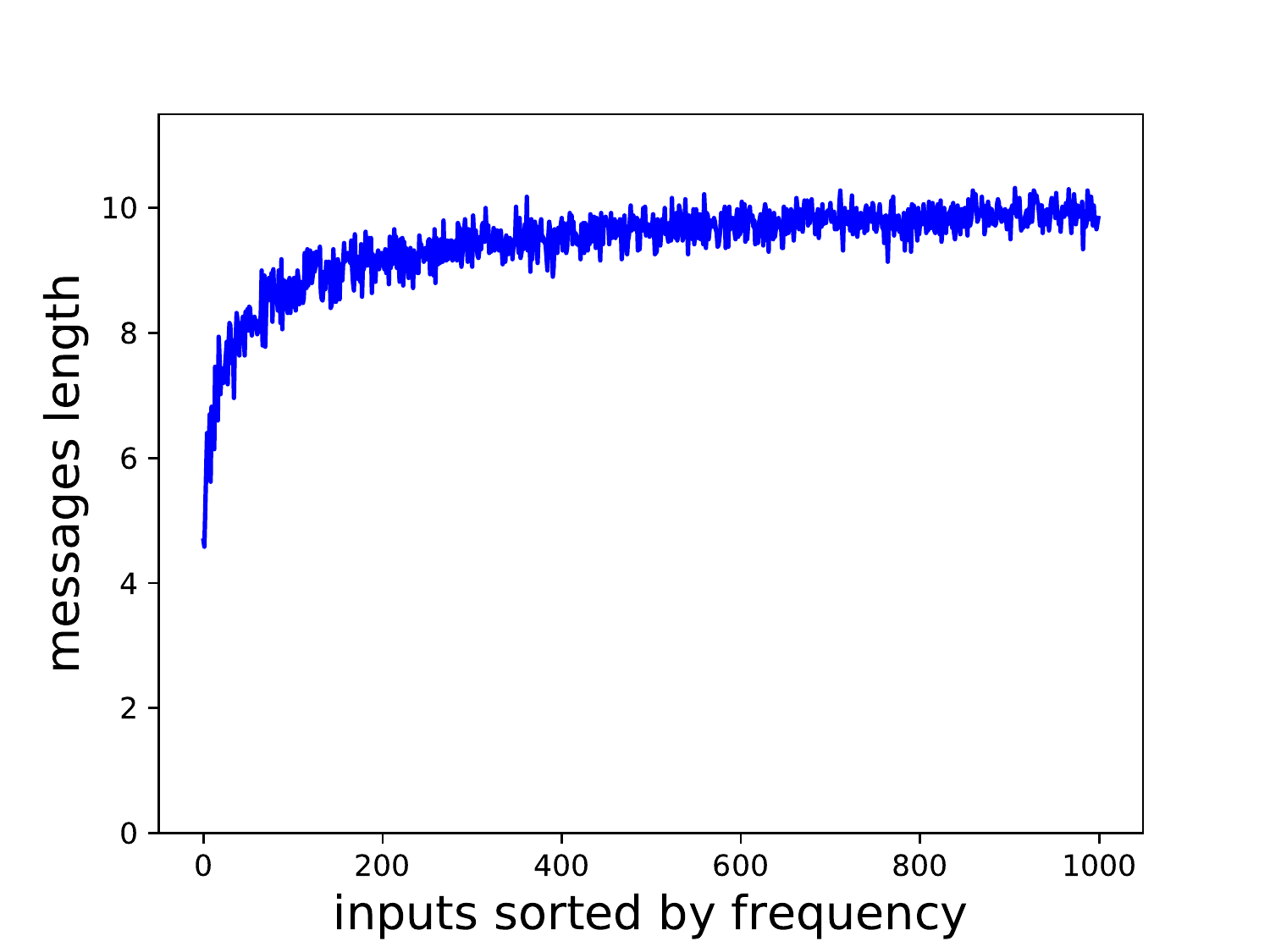}    
}
\subfigure[\hspace{-.2\baselineskip}$\texttt{max_len}$=$11$, $a$=$5$]{
    \hspace{-1.5\baselineskip}
    \includegraphics[width=0.25\textwidth, keepaspectratio]{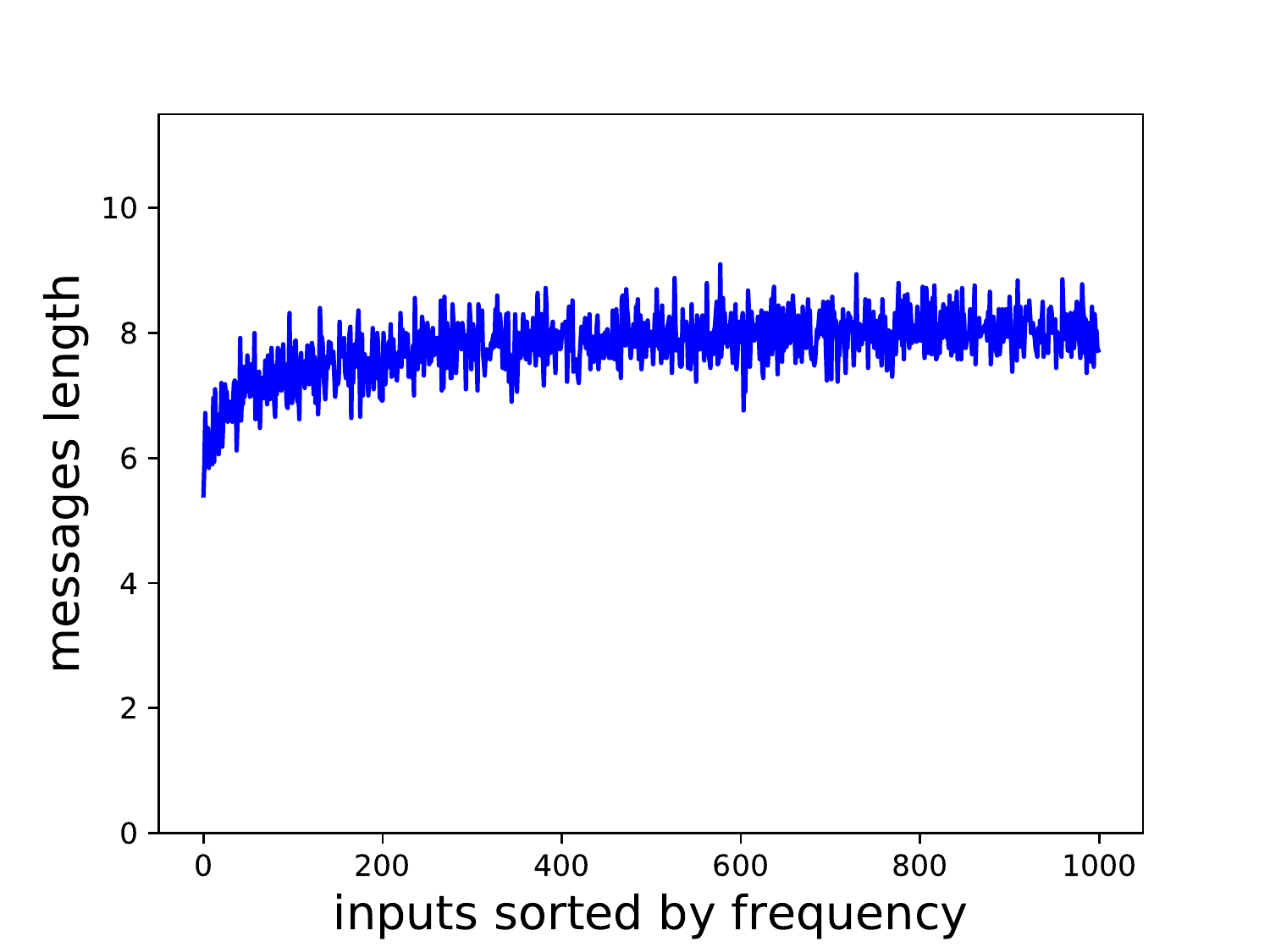}    
}
\subfigure[\hspace{-.3\baselineskip}$\texttt{max_len}$=$11$, $a$=$10$]{
    \hspace{-1.5\baselineskip}
    \includegraphics[width=0.25\textwidth, keepaspectratio]{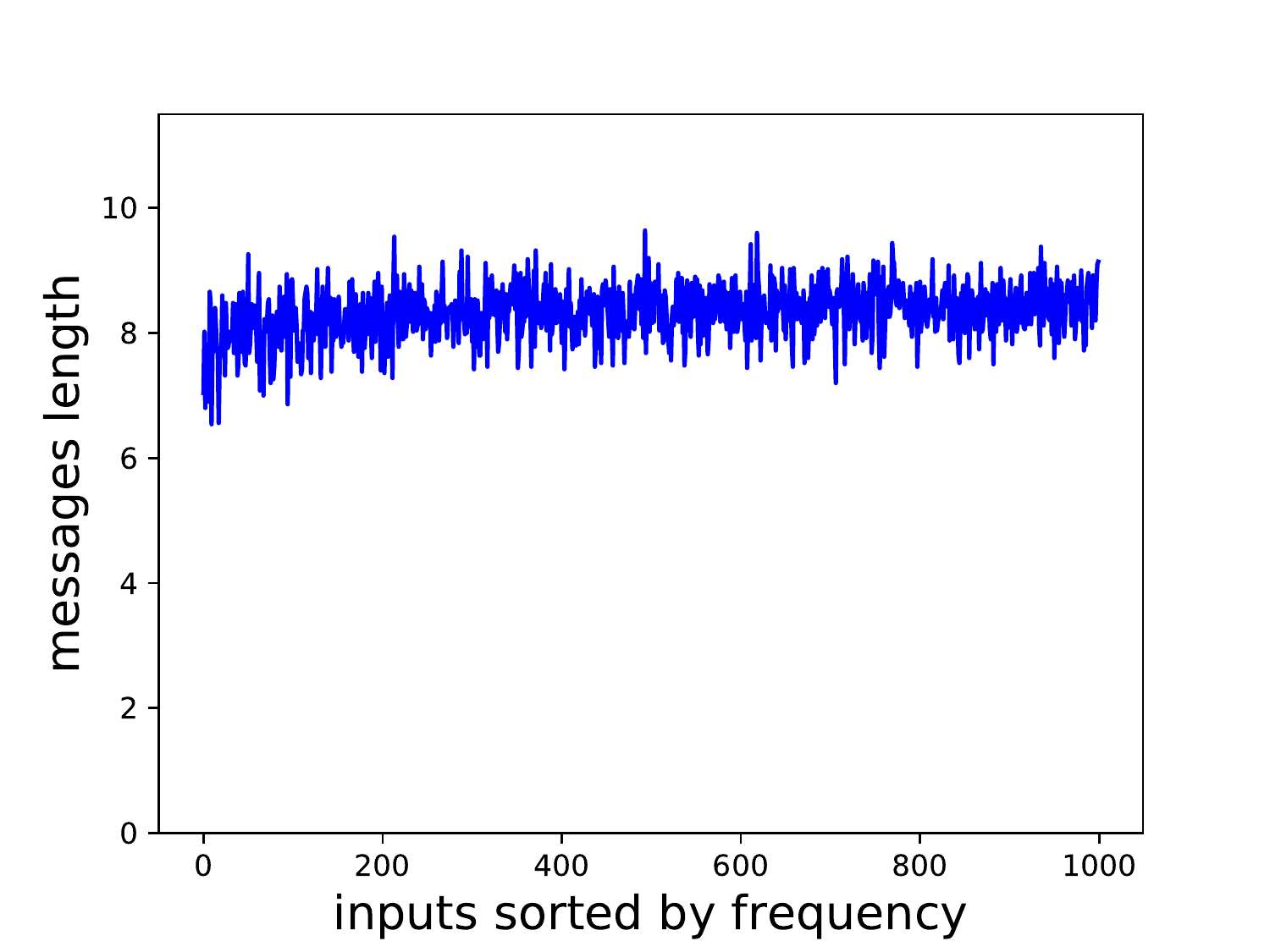}    
}
\subfigure[\hspace{-.3\baselineskip}$\texttt{max_len}$=$11$, $a$=$40$]{
    \hspace{-1.5\baselineskip}
    \includegraphics[width=0.25\textwidth, keepaspectratio]{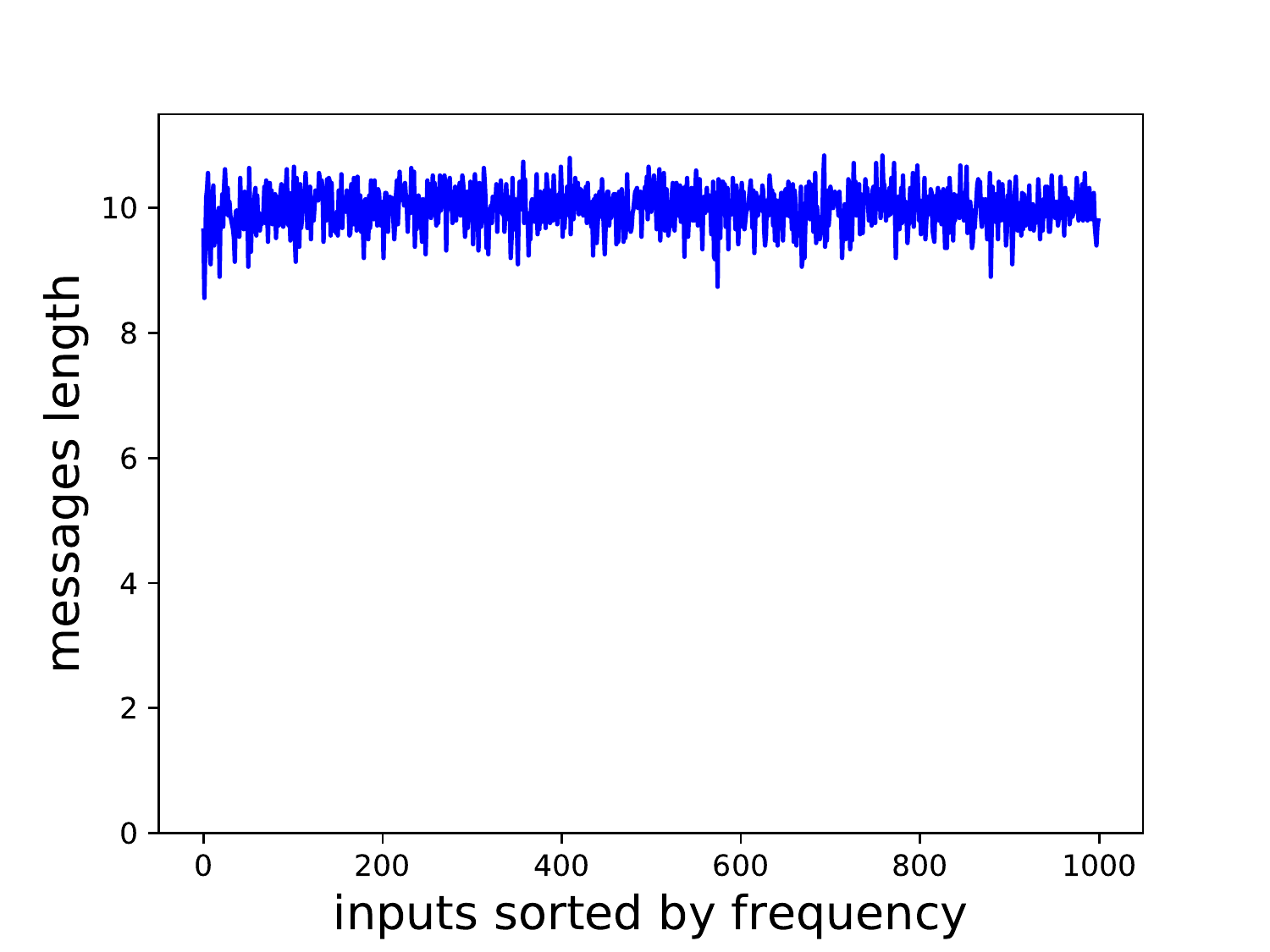}    
}
\subfigure[\hspace{-.3\baselineskip}$\texttt{max_len}$=$30$, $a$=$3$]{
    \includegraphics[width=0.25\textwidth, keepaspectratio]{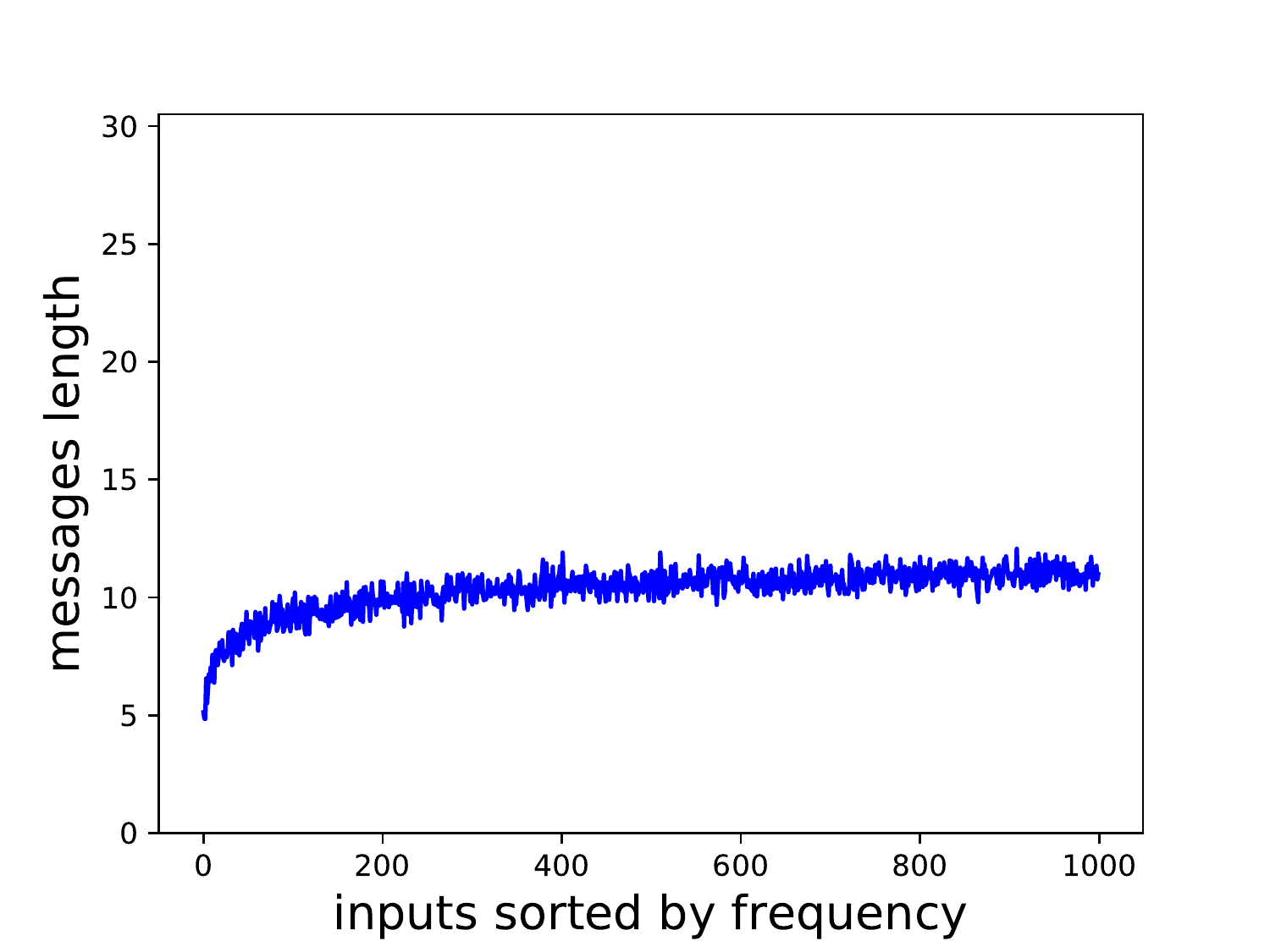}    
}
\subfigure[\hspace{-.4\baselineskip}$\texttt{max_len}$=$30$, $a$=$5$]{
    \hspace{-1.5\baselineskip}
    \includegraphics[width=0.25\textwidth, keepaspectratio]{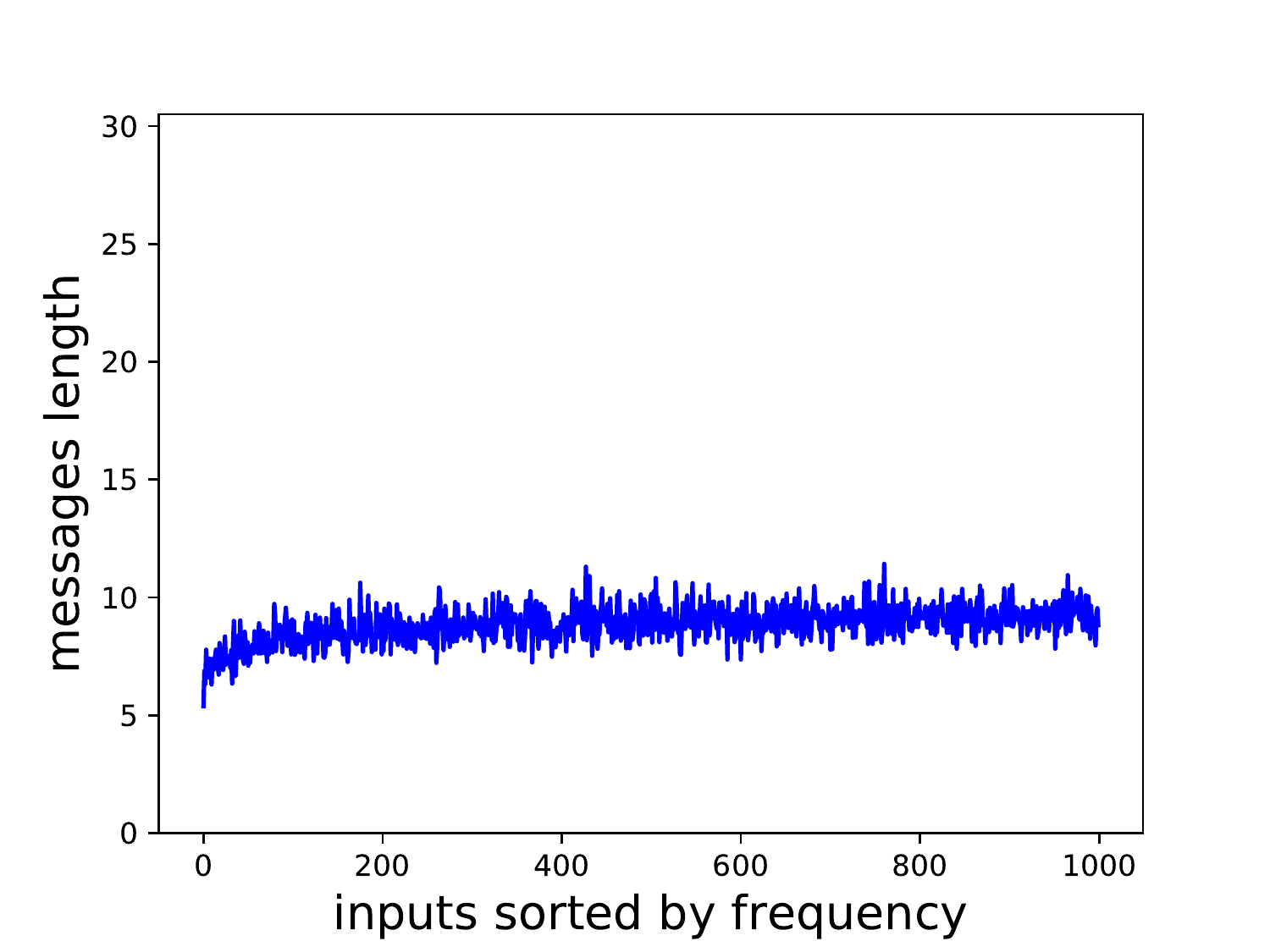}    
}
\subfigure[\hspace{-0.3\baselineskip}$\texttt{max_len}$=$30$, $a$=$10$]{
    \hspace{-1.5\baselineskip}
    \includegraphics[width=0.25\textwidth, keepaspectratio]{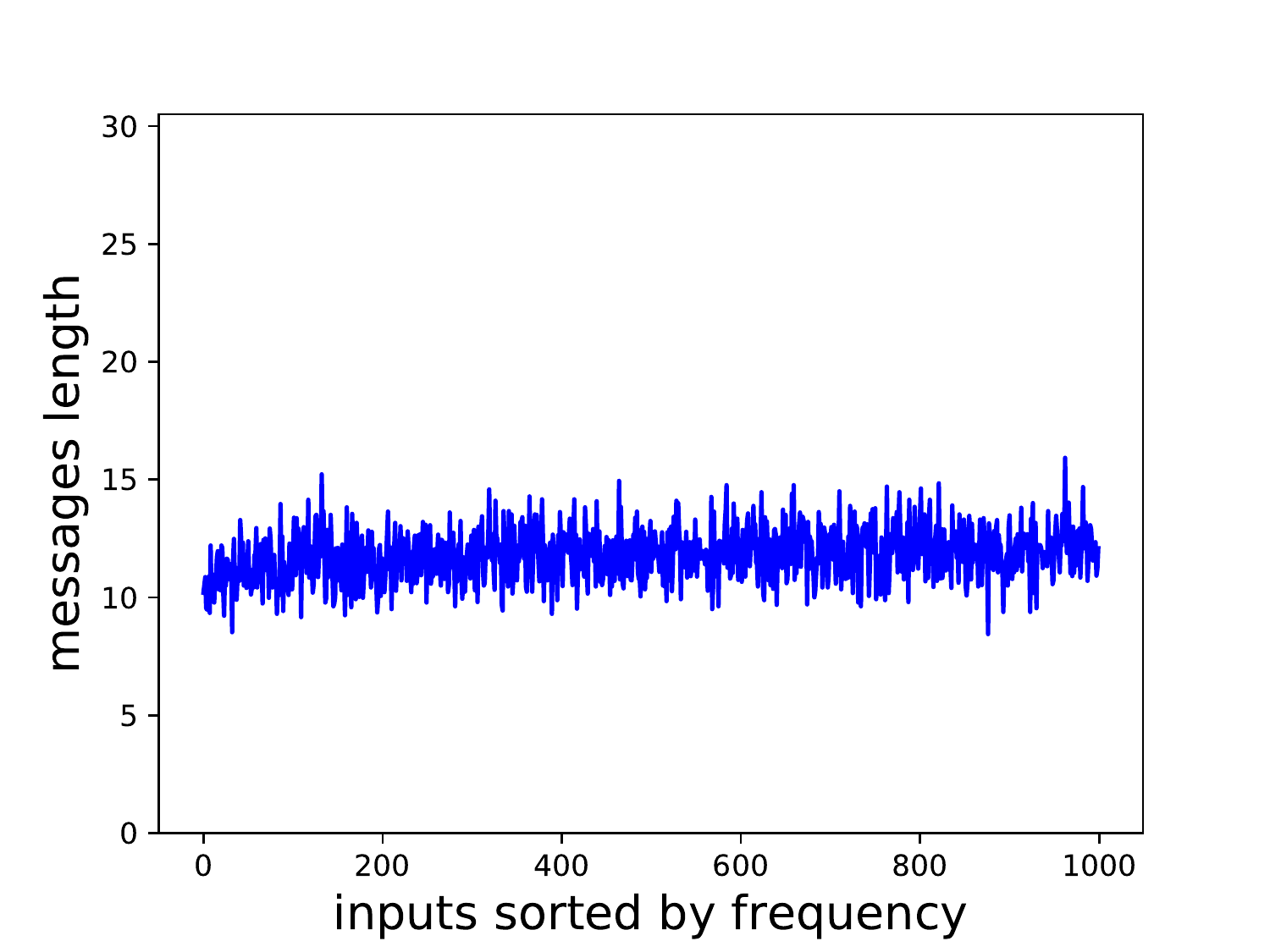} 
}
\subfigure[\hspace{-0.3\baselineskip}$\texttt{max_len}$=$30$, $a$=$40$]{
    \hspace{-1.5\baselineskip}
    \includegraphics[width=0.25\textwidth, keepaspectratio]{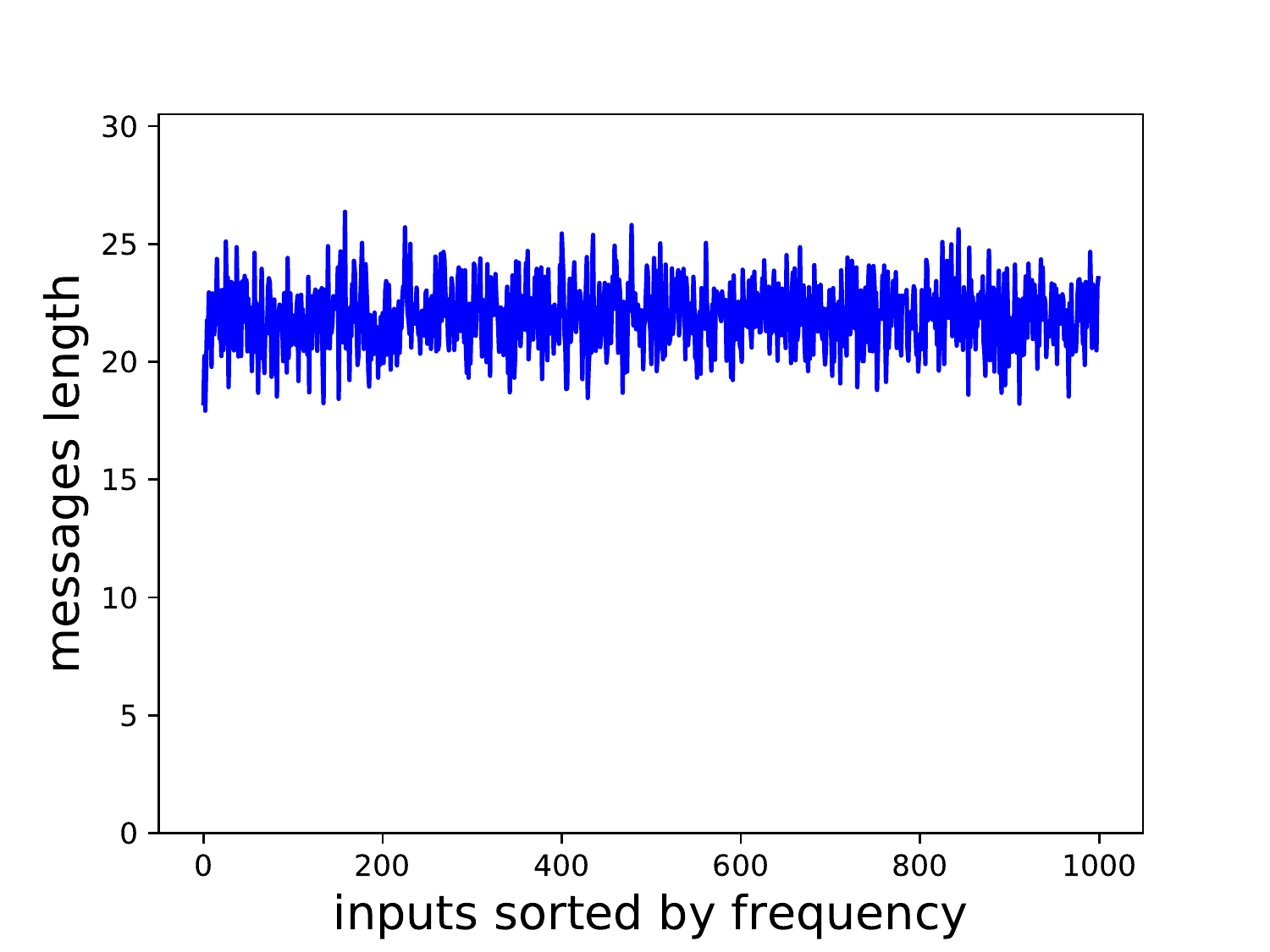} 
}
\vspace{-.5\baselineskip}
\caption{Monkey typing encoding: Mean message length across 50 simulations as a function of input frequency rank.}\label{fig:intermittentSilence}
\end{figure*}

We adapt the Monkey typing (MT) process by adding the $\texttt{max_len}$ constraint. This makes it a ZLA-like distribution only when vocabulary size $a$ is small. Figure \ref{fig:intermittentSilence} illustrates this behavior. We see that the higher $a$ is, the further the MT distribution departs from a ZLA pattern.
 
\subsection{Natural language distributions}
\label{sec:NatLang}
\begin{figure*}[ht]
\centering
\includegraphics[width=0.5\textwidth, keepaspectratio]{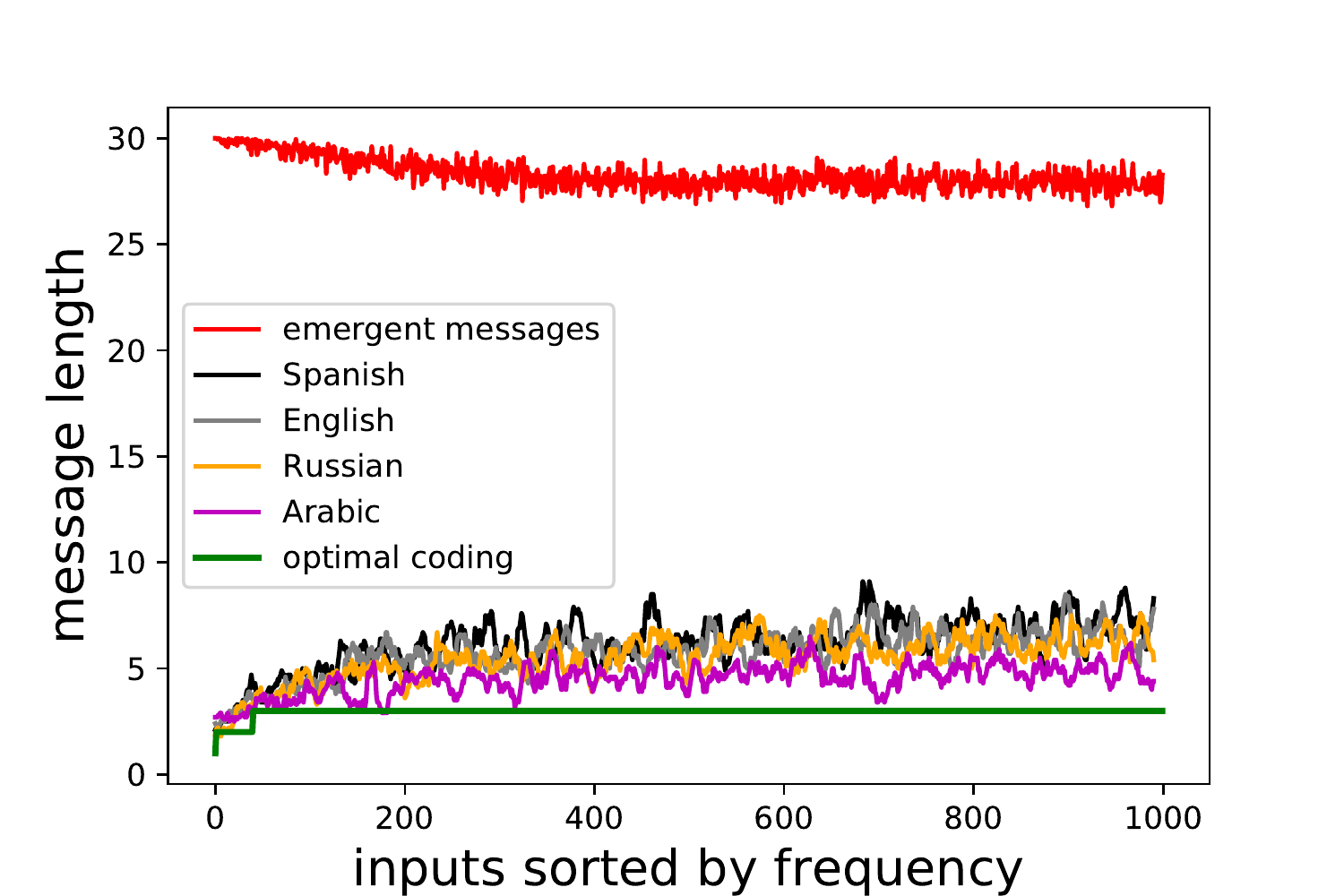} 
\caption{Word length in natural languages in function of word frequency rank, compared to average emergent code and OC in the $(\texttt{max_len}=30, a=40)$ setting. For readability, we smooth natural language distributions by reporting the sliding average of $10$ consecutive lengths.}
\label{fig:natlang}
\end{figure*}

We report in Figure \ref{fig:natlang} word length distributions for all the natural languages we considered, and compare them with (1) optimal encoding (OC) and (2) emergent language in the most comparable simulation setting: $(\texttt{max_len}=30, a=40)$. Despite their different alphabet sizes, natural languages pattern similarly: They follow ZLA, and approximate OC.

\subsection{Anti-efficient emergent language}
\label{sec:antieff}
\begin{figure*}[ht]
\centering
\subfigure[$\texttt{max_len}$=$6$, $a$=$40$]{
    \hspace{-1.5\baselineskip}
    \includegraphics[width=0.50\textwidth, keepaspectratio]{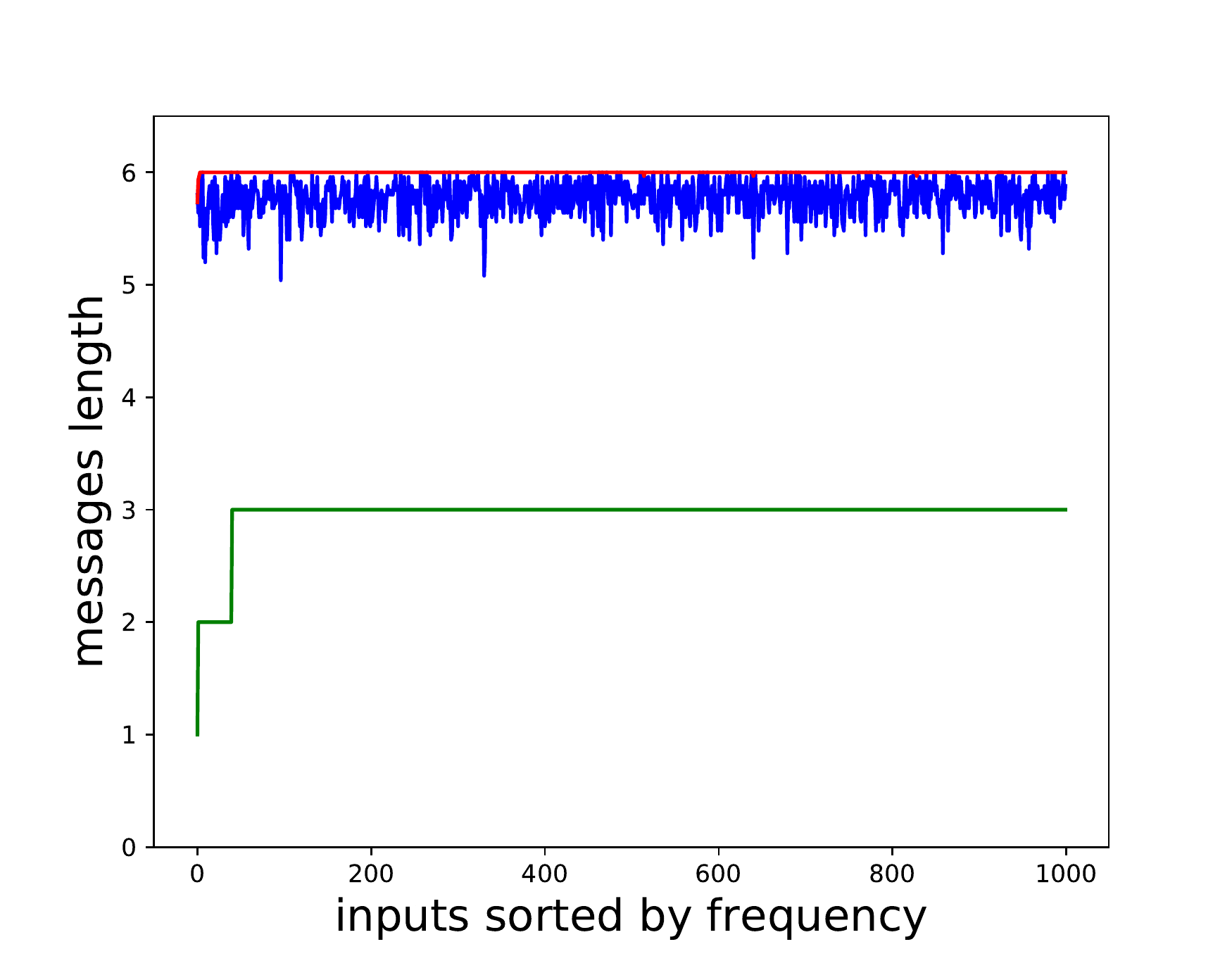} 
}
\subfigure[$\texttt{max_len}$=$6$, $a$=$1000$]{
    \hspace{-1.5\baselineskip}	
    \includegraphics[width=0.50\textwidth, keepaspectratio]{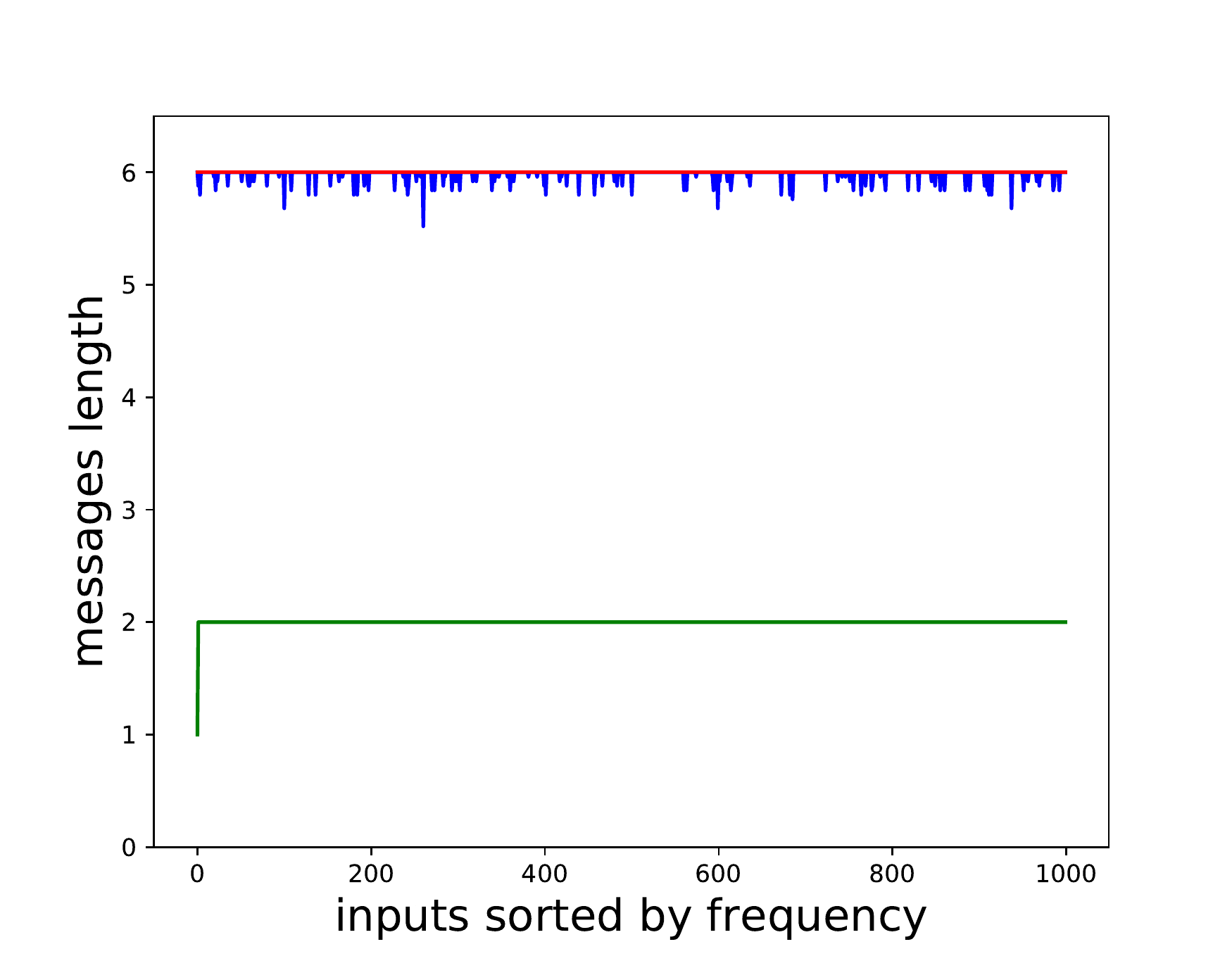}
}
\subfigure[$\texttt{max_len}$=$11$, $a$=$10$]{
    \hspace{-1.5\baselineskip}
    \includegraphics[width=0.33\textwidth, keepaspectratio]{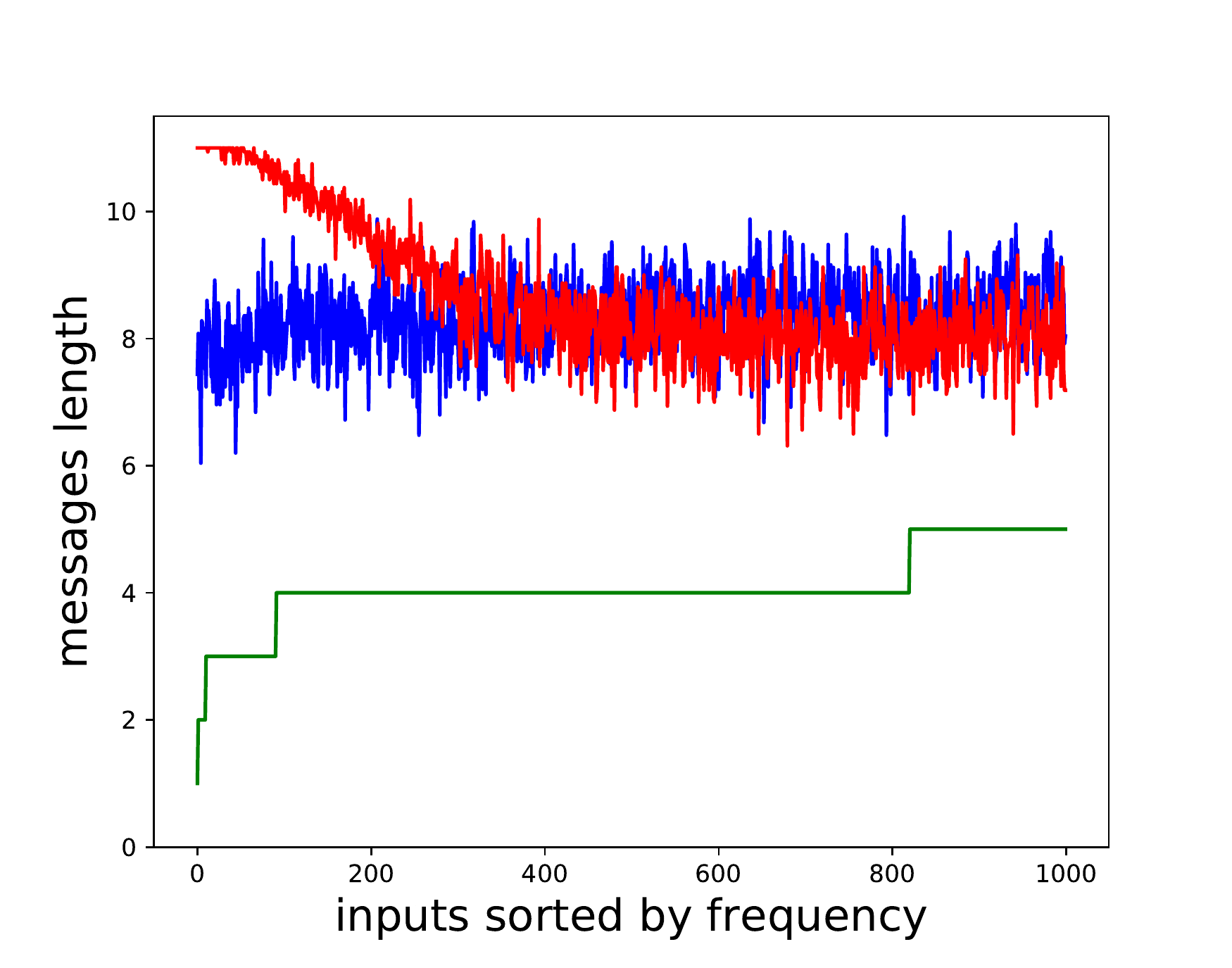}    
}
\subfigure[$\texttt{max_len}$=$11$, $a$=$40$]{
    \hspace{-1.5\baselineskip}
    \includegraphics[width=0.33\textwidth, keepaspectratio]{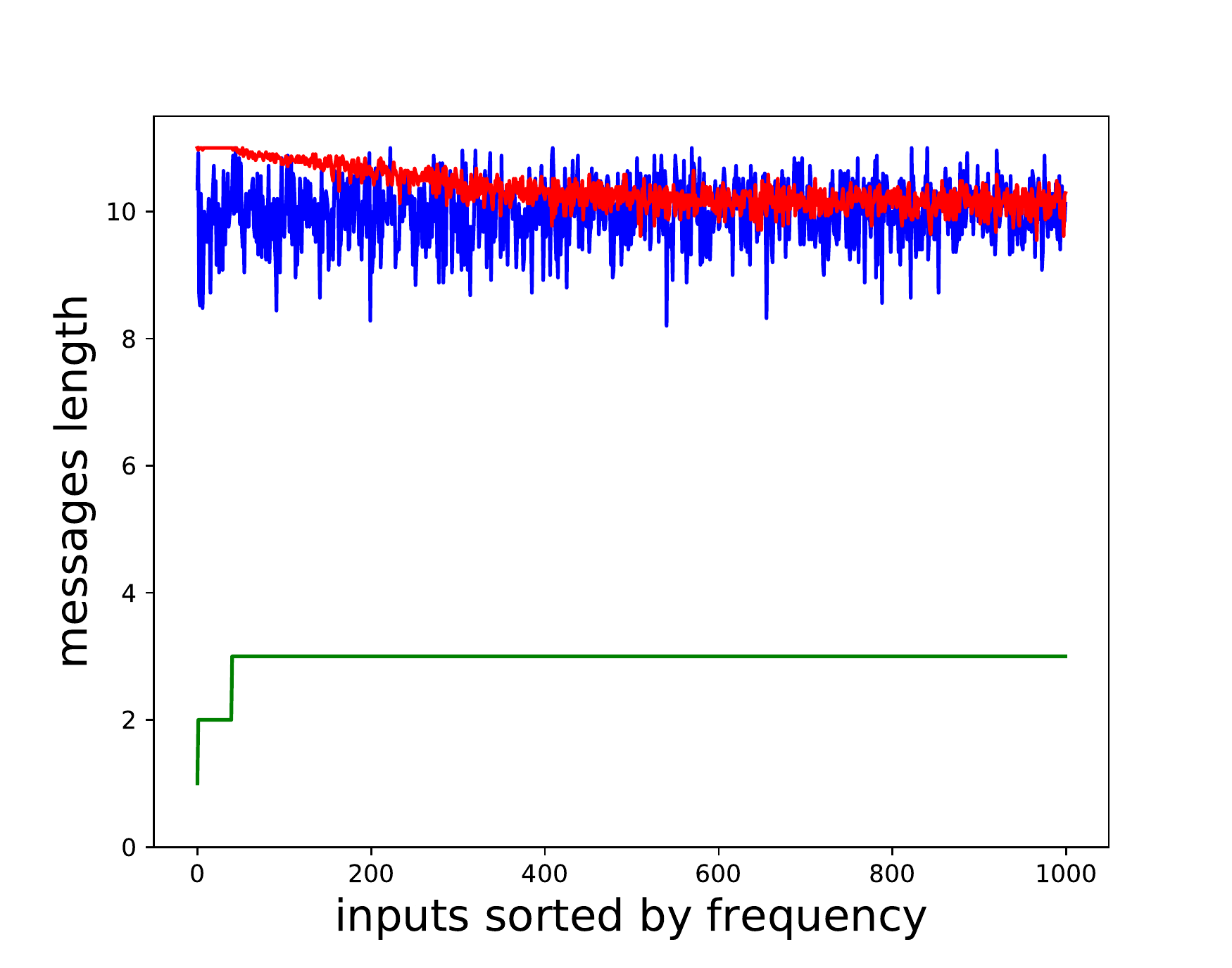}    
}
\subfigure[\hspace{-.2\baselineskip}$\texttt{max_len}$=$11$, $a$=$1000$]{
    \hspace{-1.5\baselineskip}
    \includegraphics[width=0.33\textwidth, keepaspectratio]{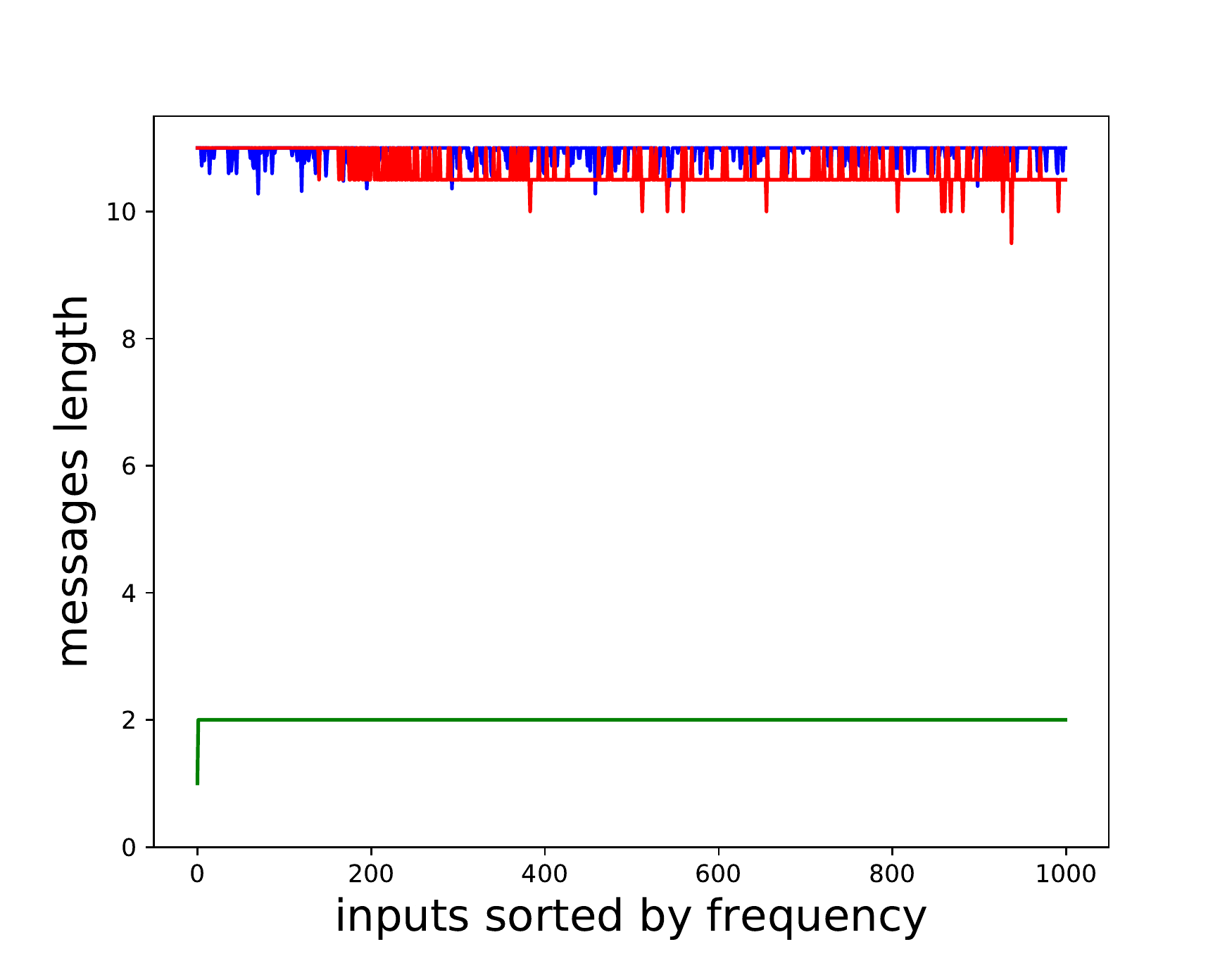}    
}
\subfigure[\hspace{-.3\baselineskip}$\texttt{max_len}$=$30$, $a$=$5$]{
    \hspace{-1.5\baselineskip}
    \includegraphics[width=0.25\textwidth, keepaspectratio]{figs/length3encsmallML30VS5.pdf}    
}
\subfigure[\hspace{-.3\baselineskip}$\texttt{max_len}$=$30$, $a$=$10$]{
    \hspace{-1.5\baselineskip}
    \includegraphics[width=0.25\textwidth, keepaspectratio]{figs/length3encsmallML30VS10.pdf}    
}
\subfigure[\hspace{-.4\baselineskip}$\texttt{max_len}$=$30$, $a$=$40$]{
    \hspace{-1.5\baselineskip}
    \includegraphics[width=0.25\textwidth, keepaspectratio]{figs/length3encsmallML30VS40.pdf}    
}
\subfigure[\hspace{-0.3\baselineskip}$\texttt{max_len}$=$30$, $a$=$1000$]{
    \hspace{-1.5\baselineskip}
    \includegraphics[width=0.25\textwidth, keepaspectratio]{figs/length3encsmallML30VS1000.pdf} 
}
{
    \includegraphics[width=\textwidth]{figs/length3encsmallML.pdf}    
}
\vspace{-.5\baselineskip}
\caption{Mean message length across successful runs as a function of input frequency rank, with reference distributions. Natural language distributions are smoothed as in Fig. \ref{fig:natlang}.}\label{fig:AllFigs}
\end{figure*}

Figure \ref{fig:AllFigs} shows message length distribution (averaged across all successful runs) in function of input frequency rank, and compares it with some reference distributions. The results are in line with our finding in Section \ref{sec:emergentenc} of the main paper.

\subsection{Emergent language with uniform input distribution}
\label{sec:unifrom}
\begin{figure*}[ht]
\centering
\subfigure[$a$=$3$]{
    \hspace{-1.5\baselineskip}
    \includegraphics[width=0.25\textwidth, keepaspectratio]{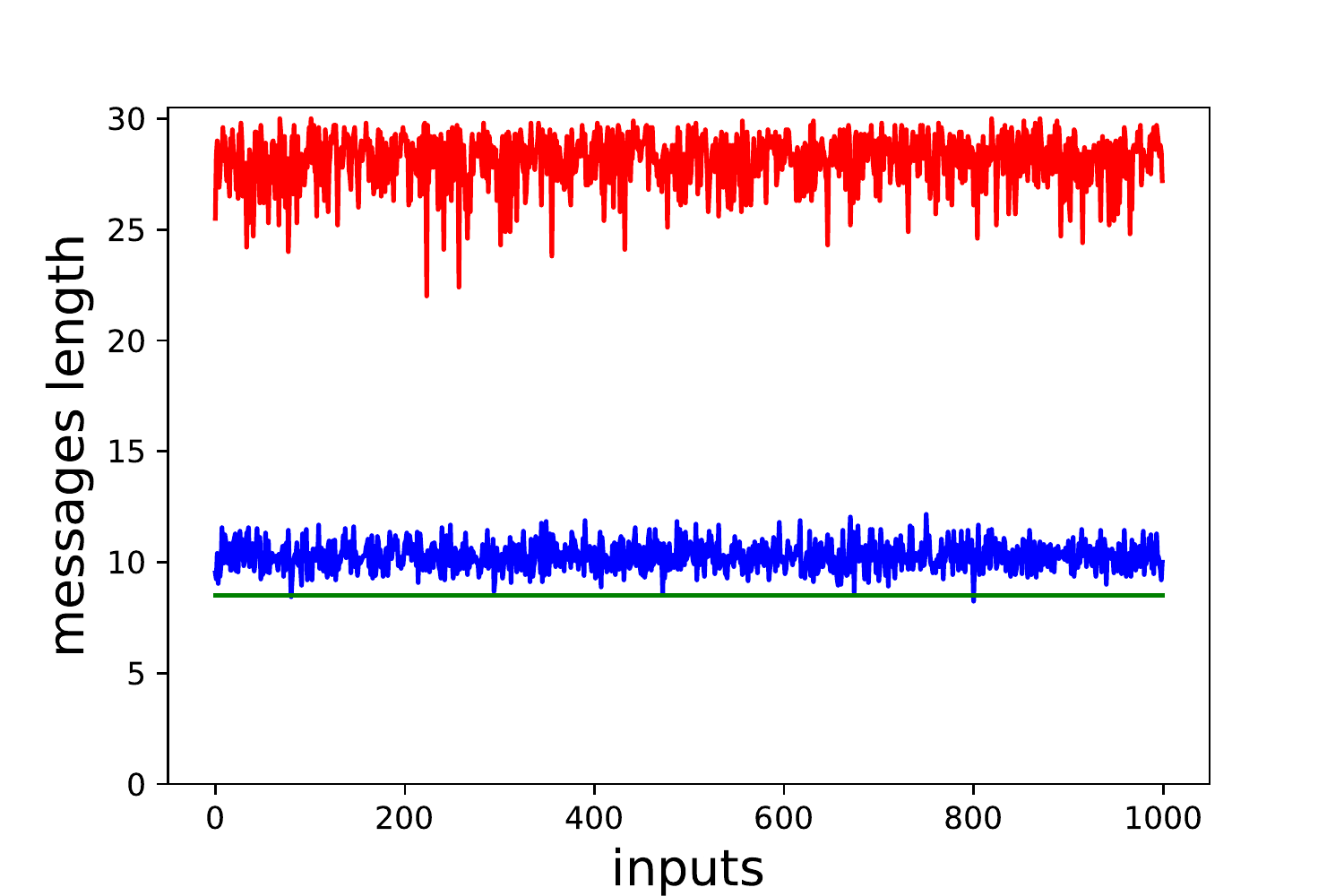} 
}
\subfigure[$a$=$5$]{
    \hspace{-1.5\baselineskip}	
    \includegraphics[width=0.25\textwidth, keepaspectratio]{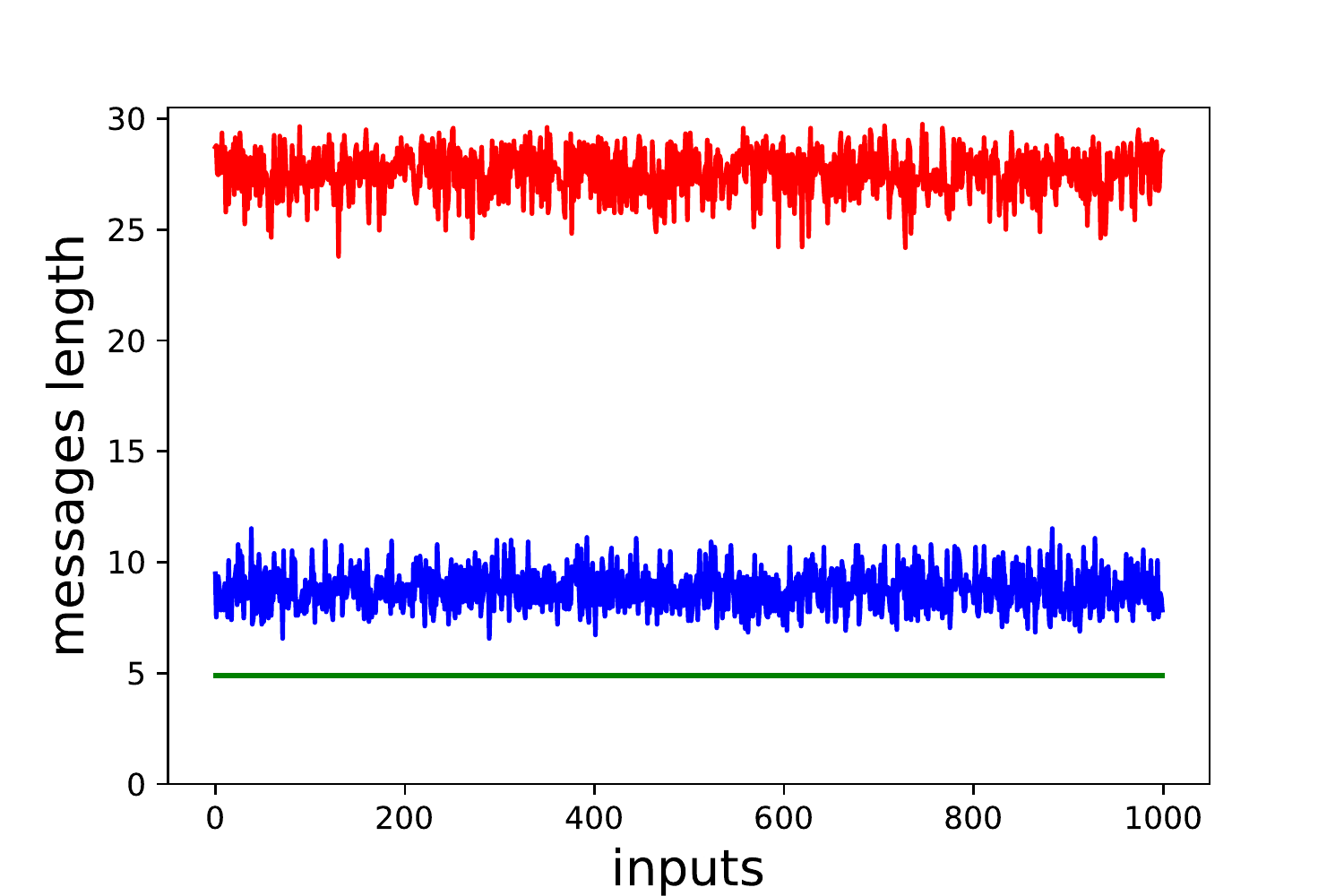}
}
\subfigure[ $a$=$10$]{
    \hspace{-1.5\baselineskip}
    \includegraphics[width=0.25\textwidth, keepaspectratio]{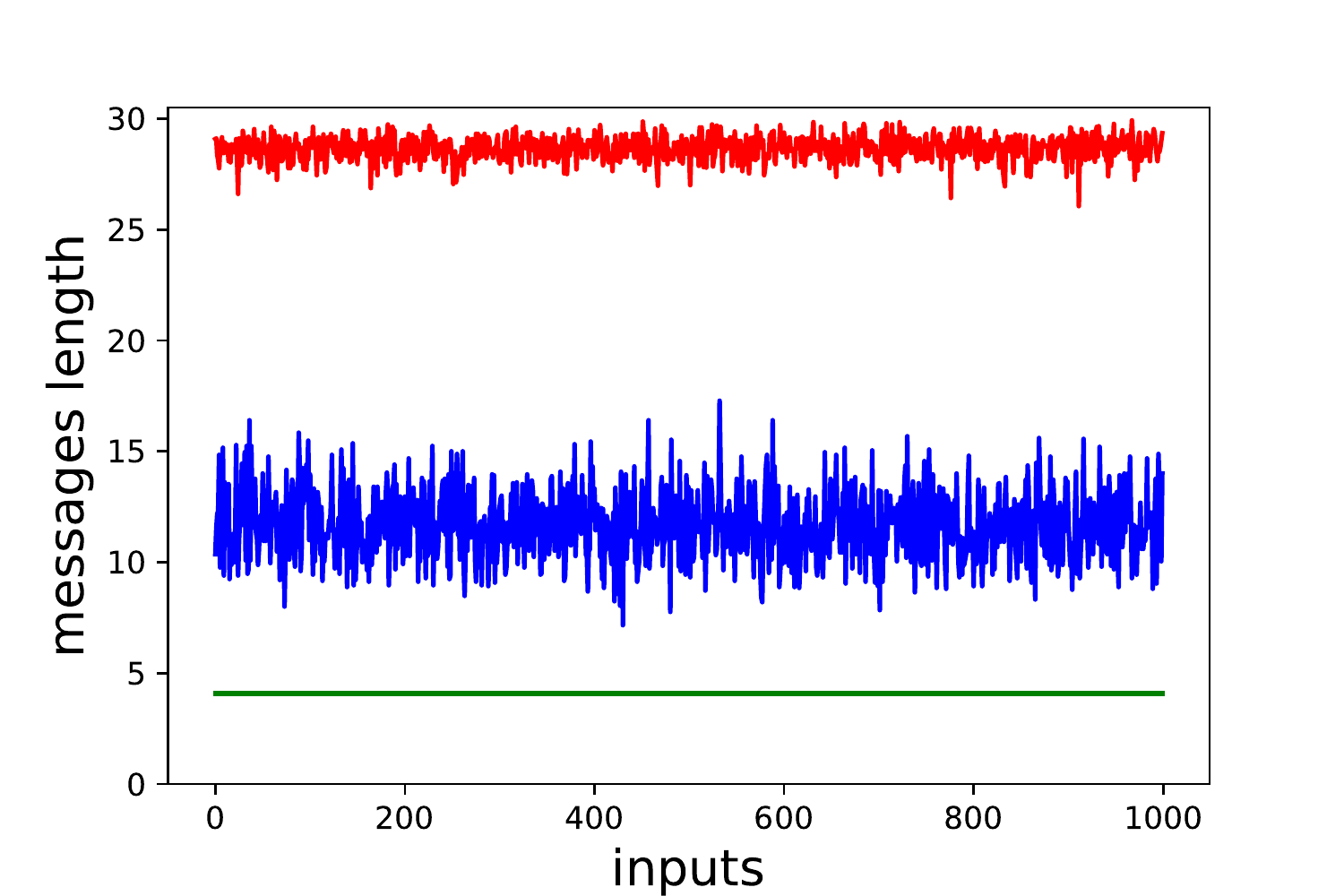}    
}
\subfigure[ $a$=$40$]{
    \hspace{-1.5\baselineskip}
    \includegraphics[width=0.25\textwidth, keepaspectratio]{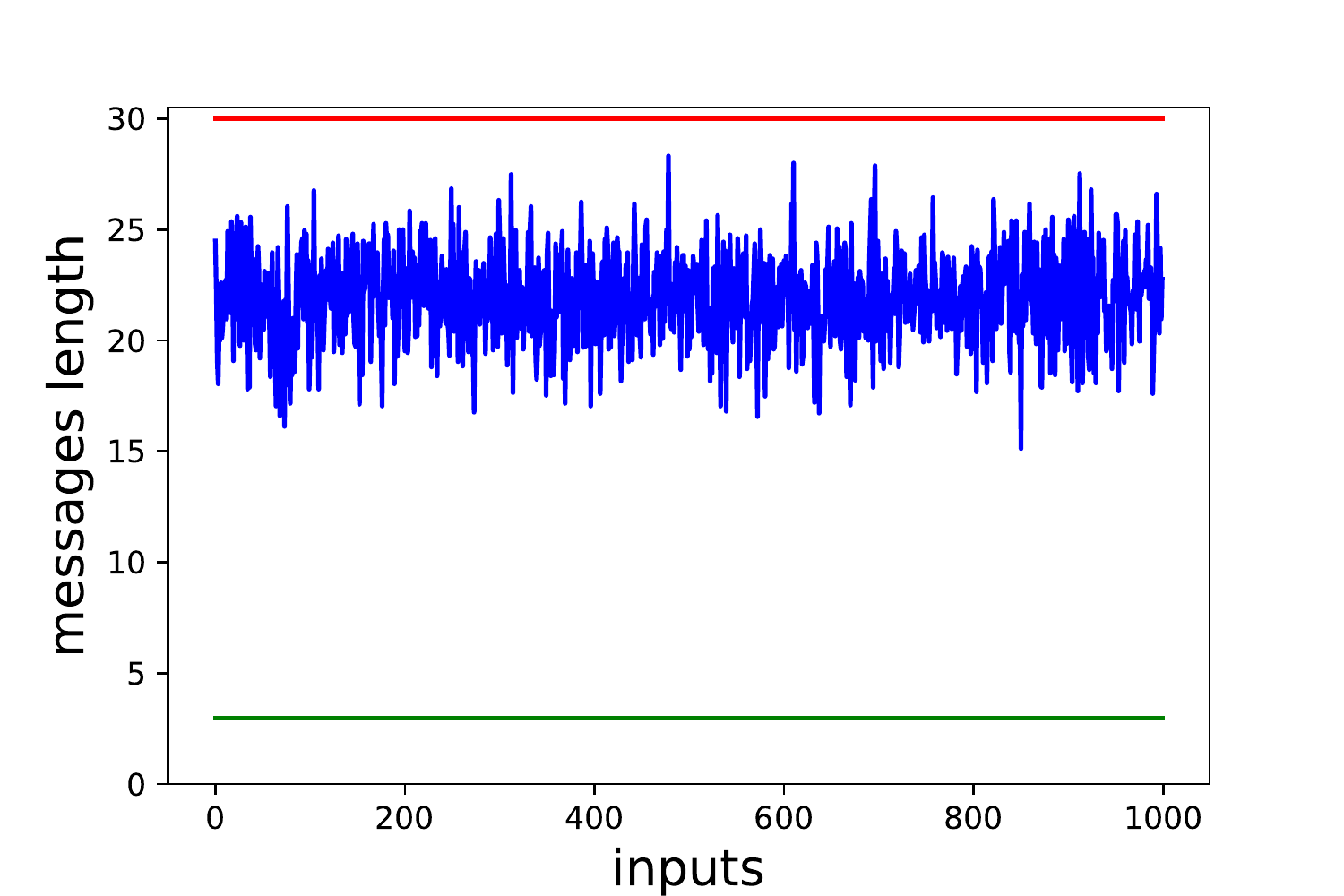}    
}{
    \includegraphics[width=\textwidth]{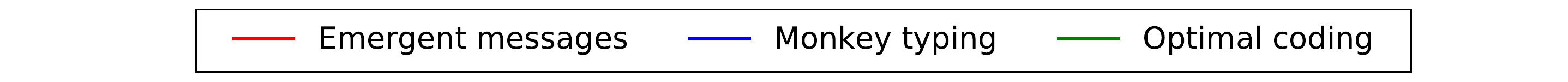}    
}
\vspace{-.5\baselineskip}
\caption{Mean message length per input across successful runs for \texttt{max_len=30} and different $a$. Inputs are uniformly
distributed.}\label{fig:unif}
\end{figure*}

Agents’ messages are very long also when the input distribution is uniform, see Figure \ref{fig:unif}. Their average length is significantly larger than MT messages with uniform inputs (t-test, $p < 10^{-9}$).

\subsection{Randomization test}
\label{sec:radomtest}

In the main paper, we observe a tendency for Speaker to use longer messages for frequent inputs, making its code obey a sort of ``\emph{anti}-ZLA''. In this section, we provide quantitative support for this observation. We run the randomization test of  \citet{ferrer2013}.  We note $E=\sum_{i=1}^{1000} p_i \times {l_i}$ the mean length of messages, where $p_i$ is the probability of the type $i$ and $l_i$ is the length of the corresponding message. A language that respects ZLA is characterized by a small $E$ (optimal coding, OC, is associated with $min(E)$).
Under $H_0$, the mean length of the encoding coincides with the mean length of a random permutation of messages across types. To be comparable with \citet{ferrer2013}, we use the same number of permutations ($=10^{5}$). Also, we adopt their definition of ``left p-value'' and ``right p-value''. If left p-value$\leq 0.005$, the studied encoding is \emph{significantly small} (characterized by significantly smaller $E$ than random permutations), if right p-value$ \leq 0.005$, it is \emph{significantly large}, corresponding to our notion of anti-efficiency. 

We observe in Table \ref{table:significant} that $H_0$ is only rejected for MT with $a \geq 40$, which, as we mentioned in the main paper, approaches a random length distribution for those cases, and for emergent messages with $a=1000$. OC, natural languages, and emergent language \emph{with} Speaker-length regularization are, in all the considered settings, significantly more efficient than chance. Importantly, the Emergent language results confirm LSTMs' natural preference for long messages ($E$ approaching $\texttt{max_len}$) and \emph{significant} anti-efficiency for $a \leq 40$ (right p-value$\approx{}0$). When $a=1000$, there is no frequency rank/length relation and all lengths $\approx{}\texttt{max_len}$.

\begin{table}[tb]
\tiny
    \caption{Results of the randomization test for different codes when $\texttt{max_len}=30$ and with different alphabet sizes $a$. Left/right p-values significant at $\alpha=0.01$ suffixed by asterisk. See Table 1 of \citet{ferrer2013}  for more codes to be compared with our results. } \label{table:significant}
    \centering
\begin{adjustbox}{width=1\textwidth}
\begin{tabular}{| r || r | c | c | c |}
\hline
  Setting              & Code & $E$ & Left p-Value & Right p-Value\\
\hline
\multirow{3}{*}{$a=5$} & OC & $3.55$ & $<10^{-5}$*  & $>1-10^{-5}$  \\
            & MT & $7.56$ & $<10^{-5}*$ & $>1-10^{-5}$  \\
           & Emergent & $26.98$ & $>1-10^{-5}$ & $<10^{-5}*$ \\
\hline
\multirow{3}{*}{$a=10$} & OC & $2.82$ & $<10^{-5}*$ & $>1-10^{-5}$  \\
            & MT & $11.27$ & $0.0002$* & $0.998$   \\
           & Emergent & $26.73$ & $>1-10^{-5}$ & $<10^{-5}$*  \\
\hline
\multirow{6}{*}{$a=40$} & OC & $2.29$ & $<10^{-5}$* & $>1-10^{-5}$  \\
            & MT & $21.30$ & $0.814$ & $0.186$  \\
           & Emergent & $29.40$ & $>1-10^{-5}$ & $<10^{-5}$*  \\
           & Regularized ($\alpha$=$0.5$) & $7.22$ &  $<10^{-5}$* & $>1-10^{-5}$ \\
     & English & $3.68$ & $<10^{-5}$* & $>1-10^{-5}$   \\
     & Arabic & $3.14$ & $<10^{-5}$* & $>1-10^{-5}$  \\     
\hline
\multirow{3}{*}{$a=1000$} & OC & $1.86$ & $0.001$*  & $0.999$  \\
            & MT & $29.67$ & $0.750$ & $0.250$  \\
           & Emergent & $29.98$ & $0.072$ & $0.928$ \\
\hline
\end{tabular}
\end{adjustbox}
\end{table}

\subsection{Speaker initial length distribution}
\label{sec:speakerb}

\begin{figure*}[ht]
\centering
\subfigure[$(6, 5)$]{
    \includegraphics[width=0.25\textwidth, keepaspectratio]{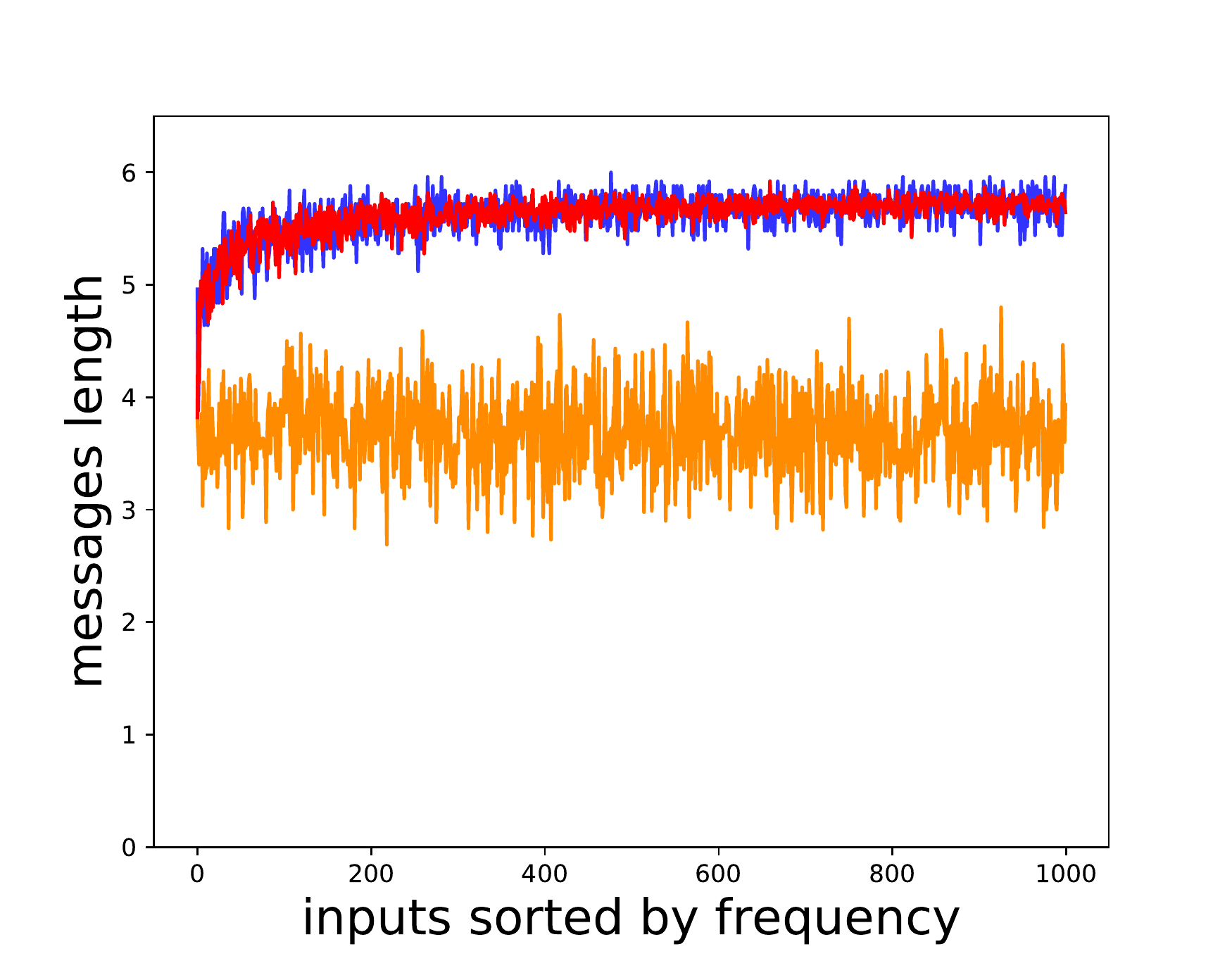} 
    \label{fig:A}
}
\subfigure[$(6, 10)$]{
 \hspace{-1.5\baselineskip}
    \includegraphics[width=0.25\textwidth, keepaspectratio]{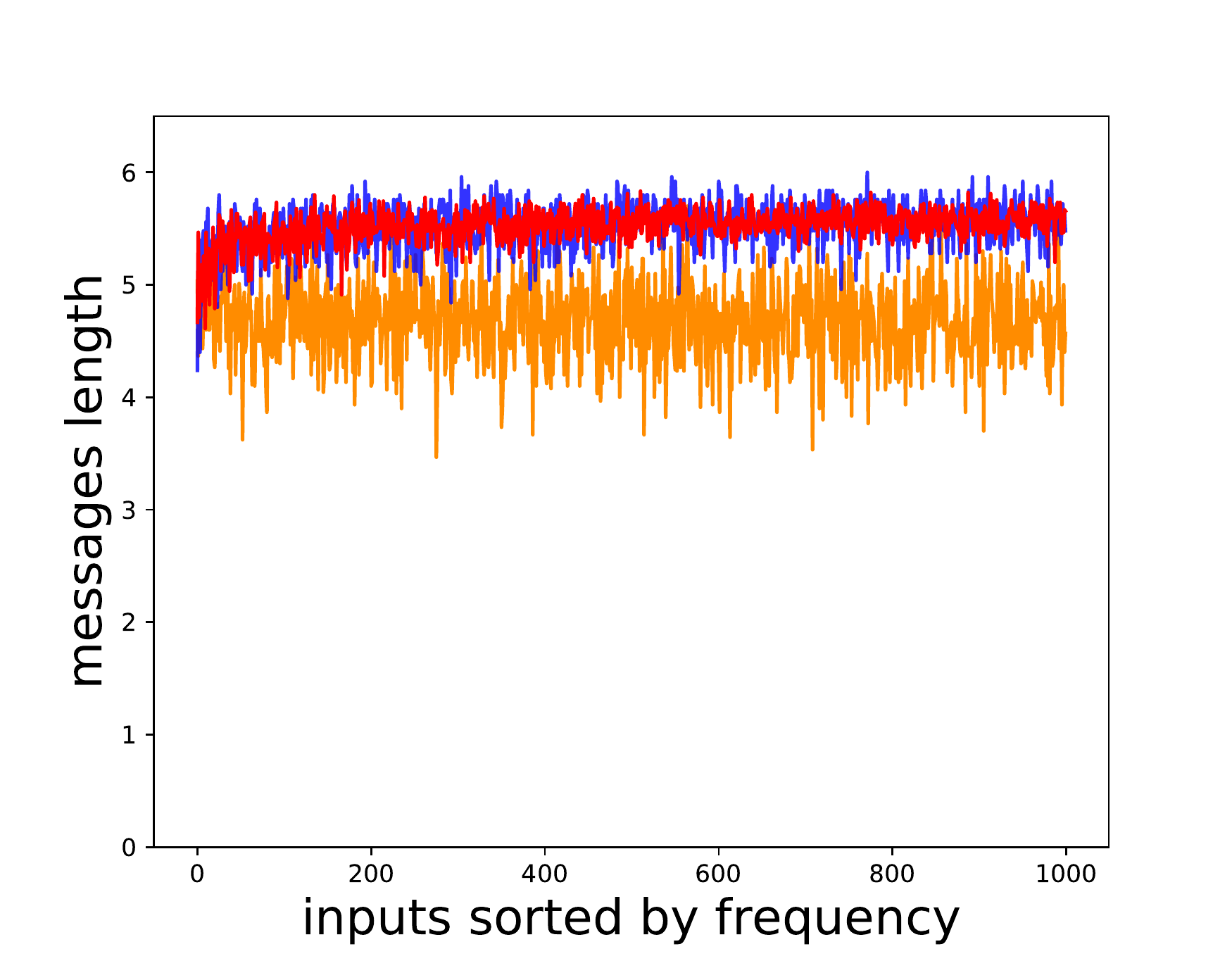} 
    \label{fig:A}
}
\subfigure[$(6, 40)$]{
    \hspace{-1.5\baselineskip}	
    \includegraphics[width=0.25\textwidth, keepaspectratio]{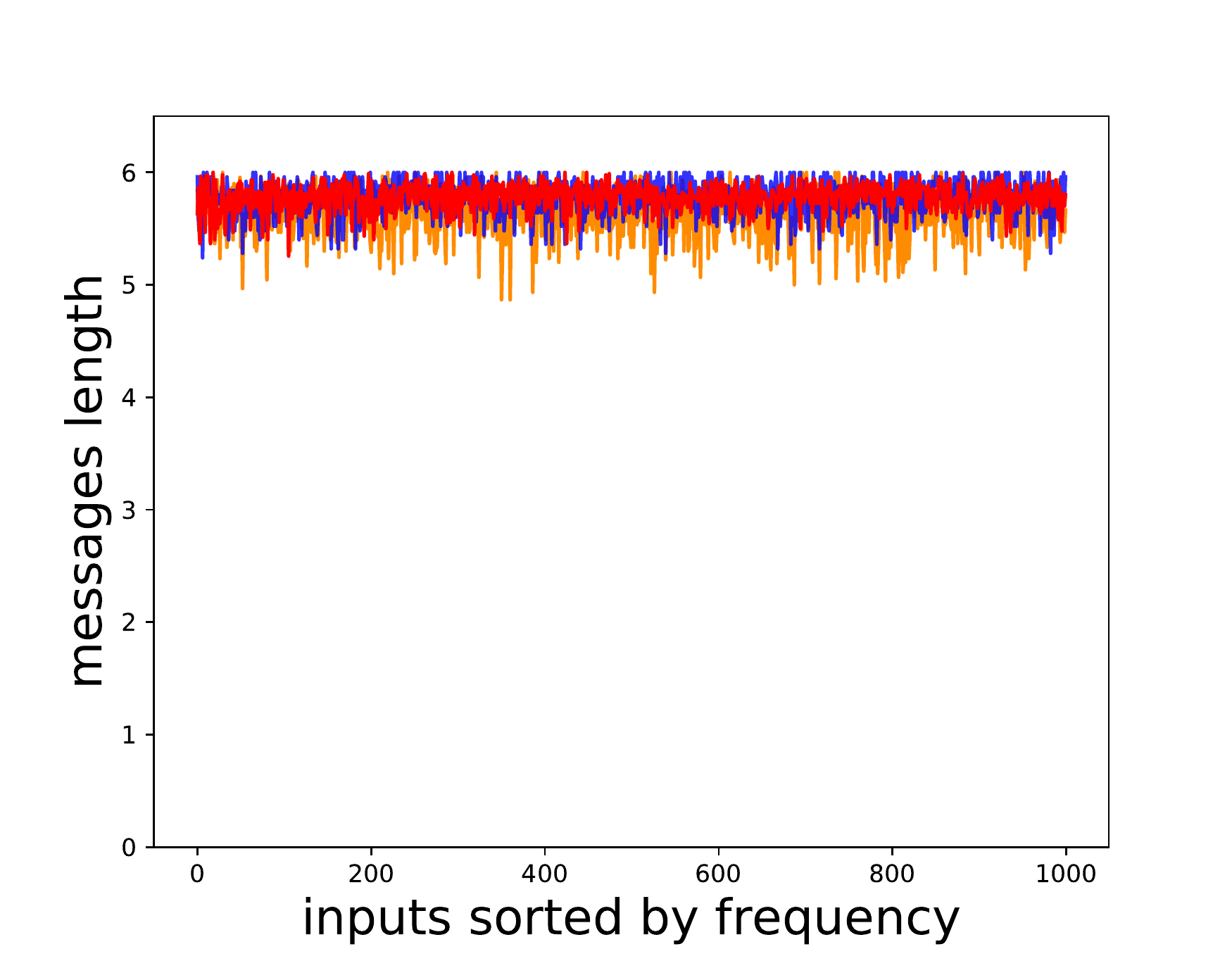}
}
\subfigure[$(6, 1000)$]{
 \hspace{-1.5\baselineskip}
   \includegraphics[width=0.25\textwidth, keepaspectratio]{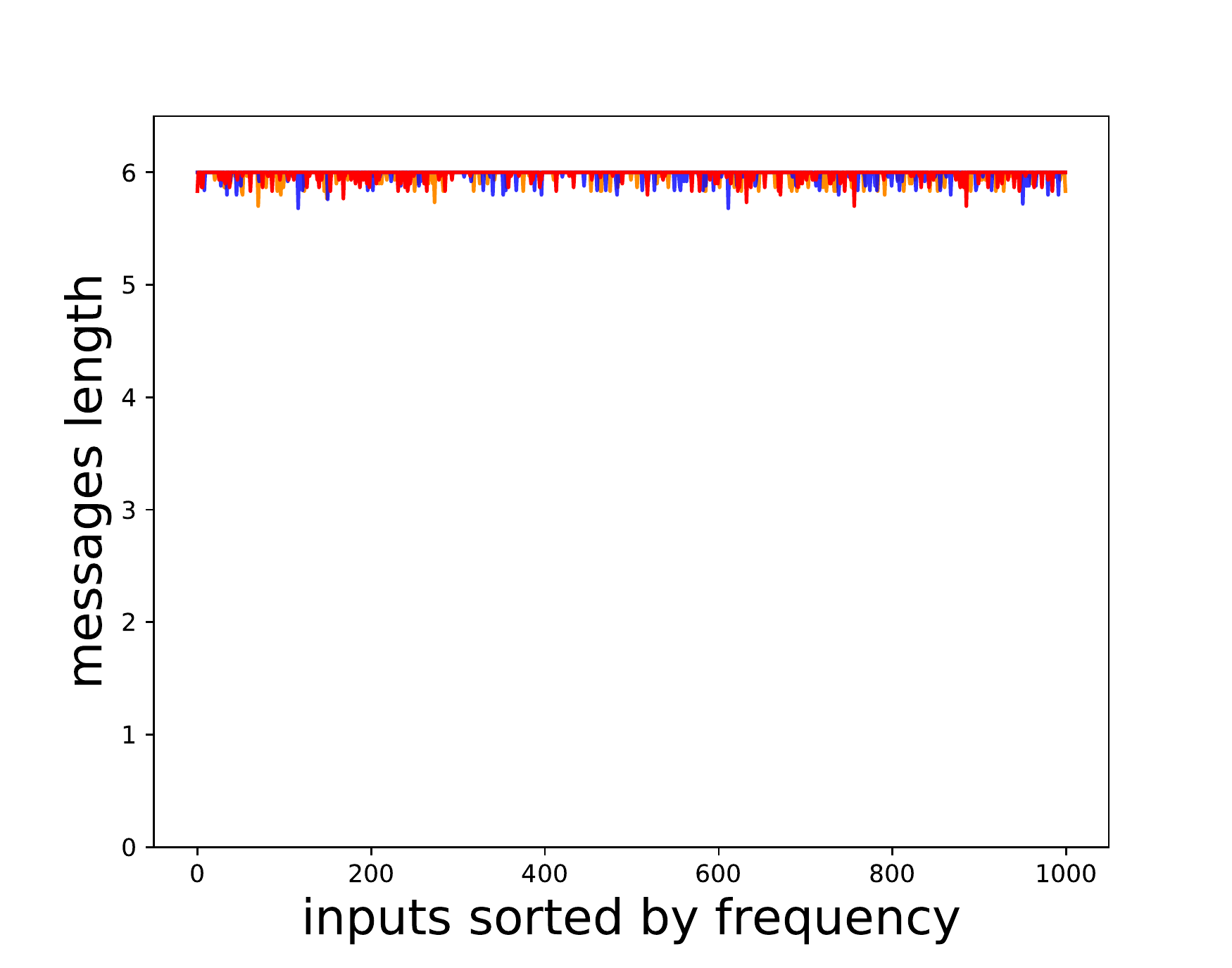}
}
\subfigure[$(11, 3)$]{
    \includegraphics[width=0.2\textwidth, keepaspectratio]{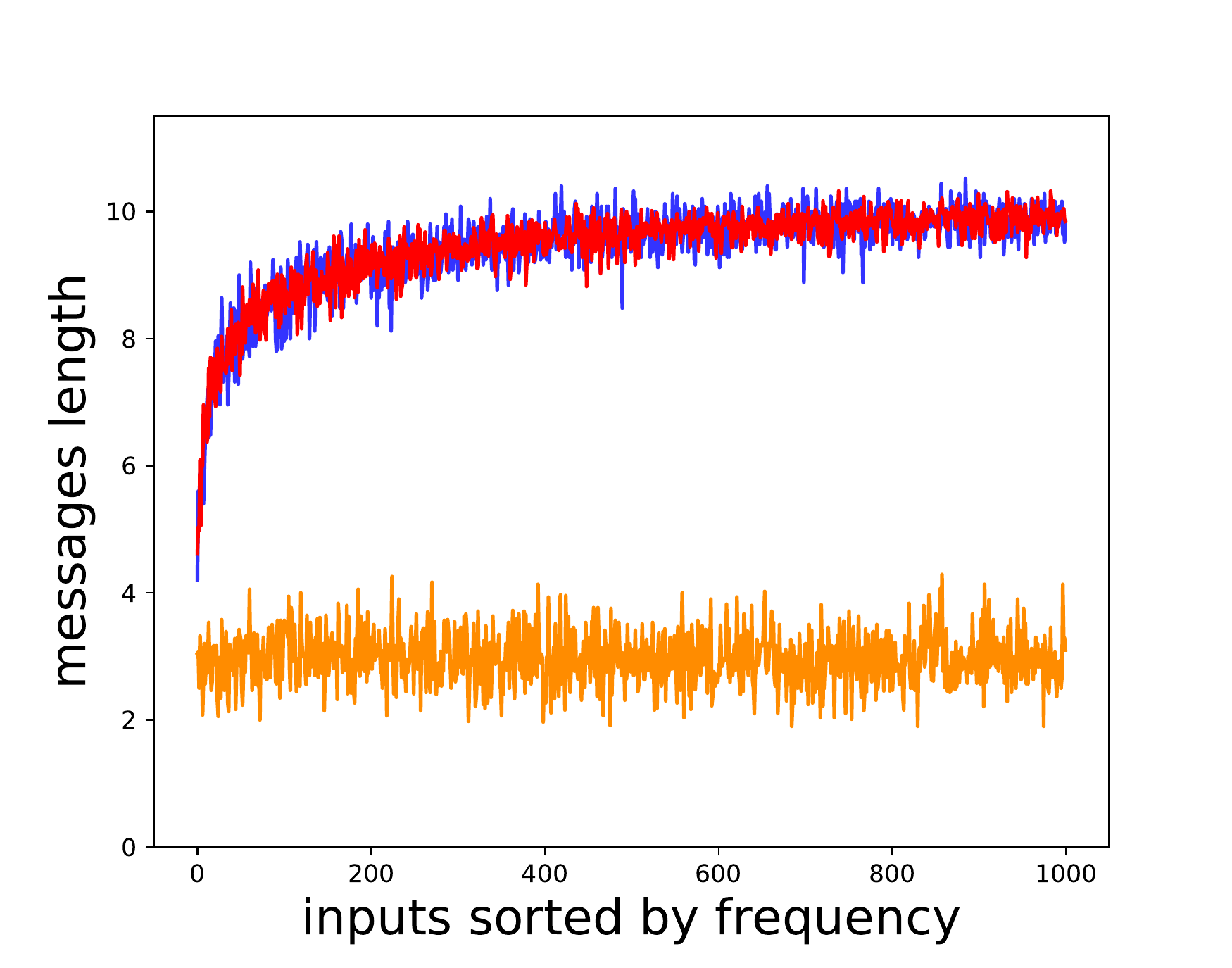}    
}
\subfigure[$(11, 5)$]{
 \hspace{-1.5\baselineskip}
    \includegraphics[width=0.2\textwidth, keepaspectratio]{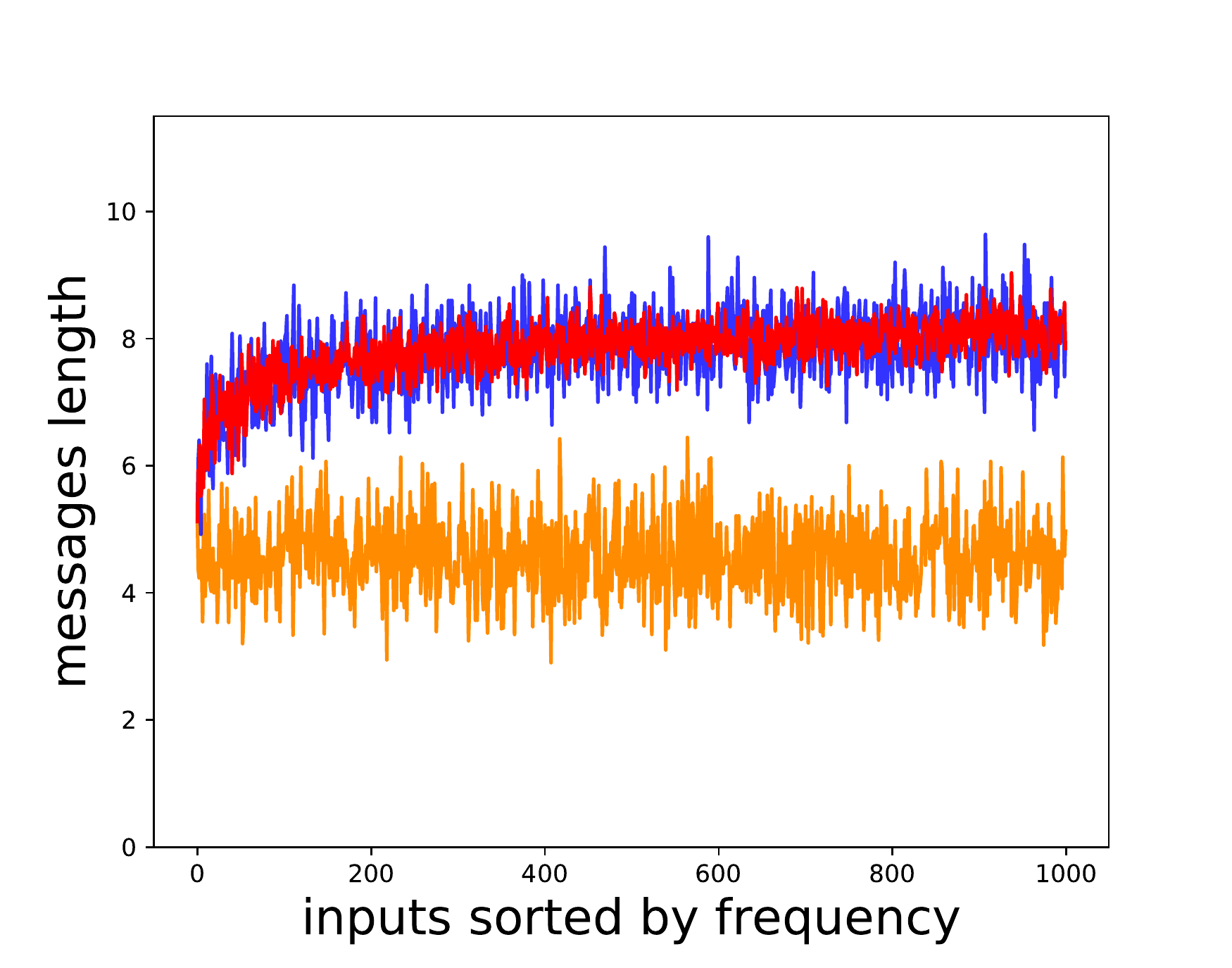}    
}
\subfigure[$(11, 10)$]{
    \hspace{-1.5\baselineskip}
    \includegraphics[width=0.2\textwidth, keepaspectratio]{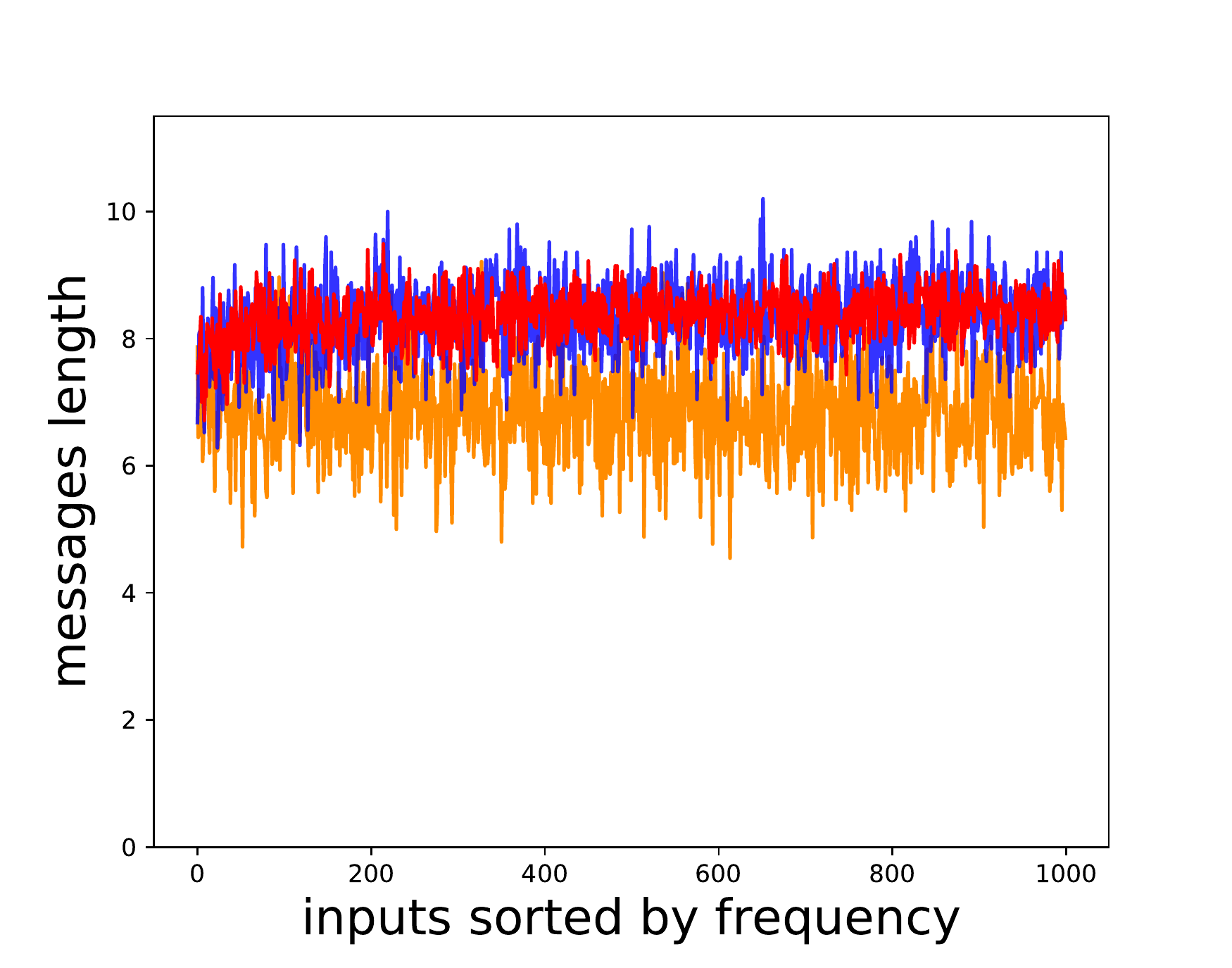}    
}
\subfigure[$(11, 40)$]{
    \hspace{-1.5\baselineskip}
    \includegraphics[width=0.2\textwidth, keepaspectratio]{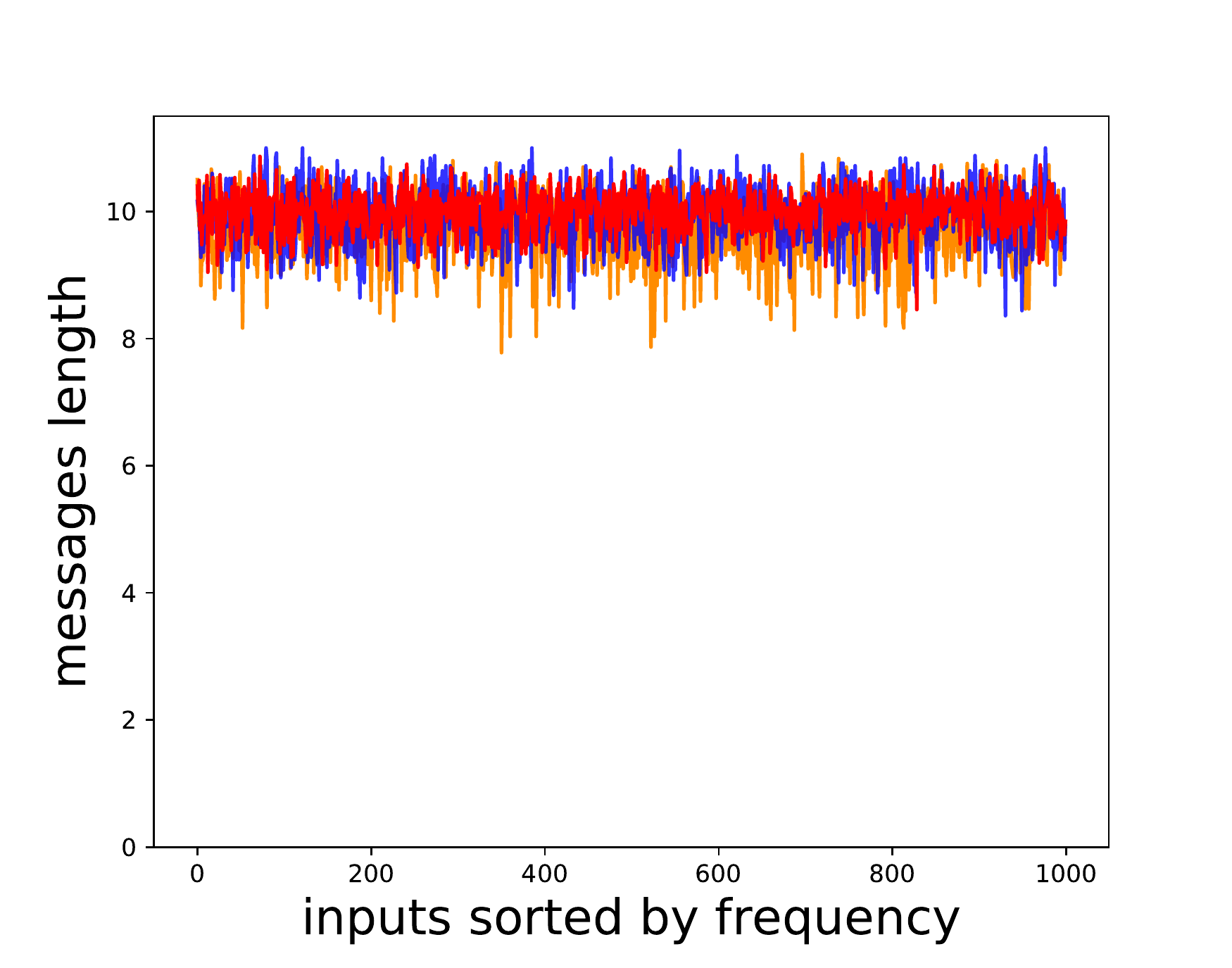}    
}
\subfigure[$(11, 1000)$]{
    \hspace{-1.5\baselineskip}
    \includegraphics[width=0.2\textwidth, keepaspectratio]{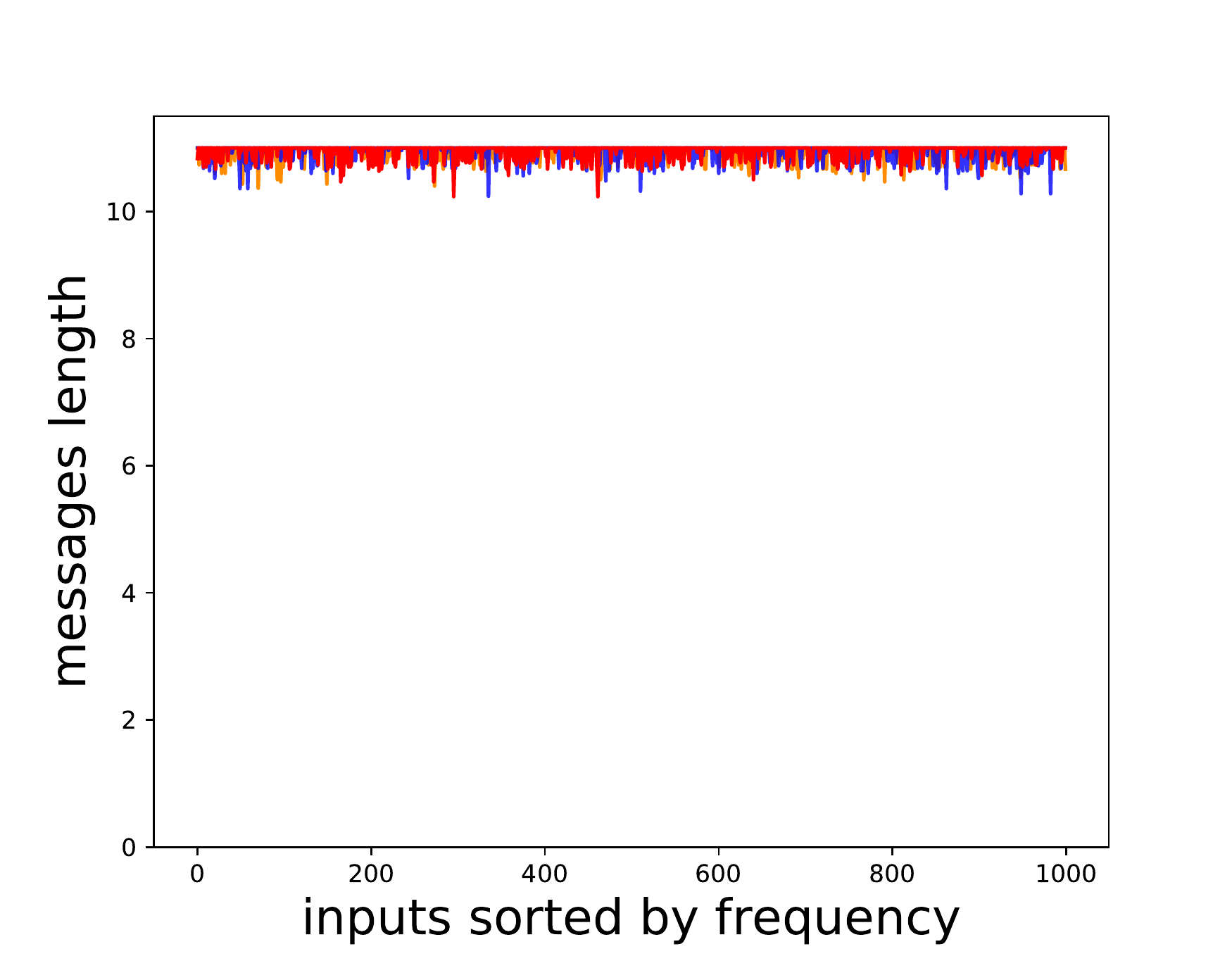}    
}
\subfigure[$(30, 3)$]{
    \includegraphics[width=0.20\textwidth, keepaspectratio]{figs/AsenderbiasesML30VS3.pdf}    
}
\subfigure[$(30, 5)$]{
    \hspace{-1.5\baselineskip}
    \includegraphics[width=0.20\textwidth, keepaspectratio]{figs/AsenderbiasesML30VS5.pdf}    
}
\subfigure[$(30, 10)$]{
    \hspace{-1.5\baselineskip}
    \includegraphics[width=0.20\textwidth, keepaspectratio]{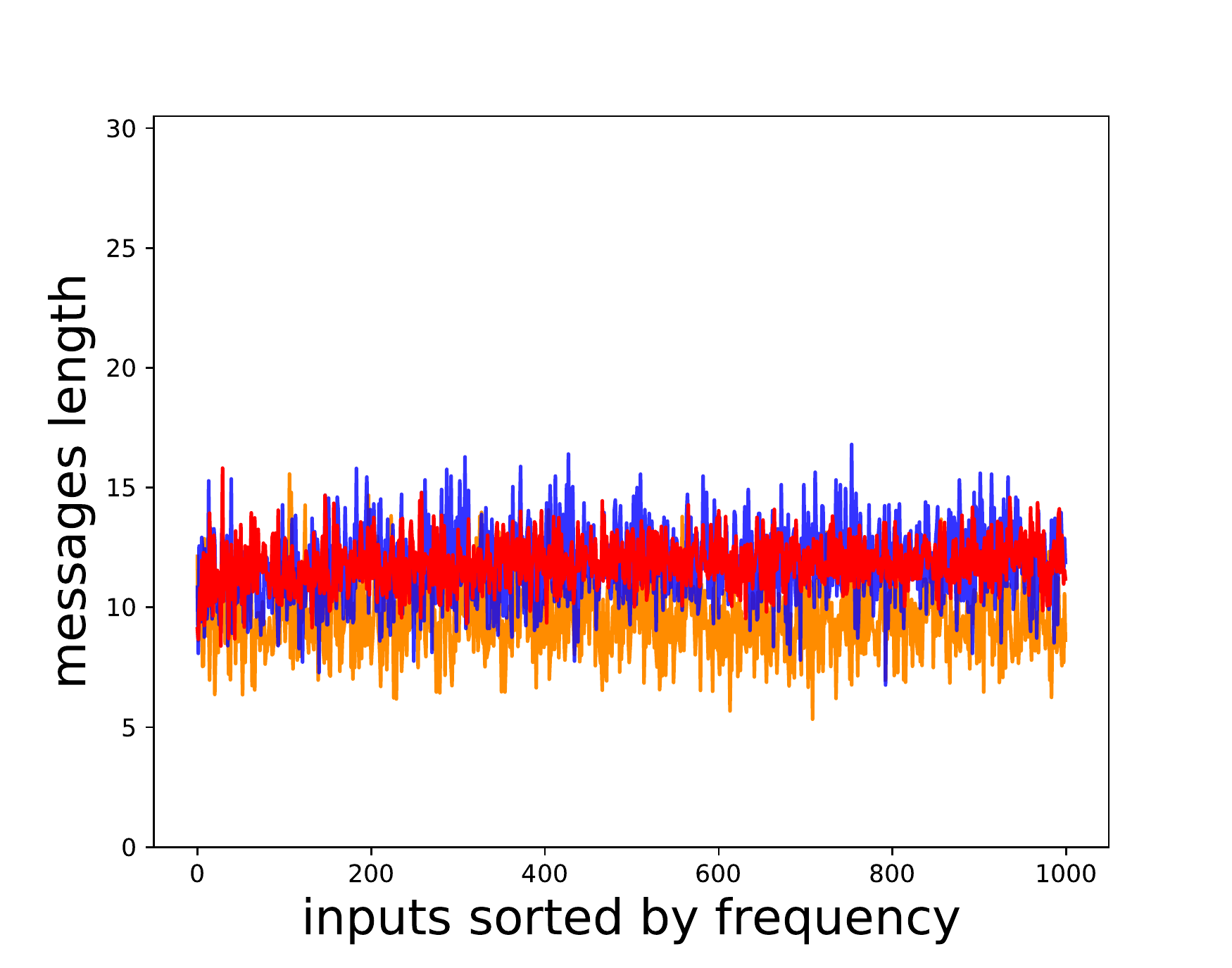}    
}
\subfigure[$(30, 40)$]{
    \hspace{-1.5\baselineskip}
    \includegraphics[width=0.20\textwidth, keepaspectratio]{figs/AsenderbiasesML30VS40.pdf}    
}
\subfigure[$(30, 1000)$]{
    \hspace{-1.5\baselineskip}
    \includegraphics[width=0.20\textwidth, keepaspectratio]{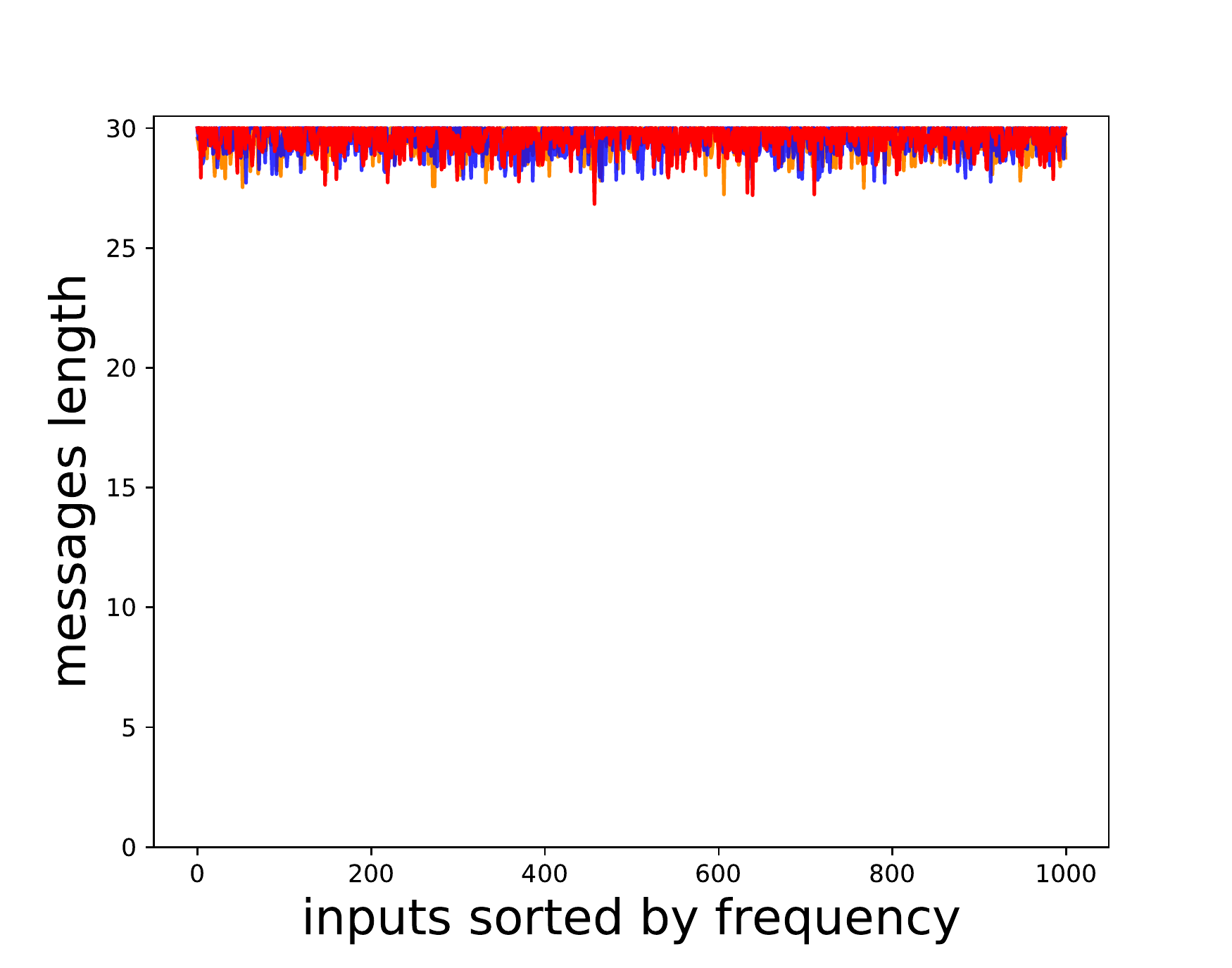}    
}
{
    \includegraphics[width=\textwidth]{figs/legendsenderbiases.pdf}    
}
\vspace{-.5\baselineskip}
\caption{Average length of messages in function of input frequency rank for untrained Speakers, compared to MT. In each figure we report the results in a specific setting $(\texttt{max_len}, a)$.}\label{fig:SendBiasInAnnex}
\end{figure*}

Figure \ref{fig:SendBiasInAnnex} plots message length in function of input frequency rank for several settings. In particular, we report \emph{all} settings $(\texttt{max_len}, a)$ that succeeded when training the Speaker-Listener system. Here, however, no training is performed, so that we can observe Speaker's initial biases. The results are in line with our finding in Section \ref{sec:speakerbias} of the main paper.

\subsection{The effect of length regularization}
\label{sec:regeffect}

\begin{figure*}[ht]
\centering
\subfigure[$\alpha$=$0.1$]{
    \includegraphics[width=0.33\textwidth, keepaspectratio]{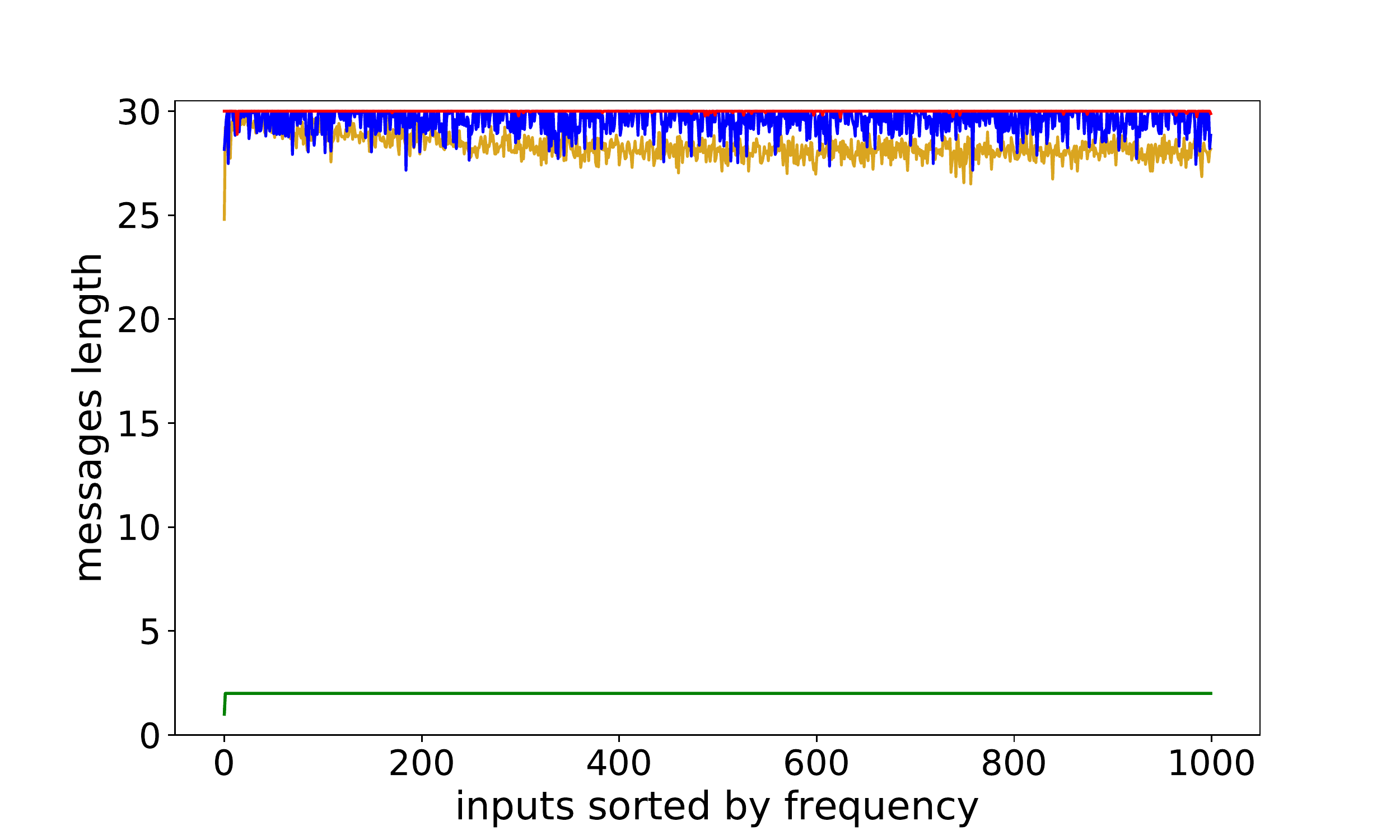} 
}
\subfigure[$\alpha$=$0.5$]{
   \hspace{-1.5\baselineskip}
    \includegraphics[width=0.33\textwidth, keepaspectratio]{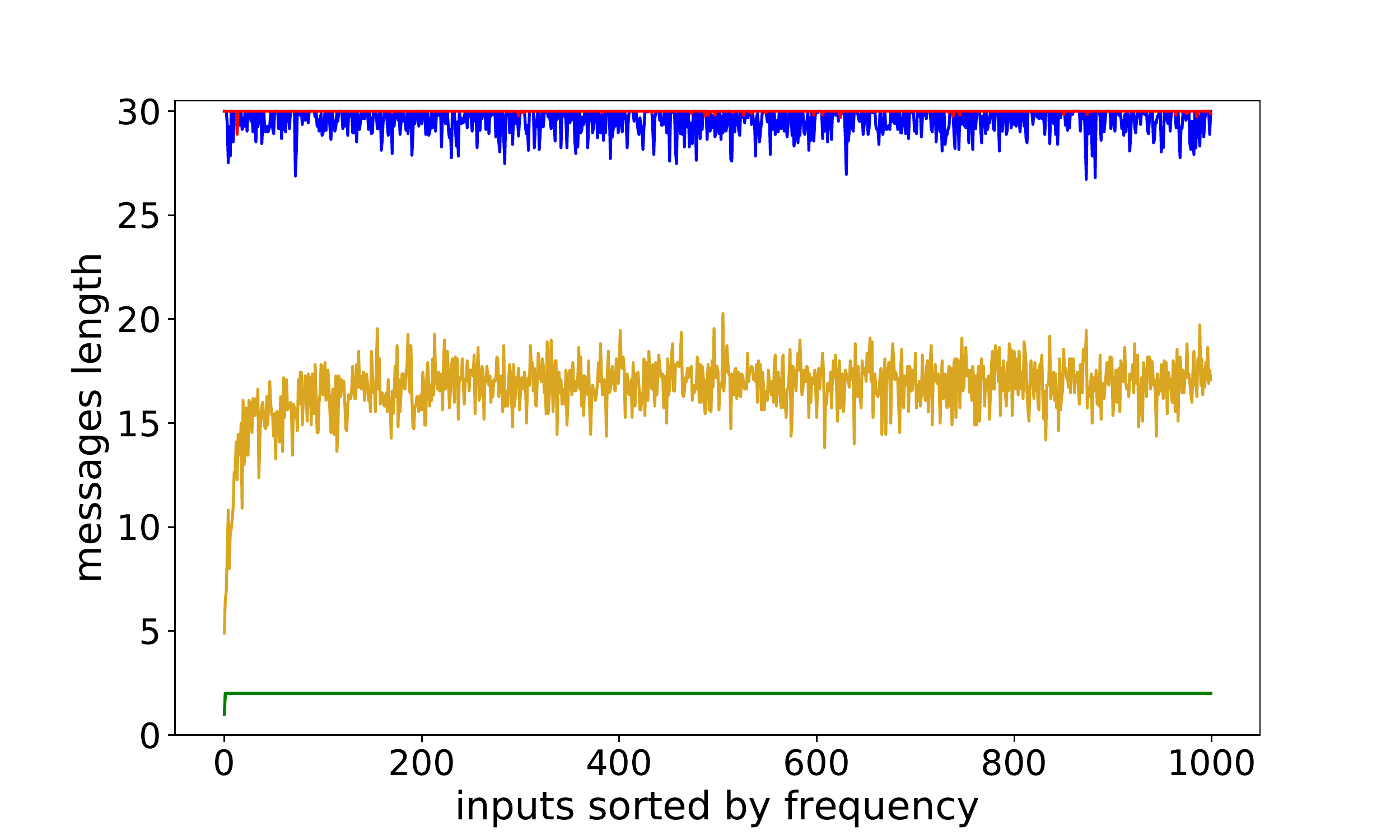} 
}
\subfigure[$\alpha$=$1$]{
    \hspace{-1.5\baselineskip}	
    \includegraphics[width=0.33\textwidth, keepaspectratio]{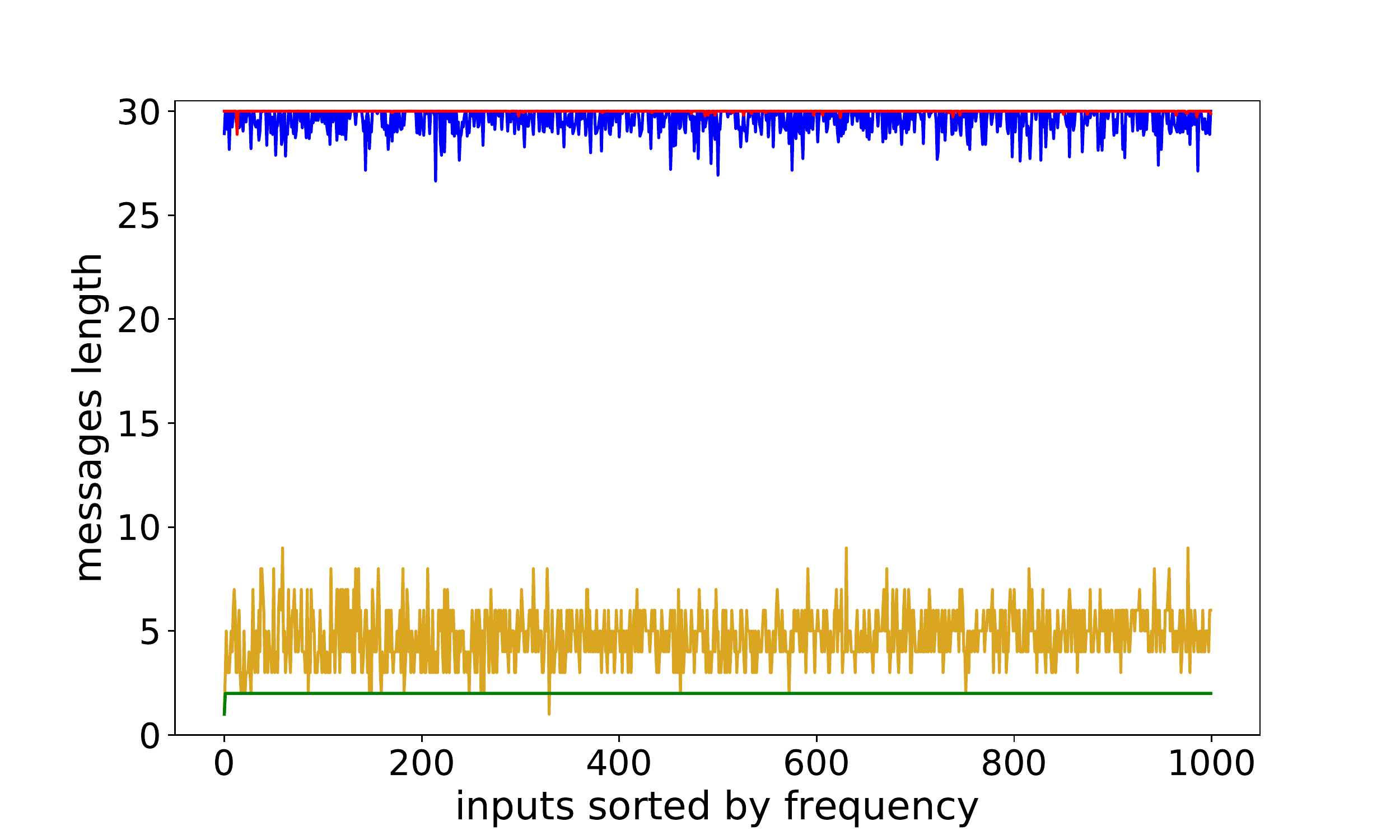}
}

{
    \includegraphics[width=\textwidth]{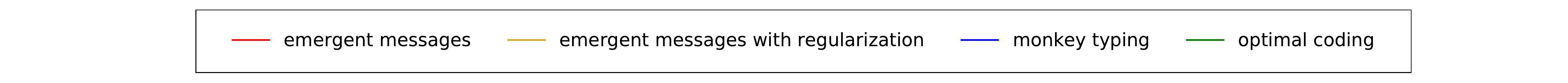}    
}
\vspace{-.5\baselineskip}
\caption{Length of messages as a function of input frequency for $\texttt{max_len}=30$ and $a=1000$, when varying $\alpha$ in the length regularization case.}\label{fig:RegularizationInAnnex}
\end{figure*}

We look here at the effect of the regularization coefficient $\alpha$ on the nature of the emergent encoding. To this end, we consider the setting that is least efficient when no optimization is applied: $(\texttt{max_len}=30, a=1000)$. The same pattern is also observed with different choices of \texttt{max_len} and \emph{a}. Figure \ref{fig:RegularizationInAnnex} shows, for $\alpha=1$, that emergent messages \emph{approximate optimal coding}. For even larger values, we were not able to successfully train the system to communicate. This is in line with Zipf's view of \emph{competing} pressures for accurate communication vs.~efficiency. The emergent messages follow ZLA only when both pressures are at work. If the efficiency pressure is not present, agents come up with a communicatively effective but non-efficient encoding, as shown in Section \ref{sec:antieff} and Section \ref{sec:emergentenc} of the main paper. However, if  the efficiency pressure is too high, agents cannot converge on a  protocol that is successful from the point of view of communication.

\section{Repetition in emergent messages}
\label{sec:examples}
\begin{figure*}[ht]
\centering
\includegraphics[width=0.6\textwidth]{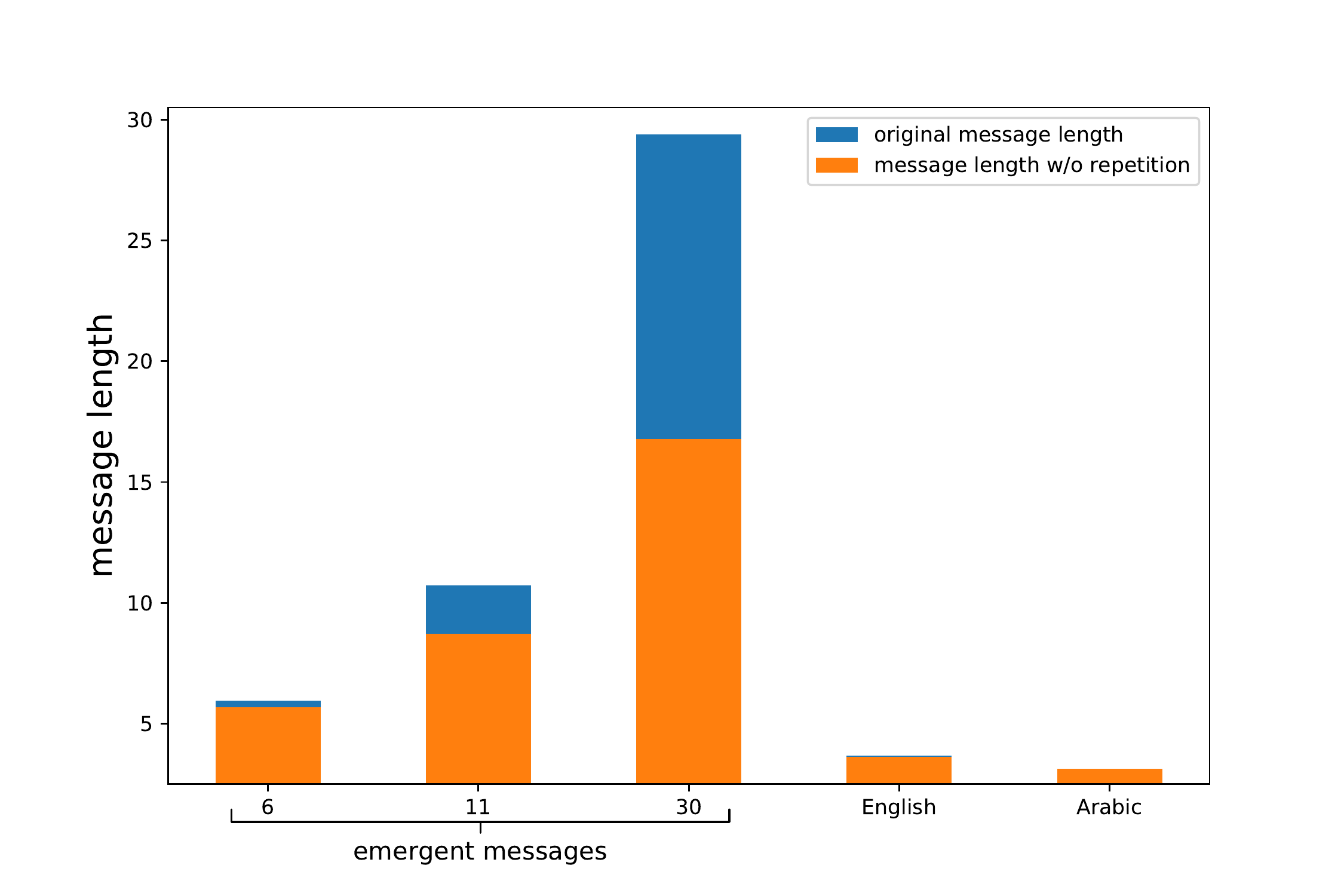} 
\caption{Mean message length (weighted by input probability, and
  averaged across successful runs) for various $\texttt{max_len}$
  and fixed $a=40$, before and after removing all repetitions. A repetition
  here refers to a sequence of 2 or more consecutive identical
  symbols. Emergent messages are indexed by their $\texttt{max_len}$, and
  we add the same statistics in two human languages for
  comparison.}\label{fig:rep}
\end{figure*}

We report in listings \ref{lst:example1}, \ref{lst:example2}, \ref{lst:example3} and \ref{lst:example4} examples of emergent messages in different settings. We notice that the agents extensively use repetition, even when $a$ (vocabulary size) is large. This repetition, that results in the very skewed bigram distributions presented in Section \ref{sec:symbolDist} of the main paper, increases with higher $\texttt{max_len}$, as shown in figure \ref{fig:rep}. Moreover, from figure \ref{fig:rep}, we see that, unlike in emergent codes, this sort of repetition does not appear in natural language.

\begin{lstlisting}[mathescape=true, caption=Emergent messages for the $4$ most frequent inputs ($\texttt{max_len}$:$11$ and $a$:$40$)., xleftmargin=.05\textwidth, label=lst:example1, basicstyle=\small]
m1: 18,5,36,36,5,5,10,5,32,8,eos
m2: 1,36,2,36,10,13,9,29,33,eos
m3: 29,1,8,1,39,39,9,15,10,19,eos
m4: 29,1,36,36,36,36,5,8,13,9,eos
\end{lstlisting}

\begin{lstlisting}[mathescape=true, caption=Emergent messages for the $4$ most frequent inputs ($\texttt{max_len}$:$11$ and $a$:$1000$)., xleftmargin=.05\textwidth, label=lst:example2, basicstyle=\small]
m1: 431,431,305,305,70,70,331,391,134,581,eos
m2: 867,288,466,466,466,737,113,77,615,615,eos
m3: 288,466,466,466,418,144,113,615,638,615, eos
m4: 4,4,152,152,152,468,642,615,422,134,eos
\end{lstlisting}

\begin{lstlisting}[mathescape=true, caption=Emergent messages for the $4$ most frequent inputs ($\texttt{max_len}$:$30$ and $a$:$5$)., xleftmargin=.05\textwidth, label=lst:example3, basicstyle=\small]
m1: 3,4,4,4,1,1,1,1,1,1,1,1,4,4,4,4,4,4,4,4,4,4,4,4,4,3,4,3,4,eos
m2: 3,1,3,3,1,1,1,1,1,1,4,4,4,4,4,2,4,2,4,2,4,2,4,2,4,2,4,3,2,eos
m3: 1,4,4,1,1,1,1,1,1,1,4,4,4,4,4,2,4,4,4,4,4,4,4,4,4,2,4,3,1,eos
m4: 1,4,4,1,1,1,1,1,1,1,4,4,4,4,4,4,4,4,4,4,4,4,2,4,2,2,4,1,4,eos
\end{lstlisting}

\begin{lstlisting}[mathescape=true, caption=Emergent messages for the $4$ most frequent inputs ($\texttt{max_len}$:$30$ and $a$:$40$)., xleftmargin=.05\textwidth, label=lst:example4, basicstyle=\tiny]
m1: 11,11,12,24,8,8,12,24,12,12,12,12,12,12,36,24,24,35,35,35,36,36,20,15,36,19,11,31,13, eos
m2: 13,31,31,24,8,8,8,8,8,8,8,8,8,19,24,3,3,36,36,19,29,15,31,30,31,15,19,11,13,eos
m3: 39,8,12,8,8,8,8,25,25,25,25,25,25,25,36,24,12,12,35,35,35,18,18,11,3,7,11,7,11,eos
m4: 14,31,8,8,8,8,8,8,24,25,25,25,36,36,36,36,36,36,36,36,36,36,3,2,35,30,31,21,29,eos
\end{lstlisting}

\subsection{Entropy of symbol distributions in different codes}
\label{sec:entropy}
\begin{figure*}[ht]
\centering
\subfigure[Entropy of unigram distributions]{
    \includegraphics[width=0.50\textwidth, keepaspectratio]{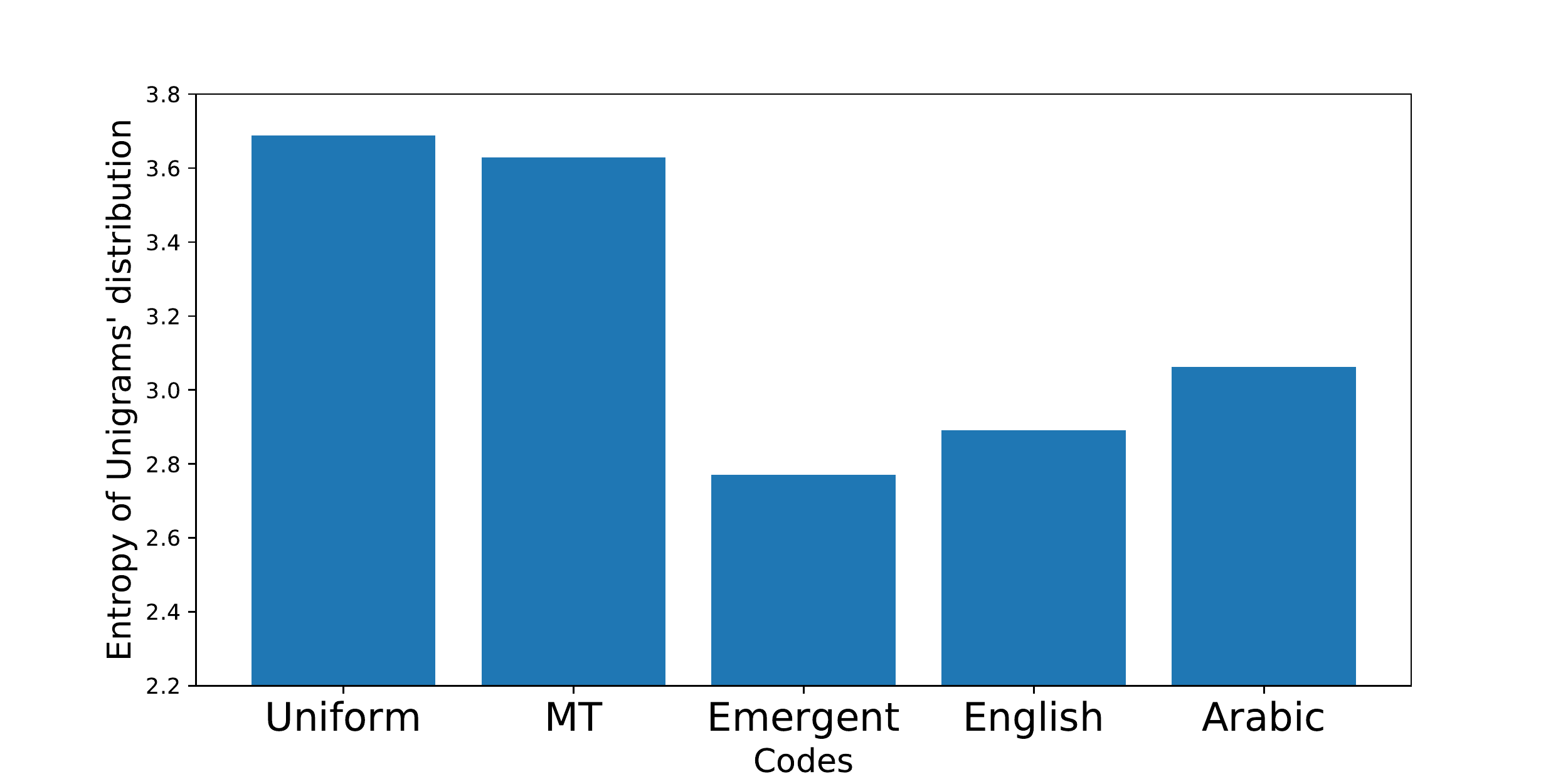} 
    \label{fig:unigramEnt}
}
\subfigure[Entropy of bigram distributions]{
   \hspace{-1.5\baselineskip}
    \includegraphics[width=0.50\textwidth, keepaspectratio]{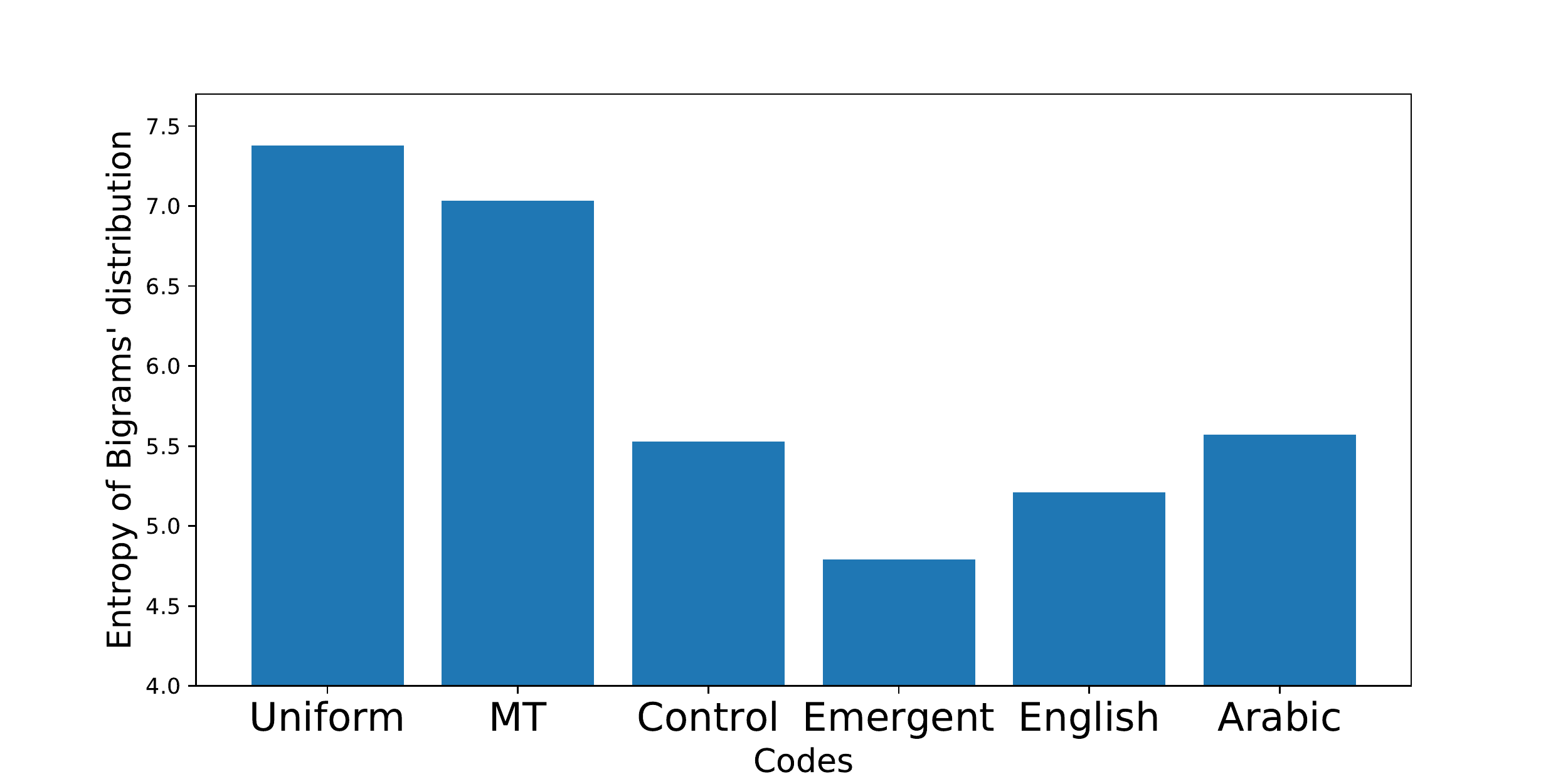} 
    \label{fig:bigramEnt}
}
\vspace{-.5\baselineskip}
\caption{Entropy of symbol unigram and bigram distributions for different codes (in natural log). The higher the entropy, the more uniform the corresponding distribution is. The entropy of the uniform code is computed by assuming a uniform distribution over $40$ symbols (unigram) and $1600$ sequences of $2$ symbols (bigram). MT and control messages (see Section \ref{sec:symbolDist} of main text) are averaged across $25$ different simulations in the ($\texttt{max_len}$=$30$,$a$=$40$) setting. Emergent messages are averaged across successful runs in the same setting.}\label{fig:vocabUse}
\end{figure*}

We report the entropy of symbol unigram and bigram distributions for different codes in figures \ref{fig:unigramEnt} and \ref{fig:bigramEnt}, respectively. We observe that, in both cases, the emergent code symbol distribution is more skewed than in any considered reference code.

\end{document}